\documentclass[review]{elsarticle}

\graphicspath{ {pic/} }
\usepackage{xltabular}
\usepackage{dcolumn}
\usepackage{threeparttablex}
\usepackage{caption}

\usepackage{amsmath}
\usepackage{booktabs}
\usepackage{subcaption}
\usepackage[colorinlistoftodos]{todonotes}
\usepackage{multirow}
\usepackage{tabularx}
\usepackage{rotating}
\usepackage[]{hyperref}
\usepackage{enumitem}
\hypersetup{
    pdftitle={Computational Models of Solving Raven's Progressive Matrices: A Comprehensive Introduction},
    pdfauthor={Yuan Yang},
    bookmarksnumbered=true,     
    bookmarksopen=true,         
    bookmarksopenlevel=1,       
    colorlinks=true,  
    citecolor=[rgb]{0, 0.5, 0.5},
    linkcolor=[rgb]{0, 0.5, 0.5},
    pdfstartview=FitH,
    pdfpagelayout=SinglePage
}

\usepackage{lineno}
\modulolinenumbers[5]

\journal{Journal of Artificial Intelligence}

\biboptions{authoryear}

\begin{document}

\begin{frontmatter}

\title{Computational Models of Solving Raven's Progressive Matrices: A Comprehensive Introduction}
\tnotetext[mytitlenote]{Fully documented templates are available in the elsarticle package on \href{http://www.ctan.org/tex-archive/macros/latex/contrib/elsarticle}{CTAN}.}

\author{Yuan Yang}
\author{Mathilee Kunda}
\address{Vanderbilt University Department of Computer Science, 400 24th Ave S, Nashville, TN 37212 USA}

\begin{abstract}
As being widely used to measure human intelligence, Raven's Progressive Matrices (RPM) tests also pose a great challenge for AI systems. There is a long line of computational models for solving RPM, starting from 1960s, either to understand the involved cognitive processes or solely for problem-solving purposes. Due to the dramatic paradigm shifts in AI researches, especially the advent of deep learning models in the last decade, the computational studies on RPM have also changed a lot. Therefore, now is a good time to look back at this long line of research. As the title---``a comprehensive introduction''---indicates, this paper provides an all-in-one presentation of computational models for solving RPM, including the history of RPM, intelligence testing theories behind RPM, item design and automatic item generation of RPM-like tasks, a conceptual chronicle of computational models for solving RPM, which reveals the philosophy behind the technology evolution of these models, and suggestions for transferring human intelligence testing and AI testing.
\end{abstract}


\end{frontmatter}

\section{Introduction}

Most AI researcher, if not all, must have ruminated on fateful questions, which are disturbing but cannot be answered yet, such as ``how far are we on the way to achieve the human-level AI?'' and ``how long does it take for us to fully understand the fundamental mechanism of intelligence?'' Some are more pessimistic, like ``will human-level AI be realized in my lifetime?'' Though these questions cannot be answered for now, every AI researcher is glad to see these questions being raised and attempts being made to answer them, because, whether optimistic or pessimistic, these questions represent the conscience of AI research. 

Works to answer these questions are mainly centered around comparing AI systems and humans on daily tasks that are considered indicators of intelligence.
Among these works, the most direct way is to evaluate AI systems on human intelligence tests. The scope of intelligence tests is larger than the ability tests used in clinical setting. For example, SAT and MAT can be considered as intelligence tests. In addition, many developers and publishers do not name their tests intelligence tests for some people consider the word ``intelligence'' elitism and racism, and prefer to use more accurate words, like  ``tests of learning abilities'', ``assessment of memory and attention'', and ``development motor scales''.
Intelligence tests are usually classified into two categories---single-format tests and battery-types tests. The single-format test contains items of the same format while the batter-type test contains multiple subtests of different formats. As current AI systems require the problem format to be clearly defined, evaluations of AI systems on intelligence tests are mainly on the single-format tests or a subtest of battery-type tests. 
Raven's progressive matrices (RPM) are a family of single-format tests that have been used to test AI systems in a substantial amount of works. Meanwhile, RPM has also become an impetus for developing more intelligent systems that could solve RPM as well as humans. The length of this research line dates back to 1960s; the width of this research line ranges across multiple disciplinarians, such as AI, cognitive science, neuroscience, psychometrics and so on. However, there has been lacking a work that inspects this research line in a joint view of its entire temporal and spatial span and establishes the theoretical depth of it. Given the recent development of this research line, we believe now it is a good point to do this a work. 

We will start this work by reviewing the basics of RPM in the context of human intelligence testing in Section~\ref{sec:rpm}. The purpose of section is to answer the two theoretical questions that one would first ask about RPM---what RPM measures and how RPM measures it. The answers go well beyond the ones like ``it measures human intelligence'' and ``it asks participants to solve problems''. By answering these two questions, we intend to explain the rationale of using RPM as a human intelligence measure. We believe this is necessary for analyzing the rationale of using RPM as a AI measure, and, more generally, for establishing the theoretical foundation of AI testing. 

We extend the discussion to the entire problem domain represented by RPM in Section~\ref{sec:rpm-like}. This domain includes several more tasks that are similar to RPM and also used for human intelligence testing and AI testing. To distinguish them from original RPM, we refer to them as RPM-like tasks. In these tasks, while items for human intelligence testing are mostly handcrafted by human experts, algorithmically-generated items are more and more useful in some special testing scenarios such as computer-based, adaptive, large-scale and/or repeated testing. Algorithmically generated items are also a realistic incentive for studies of deep learning models for solving RPM-like problems. Thus, in the second half of this section, we also reviewed the important works for algorithmic generation of matrix reasoning items, which exactly replicate the format of original RPM. In this section, we intend to (a) provide our readers with different choices of tasks and problem/data sets for different research purposes, (b) provide practical guidance for building algorithmic item generators, and (c) pave the way for the discussion of learning models in the following sections.  

In Section~\ref{sec:com-mod}, we propose a framework to collate all computational models for solving RPM and RPM-like tasks. We refer to this framework as a conceptual chronicle because it emphasizes the conceptual connections between computational models and the underlying logic for technological development. It is neither like the reviews that use specific taxonomies of reviewed works nor the ones that compile the reviewed works into a chronological order. Instead, it simulates the process of how a beginner's understanding of this field would naturally evolve as she knows more and more about this field. In a sense, it is more like chapter organizations of textbooks. We believe such a presentation is the best way for readers to gain a coherent understanding of this field. 

In Section~\ref{sec:dis}, we zoom away from the computational models and address more general topics of AI testing. We first tackle the fundamental issue in this research field---i.e., the validity of using intelligence tests and similar tests to evaluate AI systems. The discussion is based on the initial idea that AI systems could be measured by these tests as human intelligence is measured by them. Unless this issue is properly resolved, the practice of applying these tests on AI systems would be restricted into pure problem solving for specific problems, rather than deepening our understanding of human intelligence and AI. Secondly, on the flip side, we also discuss the implications of human intelligence manifested on intelligence tests for building AI systems. The generalization ability and robustness of human intelligence on intelligence tests are far better than what current AI systems could achieve. We believe such a discussion is crucial for future works in this research field.

\section{Raven's Progressive Matrices}
\label{sec:rpm}

\begin{figure}[t]
    \centering
    \begin{subfigure}{0.24\textwidth}
        \centering
        \includegraphics[width=0.97\textwidth]{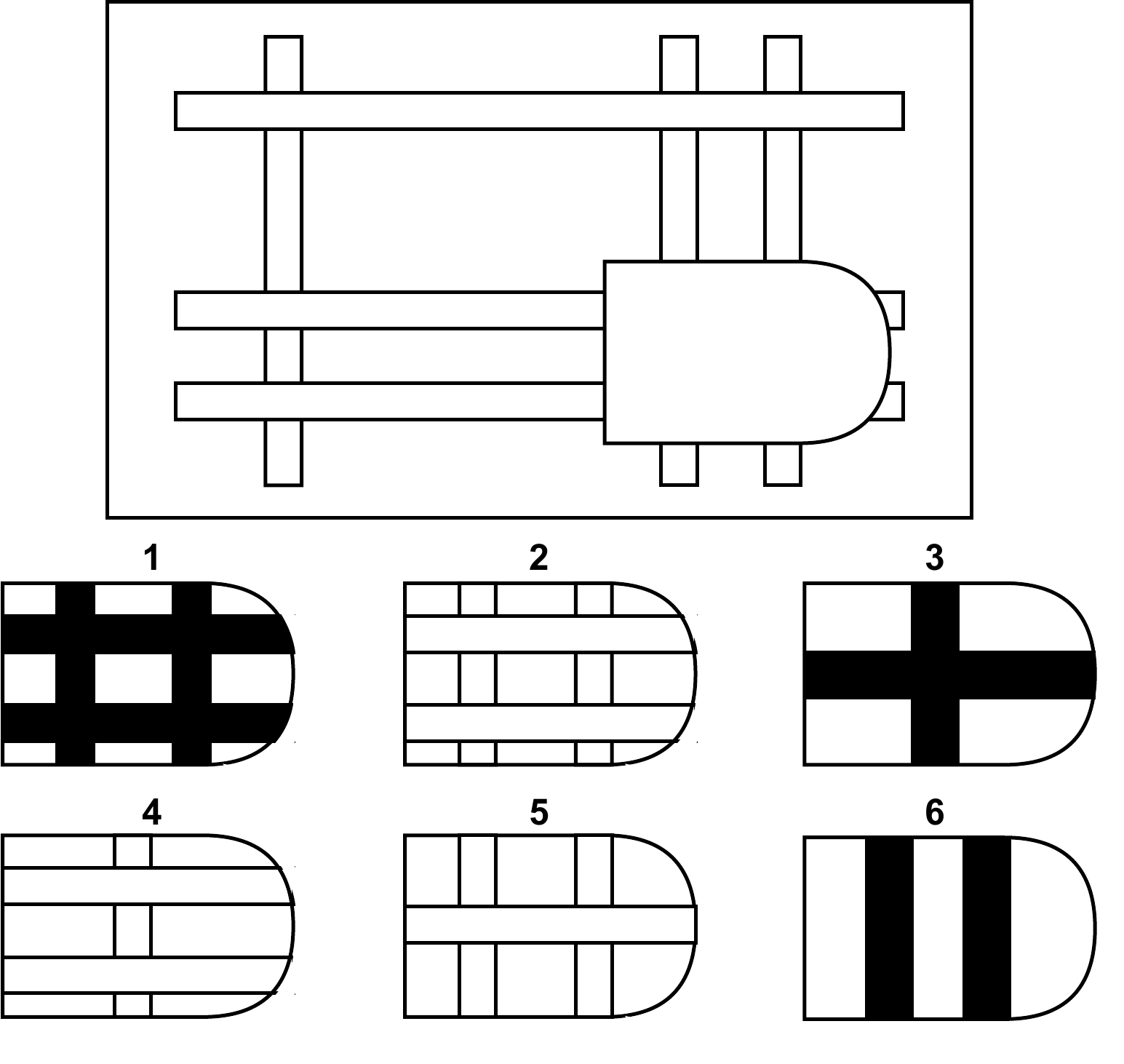} 
        \caption{}
        \label{fig:rpm22-p}
    \end{subfigure}
    \begin{subfigure}{0.24\textwidth}
        \centering
        \includegraphics[width=\textwidth]{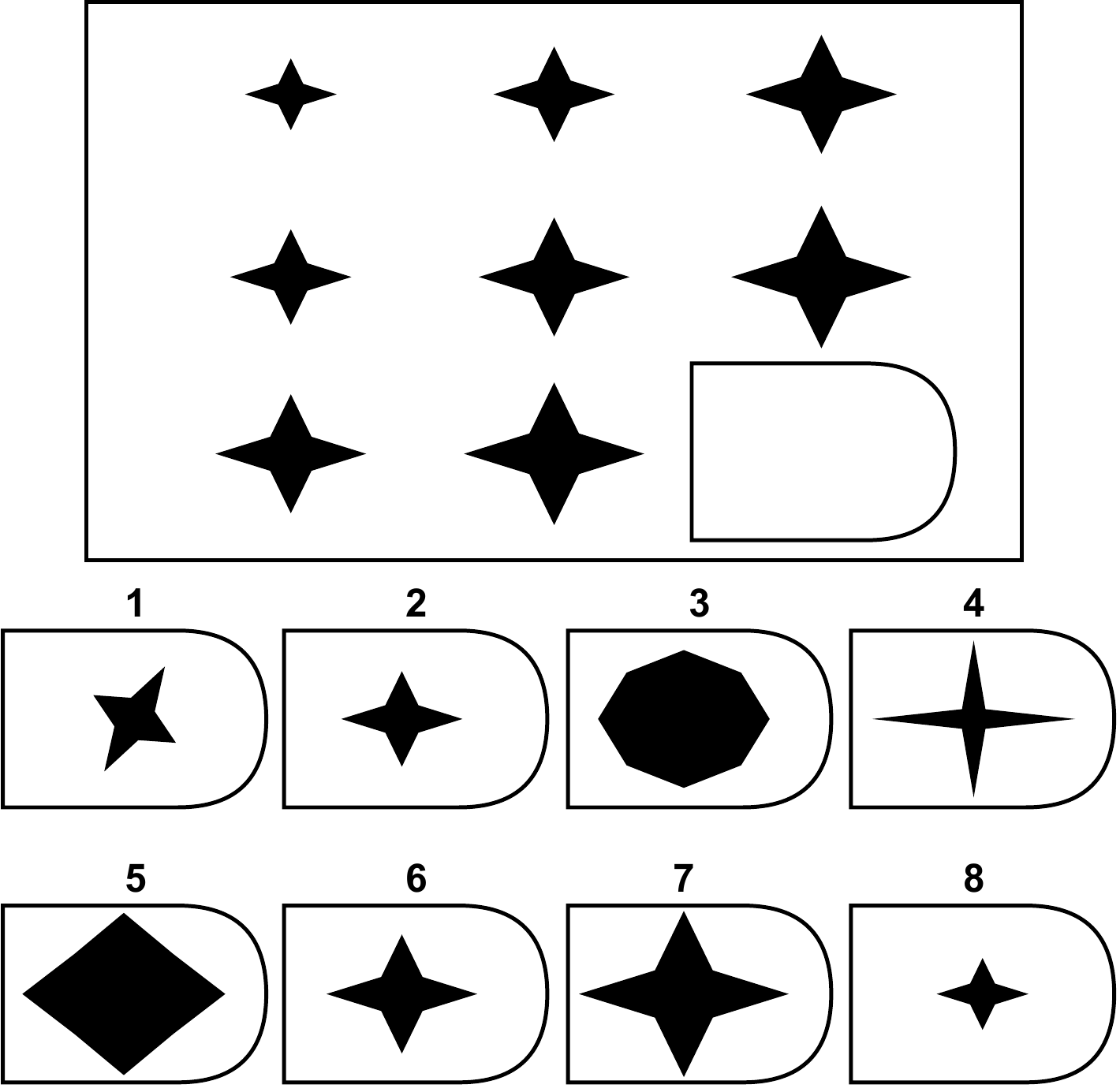}
        \caption{}
        \label{fig:rpm33-p}
    \end{subfigure}
    \begin{subfigure}{0.24\textwidth}
        \centering
        \includegraphics[width=\textwidth]{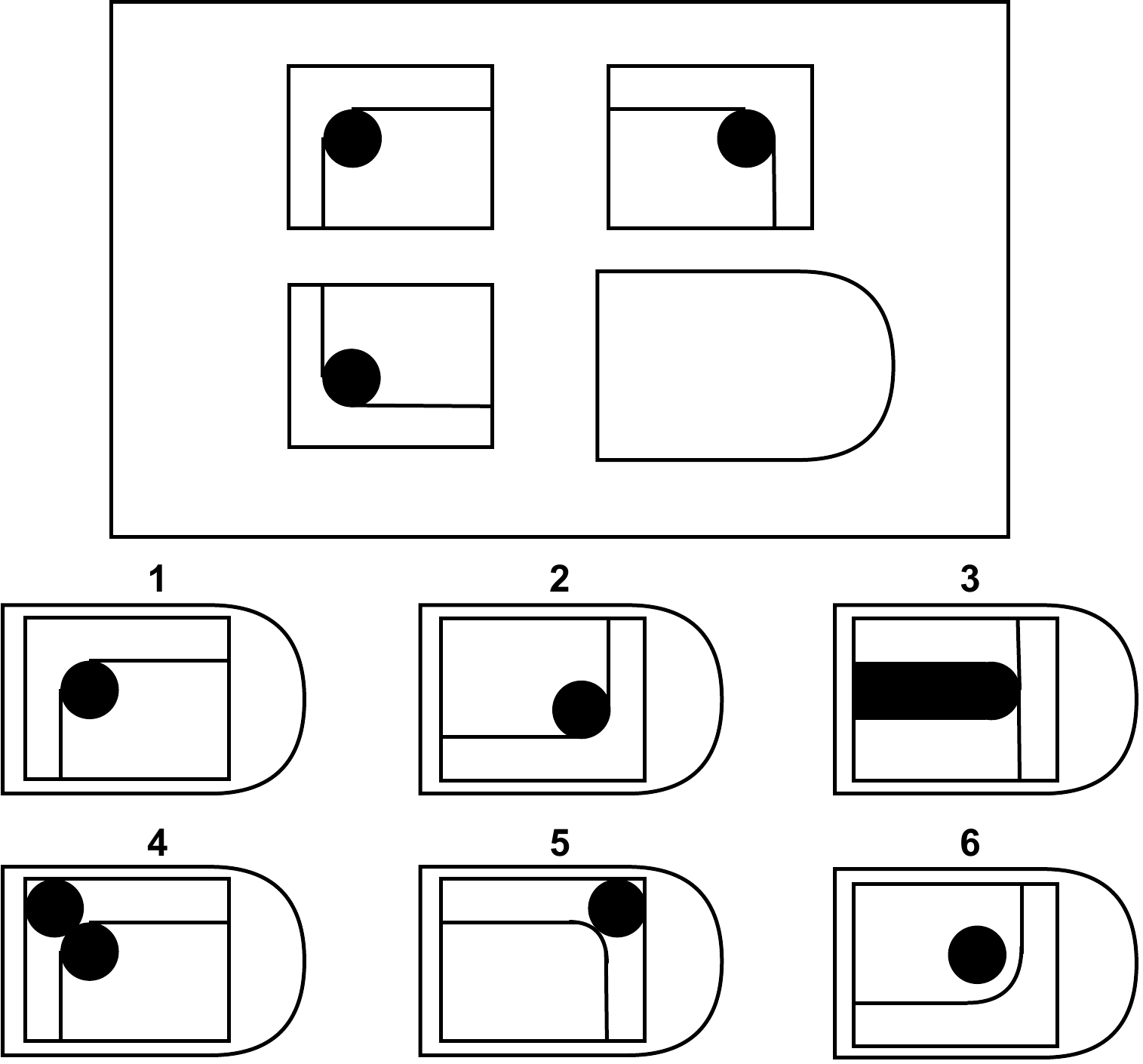}
        \caption{}
        \label{fig:rpm22-l}
    \end{subfigure}
    \begin{subfigure}{0.24\textwidth}
        \centering
        \includegraphics[width=\textwidth]{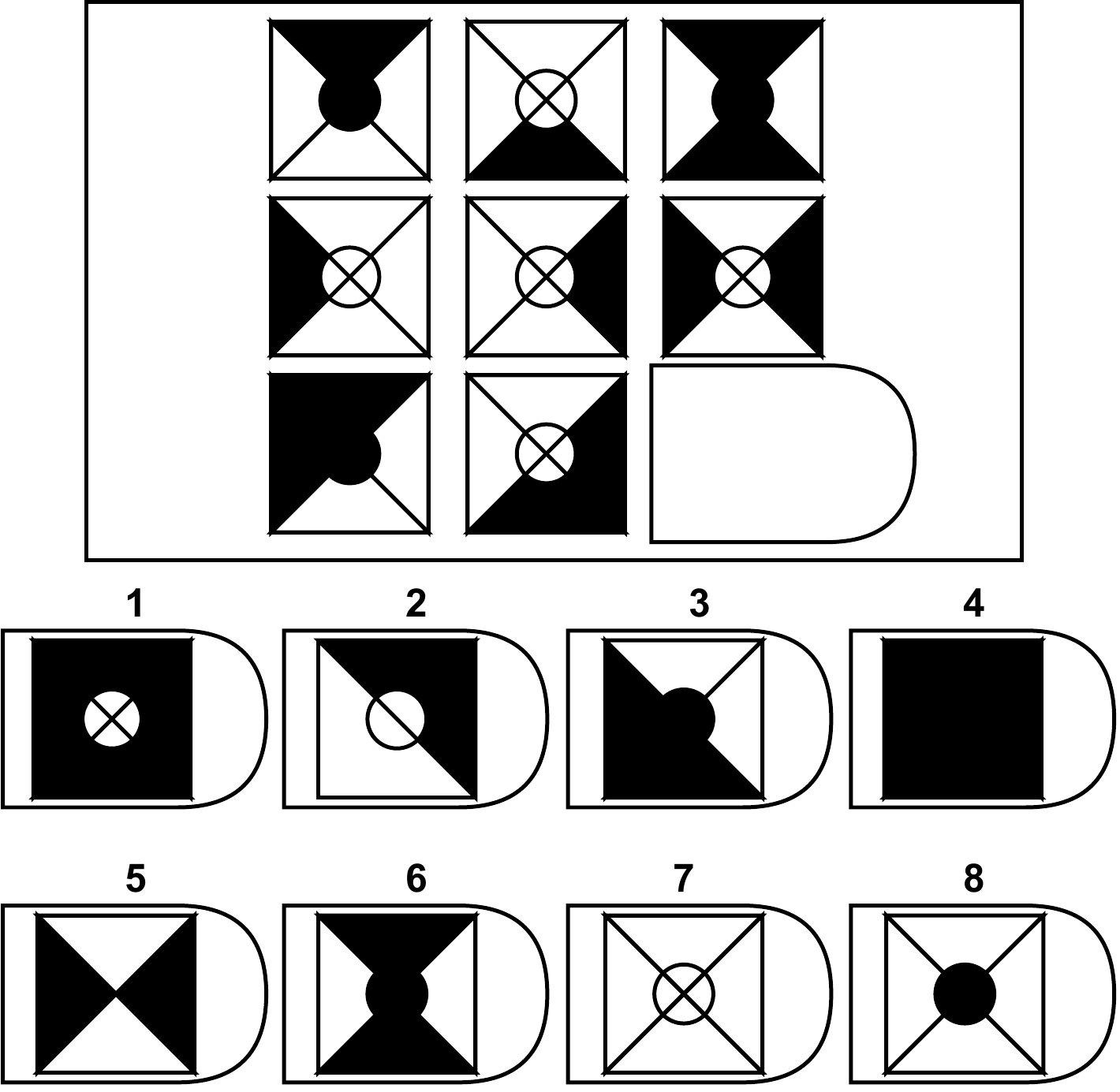}
        \caption{}
        \label{fig:rpm33-l}
    \end{subfigure}
\caption{RPM examples of different formats and stimuli.}
\label{fig:rpm-examples}
\end{figure}

For readers who are not familiar with RPM, Figure~\ref{fig:rpm-examples} shows some examples of RPM items. The original RPM tests contain items of four formats as shown in Figure~\ref{fig:rpm-examples}. The items are presented as multi-choice problems. The context can be a single image with one piece missing (Figure~\ref{fig:rpm22-p}), or a 2$\times$2 or 3$\times$3 matrix with the last entry missing (Figure~\ref{fig:rpm22-l}, \ref{fig:rpm33-p} and \ref{fig:rpm33-l}). To solve an RPM item, one needs to select an answer from the answer set to complete the context matrix. In original RPM tests, the answer sets contain 6 choices for single-image and 2$\times$2 matrices and 8 choices for 3$\times$3 images. 

Given different perceptual stimuli that populate the matrix, the item requires different cognitive abilities and skills. For example, the items in Figure~\ref{fig:rpm22-p} and \ref{fig:rpm33-p} tap into cognitive abilities of perceptual processing. Particularly, Figure~\ref{fig:rpm22-p} requires processing perceptual continuity to interpolate the missing piece in (or match the answer choices to) the context image; Figure~\ref{fig:rpm33-p} requires processing perceptual progression to extrapolate the missing image. The other two items in Figure~\ref{fig:rpm22-l} and \ref{fig:rpm33-l} differ from the first two because they requires not only the perceptual processing abilities, for example, perceptual decomposition and organization, but also abstract inductive reasoning, which involves constructing abstract symbols from raw perceptual stimuli and reasoning about these symbols.

Figure~\ref{fig:rpm-examples} represents the most typical designs of original RPM. It needs to be pointed out that RPM-like tasks are not restricted to these designs and that various designs have bee used in the RPM-like task to test different cognitive abilities and verify cognitive theories (more details in Section~\ref{sec:rpm-like}). 

It has been claimed that RPM tests are the best single-format intelligence test that one can have. This claim is based on the statistical evidence that the test scores on RPM are highly correlated with all other common intelligence tests. RPM could be visually considered located at the center on the map of all intelligence tests \citep{snow1984topography}, implying that the underlying trait behind RPM tests is also central to the traits that are measured differently. For this reason, while RPM receives much attention in clinical settings, it also receives a great deal of attention in research settings, especially in the communities of cognitive science and artificial intelligence.

\subsection{What RPM Measures?}
\label{sec:what-rpm-measures}
What RPM measures exactly? This simple question must have been haunting many researchers who are not psychologists or cognitive scientists for the first several years of their research on RPM. Well, the answer to this question may be quite straightforward to some researchers---it measures intelligence. But the others simply do not understand why these ``drop in from the sky'' items can tell about a person's intelligence. This question is probably better to be rephrased as ``why and how does solving these problems composed of simple geometric patterns measure a person's intelligence?''

The answer is not a simple one, given the complex nature of human intelligence testing. First of all, RPM represents a type of intelligence tests that are theory-motivated. That is, the test development is inspired and guided by some abstract theories about intelligence, which involve factors that are not observable. In contrast, our stereotypical impression of tests is the ones that are related to our daily experience and pragmatic purposes. For example, SAT contains sections of writing, verbal comprehension, and mathematics because competence on them is necessary for students to perform well in college and graduate; the Armed Services Vocational Aptitude Battery contains sections of electronics, auto, shop, mechanical comprehension, and assembling objects, because these knowledge and skills are necessary for the technical positions in army. The development of these tests starts off with clear purposes and understanding of what specific behavior should be measured.

However, RPM, as an intelligence test, is to measure intelligence---a factor that is not clearly defined, directly observable, or measurable. Thus, theories have been constructed to explain the relation between intelligence and observable and measurable behavior. When RPM is not introduced to someone without clarifying the theories, she would have the question at the beginning of this subsection. In particular, John C. Raven, the author of RPM  \citep{raven1936mental, raven1941standardization}, had studied with Charles Spearman, who noticed that a person's performances on tests of different cognitive abilities are correlated and thus hypothesized that a factor---general intelligence $g$ \footnote{Spearman referred to $g$ as general cognitive ability because he thought the word intelligence had been abused by many people.}---underlies all cognitive abilities. Spearman further pointed out that the $g$ factor is composed of two abilities --- \textit{eductive ability} and \textit{reproductive ability}. Eductive ability is the ability to make meaning out of confusion and generate high-level, usually nonverbal, schemata which make it easy to handle complexity. Note that the process of ``eduction'' is more often referred to as inductive reasoning. Reproductive ability is the ability to absorb, recall, and reproduce learned information and skills. 

To test eductive and reproductive abilities, Raven developed RPM and Mill Hill Vocabulary Scale, respectively. In contrast to the pragmatic tests, the development of these tests started off with the author's personal understanding of these abilities. But, it is important to point out that the development of theory-motivated tests are not idiosyncratic because the developer needs to prove that the test indeed measures what it is expected to measure. The proof is usually achieved by collecting statistical evidence that the test score is correlated with certain measurable behavior and other tests, which are determined by the purpose of the test and interpretation of test score. For example, if the test is for recruitment, the test score should be correlated with future job performance; if the test is a general intelligence test, the test score should be correlated with cognitive ability tests and medical data such as fMRI data of the brain. In the terminology of psychometrics, the developer needs to validate the test to make sure it measures what it is expected to measure. However, the studies of RPM validity would make a new book. We simply claim that RPM is well-validated test of general intelligence.

Readers might have already noticed that there are two abilities under the umbrella of $g$ and correspondingly two tests. What about the reproductive ability and its test? Why is RPM considered as the best single-format test for general intelligence instead of the other? Is eductive ability more important than reproductive ability? In his theory of general intelligence, Spearman did not treat these two abilities as separate factors. On the contrary, he believed that there is only a single factor---$g$---underlying all cognitive abilities, and eductive and reproductive abilities are two ``analytically distinguishable components'' of $g$ \citep{raven2008raven}. Eductive and reproductive abilities are better better treated as two interwoven general cognitive processes, through either of which $g$ can be measured. Since the test scores of RPM are best correlated to other intelligence tests, RPM is considered the most effective single-format intelligence test.

Now is a good point to compare to another two relevant concepts that pervade the literature of intelligence and our readers are probably more familiar with them. In the theory of general intelligence by Cattell \citep{cattell1941some, cattell1943measurement, cattell1963theory, cattell1987intelligence}, he proposed that there are two general factors (emerging from factorial analysis) subtending intellectual performances---fluid intelligence and crystallized intelligence. Fluid intelligence, $g_f$, is the ability to discriminate and perceive complex relationships when no recourse to answers is already stored in memory. Crystallized intelligence, $g_c$, consists of judgmental, discriminatory reasoning habits long established in a particular field, originally through the operation of fluid intelligence, but no longer requiring insightful perception for their successful operation. The definitions of fluid and crystallized intelligence resembles the ones of eductive and reproductive abilities. Moreover, fluid and crystallized intelligence are frequently used as synonyms of eductive and reproductive abilities in literature. But these two sets of terms are conceptually different. In particular, Spearman considered eductive and reproductive abilities as two components, while Cattell treated fluid and crystallized intelligence as factors. When we say components of a system, we mean that the components must work together for the system to work; if either of eductive and  reproductive component does not work, the whole system does not work. But when we say factors (especially in factorial analysis), we mean different dimensions that each exert separable influence on experimental outcome and thus can be studied separately. We can calculate what percentage of the variation in the data is caused by which factor (using procedures in analysis of variance), but it is conceptual wrong to do so in component systems because the components' influences are not separable. Note that this does not mean that factors are completely independent because two factors can still correlate and jointly contribute to a proportion of variation. A good example is height and weight, which are correlated, but still two different concepts and factors. As factors, their private and and shared contribution to athletic ability can be determined statistically if we collect data of athletes. Therefore, when we are using these two sets of terms interchangeably, we need to be clear about which theoretical assumption about them and have different conclusions from the the experiment if necessary.

Beside the conceptual issue behind terminology, another issue is that the boundary between theory-motivated and pragmatic tests is not so clear in practice. As more and more researches are conducted on a pragmatic test, theories will be invented to explain human responses on the tests. Similarly, as a theory-motivated test is proven to be a valid measure for some mental trait, it is also possible to use it for pragmatic purposes. For example, RPM was once used for military recruitment in UK during World War II \citep{burke1958raven}.

\subsection{A Brief History of RPM}

This paper would be incomplete if we did not say something about the history of RPM, which is almost 100 years long. Admittedly, not every detail of this history is relevant to our research of RPM in the context of AI. However, the development of RPM in human intelligence testing would provide potential enlightenment for the future of AI testing, which is largely undefined yet. We will introduce the whole family of RPM in this subsection\footnote{This subsection is mainly based on the manuals of RPM tests. For readability, we will not insert citations of the manuals in this subsection. Otherwise, it would be everywhere.}, and discuss the the motivation behind each RPM test and the connection between them. 

\citet{raven1936mental} developed the first RPM test in 1930s when he was studying with Lionel Penrose, who was a geneticist and psychiatrist. This test was used to study the genetic and environmental determinants of intellectual defect. As other genetic studies, this study required a large population of subjects, including adult parents and children at all ages, being tested at different places, such as home, school, and workplace. It is, therefore, infeasible to administer full-length intelligence tests, such as Binet tests and Wechsler tests, which require hours for a session. In addition, because some subjects then were illiterate and many workplaces were too noisy for verbal questions, the item had to be nonverbal and as self-evident as possible. These practical requirements together led to the design of the first RPM test.  

As we have mentioned, the development of RPM was theoretically inspired by the Spearman's theory of intelligence. Although the theory is instructive for understanding intelligence, the overarching $g$ factor is a latent variable, which is not directly observable and measurable. 
This makes its measurement inherently complicated because one needs to identify the measurable activities and decide how they relate to the latent variable, for example, it can be calculated by weighting scores on multiple cognitive ability tests. To simply its measurement, Raven mentioned in his personal notes that he intended to develop ``a series of overlapping homogeneous problems whose solutions required different abilities'' \citep{carpenter1990one}. In particular, these items are homogeneous in the types of perceptual stimulus and abstract relations, but their difficulty varies in a wide range. If these homogeneous items are arranged evenly in an increasing order of difficulty, they together will form a ruler of intelligence. That is, a subject is less likely to be able to solve an item if she cannot solve the items before it. As the test is administered to more and more people and more data are collected, the item difficulty is determined more accurately (relative to people's ability to solve it; through psychometric procedures). Now, the outcome of this multi-ability, homogeneous, and increasing-difficulty design is that we can measure the latent variable $g$ with a single single-format test. Intuitively, the RPM tests make the $g$ factor directly measurable and the scores more interpretable as we use a tape measure to measure height and a thermometer to measure temperature.

RPM is a family of progressive matrices tests, including three main tests---Standard Progressive Matrices (SPM), Coloured Progressive Matrices (COM), and Advanced Progressive Matrices (APM), and each test has multiple versions consisting of different items. The first RPM test is the SPM test published in 1938 \citep{raven1941standardization}, which is the ancestor of all the following RPM tests. Including the first version of SPM, all the SPM tests are composed of 60 items, which are organized into 5 set (A, B, C, D, and E) according to their difficulty. The item difficulty increases within each set and from Set A through Set E. Meanwhile, each set has a distinct theme manifested by the perceptual stimuli and conceptual relations of items in this set. 

To spread the scores and have a better precision at the lower and upper ends of the ability range, the first versions of CPM and APM were developed and published in 1947. CPM reused the Set A and B of 1938 SPM and placed a transitional set of 12 items---Set Ab---between Set A and B. The items in this set were constructed to be intermediary in difficulty between Set A and Set B. Thus, CPM has had 36 items organized into three sets. As the name indicated, CPM is printed in color to appear interesting as it is often administered to children under 11. CPM can also be administered to mentally retarded persons, the elderly, and people with brain injury. Different from SPM and APM, CPM was published in two forms --- the book form (i.e., a the paper-and-pencil test) and the board form. In the board from, each item is a board with a part removed and movable pieces as answer choices to complete the board. The board form has been proved to be equivalent to the book form, tapping the same cognitive process. Moreover, the board form has two advantages. First, the board form can be better administered without verbal instruction because the administrator can demonstrate the expected response by manipulating the board and answer pieces. This is important for people who are deaf people or unable to communicate for some reasons. 

The APM was originally drafted in 1943 for use by the British War Office Selection Boards, who needed a more difficult RPM test that can provide better discrimination at higher levels than SPM. The APM test was published in 1947, consisting of two sets --- Set I and Set II. Set I comprises 12 items covering all themes and sampled on the full test of SPM. In practice, Set I can be used to familiarize people with the test, sorting people into the ``dull'' 10\%, ``average'' 80\%, and ``bright'' 10\%, and decide whether SPM or Set II should be used next. The 1947 Set II consisted of 48 items, which resembled the items in Set C, D, and E of SPM in presentation and argument. In 1962, 12 items making no contribution to the score distribution were dropped from Set II and the remaining 36 item re-arranged. 

In the last decades, there has been a significant and steady increase in many intelligence test scores, including the SPM scores. Among all RPM tests, SPM is designed to cover the widest ability range. But this increase causes SPM to be less discriminative at upper levels of ability range (i.e., ceiling effect). In 1998, a new SPM test---SPM plus---was published to restore its discriminative power at the upper levels, and, meanwhile, keep its discriminative power at the lower levels unchanged. In particular, SPM plus includes all the items in Set A and B of SPM and replaces moderately difficult items with more difficult items in Set C, D, and E. 

As a result of its simple self-evident format, insensitivity to culture and language, and centrality among all intelligence tests, RPM has been the most widely studied single-format intelligence test and has large amounts of testing data available for research. This, however, raises a concern that the test has become too well known and the participants could be coached for solving them or memorizing the answers. This is problematic when important decisions (such as educational opportunity and job recruitment) are made upon the test results. Therefore, parallel versions of CPM and SPM was developed in 1998. These versions are designed to be parallel to the classic SPM on an item-to-item and overall score basis so that the existing data of classic SPM and CPM could be used to analyze the data of the parallel versions. 

The administration procedure of RPM tests is relatively flexible compared to other intelligence tests. RPM tests can be administered both individually and in groups. In individual test, one administrator guides one participant through the test. In group test, one administrator proctors the participants as in normal school exams. Individual tests introduce emotional factors which are not present in group testing or self-administration, and thus the scores are slightly lower than group tests, in which participant work on their own. But individual tests allow the administrator to make sure the participant understands what to do and observe the participant to collect more data, such as whether the participant uses a trial-error strategy. Thus, individual test is recommended when important decisions are made upon the test result. In both group and individual tests, instructions can be given verbally or using gestures such as pointing, nodding, and shaking head. In most cases, RPM tests are given in an untimed manner or with sufficient time to attempt every item since, when timed, the validity of scores is reduced according to statistical evidence. Moreover, it has been argued that RPM is neither a speed test nor a power test, or a combination of both. There is an exception that, after familiarization with Set I of APM, Set II of it was administered with a time limit to measure the speed of intellectual work.

To sum up, RPM is a big family of tests, including SPM, parallel SPM , SPM plus, CPM (with two forms), parallel CPM, and APM. All the RPM tests that are used today have gone through many revisions as more and more data are collected in different countries and from different groups of people. There also exist different procedures to administer the tests, which result in qualitative different results. When studying RPM in the context of artificial intelligence, it is important to point out which RPM test is used and how it is used in terms of the administration procedures.

\subsection{What RPM measures exactly?}
\label{sec:what-rpm-meansures-exactly}

At the beginning of this section, we have tried to answer the question ``what RPM measures'' from a theoretical perspective. In short, RPM measures eductive ability, which is a component of general intelligence (i.e, the $g$ factor or genera cognitive ability), and thus can be used as an index of general intelligence. However, the answer is still too abstract and does not land on the concrete items in RPM tests. To be honest, the answer at the beginning could apply to almost every test of eductive ability, fluid intelligence, or general intelligence. To tell the whole story of RPM, we further reify the answer by inspecting the concrete items and the administration procedures.

\begin{figure}[t]
    \centering
    \begin{subfigure}{0.24\textwidth}
        \centering
        \includegraphics[width=0.97\textwidth]{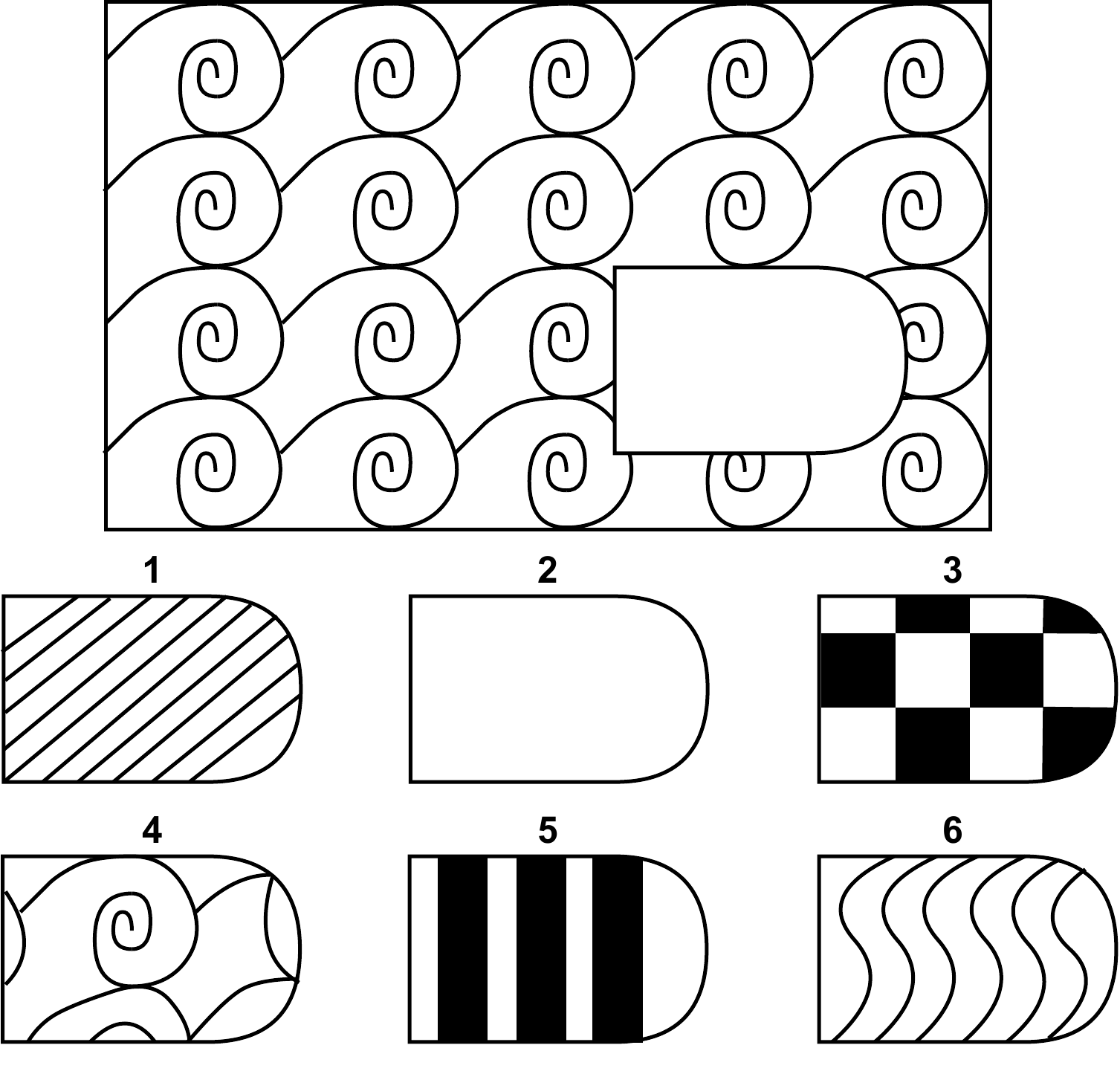} 
        \caption{}
        \label{fig:spm-1}
    \end{subfigure}
        \begin{subfigure}{0.24\textwidth}
        \centering
        \includegraphics[width=\textwidth]{rpm-perceptual-relation.pdf}
        \caption{}
        \label{fig:spm-2}
    \end{subfigure}
    \begin{subfigure}{0.24\textwidth}
        \centering
        \includegraphics[width=\textwidth]{rpm22-perceptual-relation.pdf}
        \caption{}
        \label{fig:spm-3}
    \end{subfigure}
    \begin{subfigure}{0.24\textwidth}
        \centering
        \includegraphics[width=\textwidth]{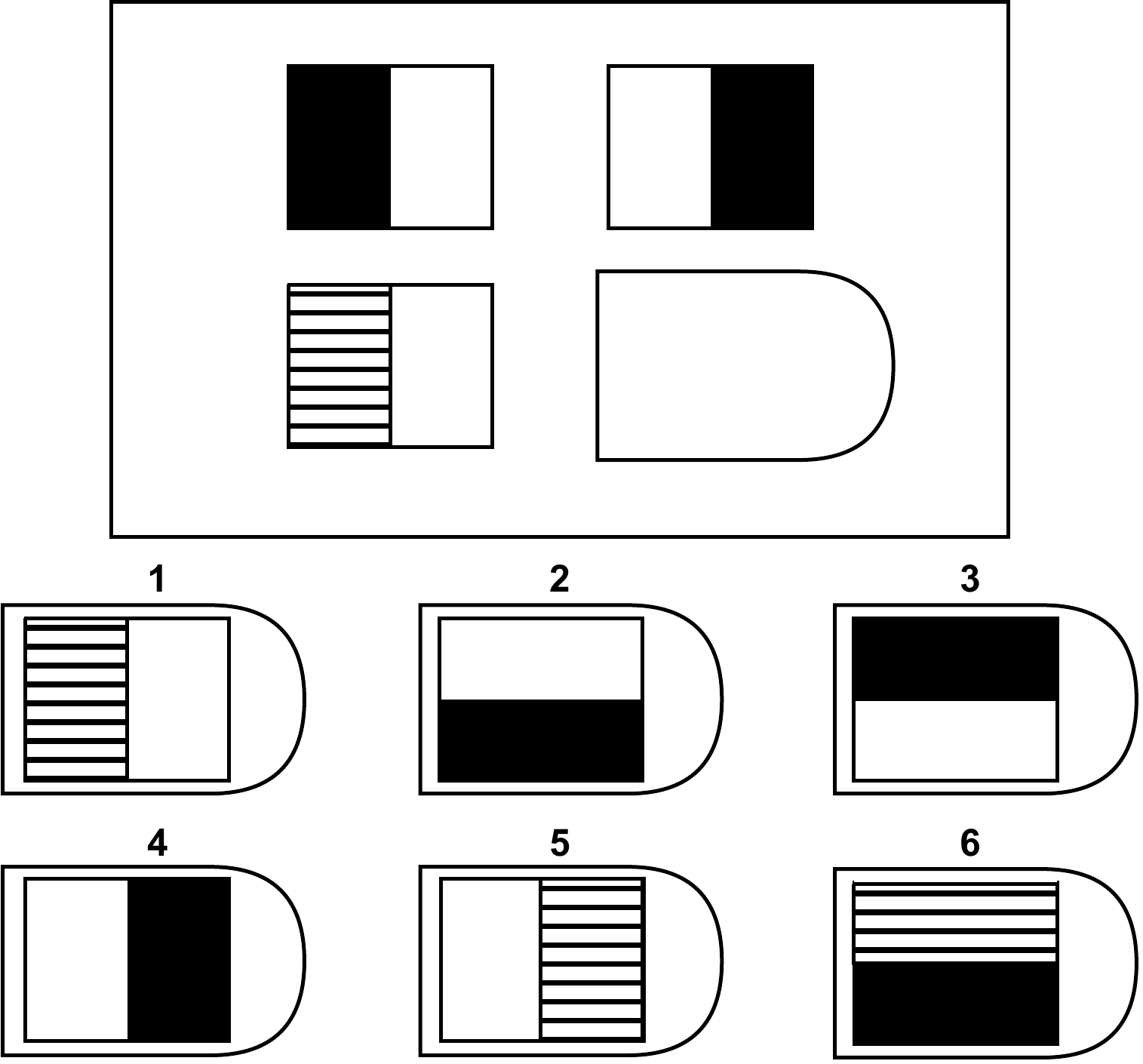}
        \caption{}
        \label{fig:spm-4}
    \end{subfigure}
    
    \begin{subfigure}{0.24\textwidth}
        \centering
        \includegraphics[width=\textwidth]{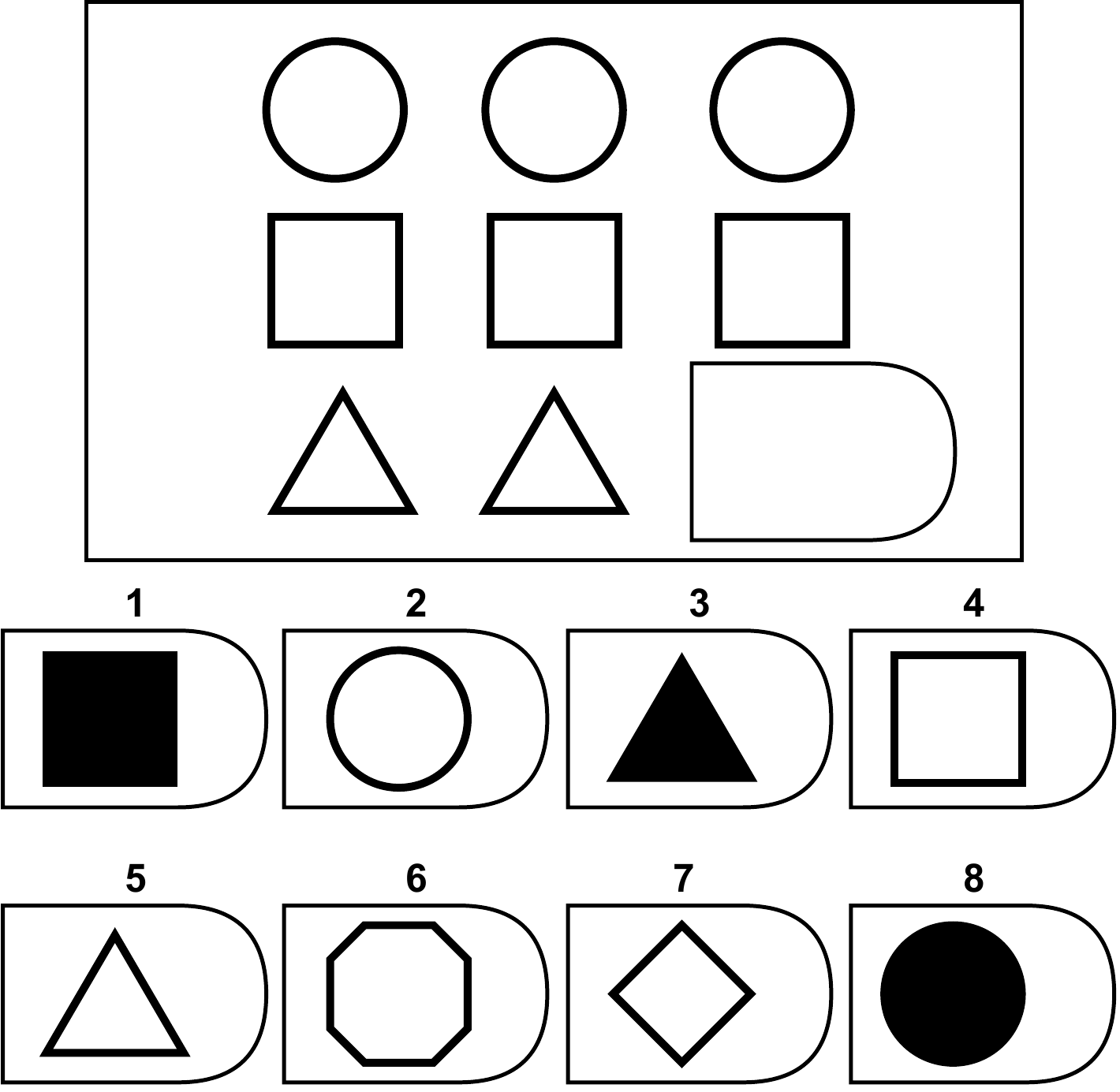}
        \caption{}
        \label{fig:spm-5}
    \end{subfigure}
    \begin{subfigure}{0.24\textwidth}
        \centering
        \includegraphics[width=\textwidth]{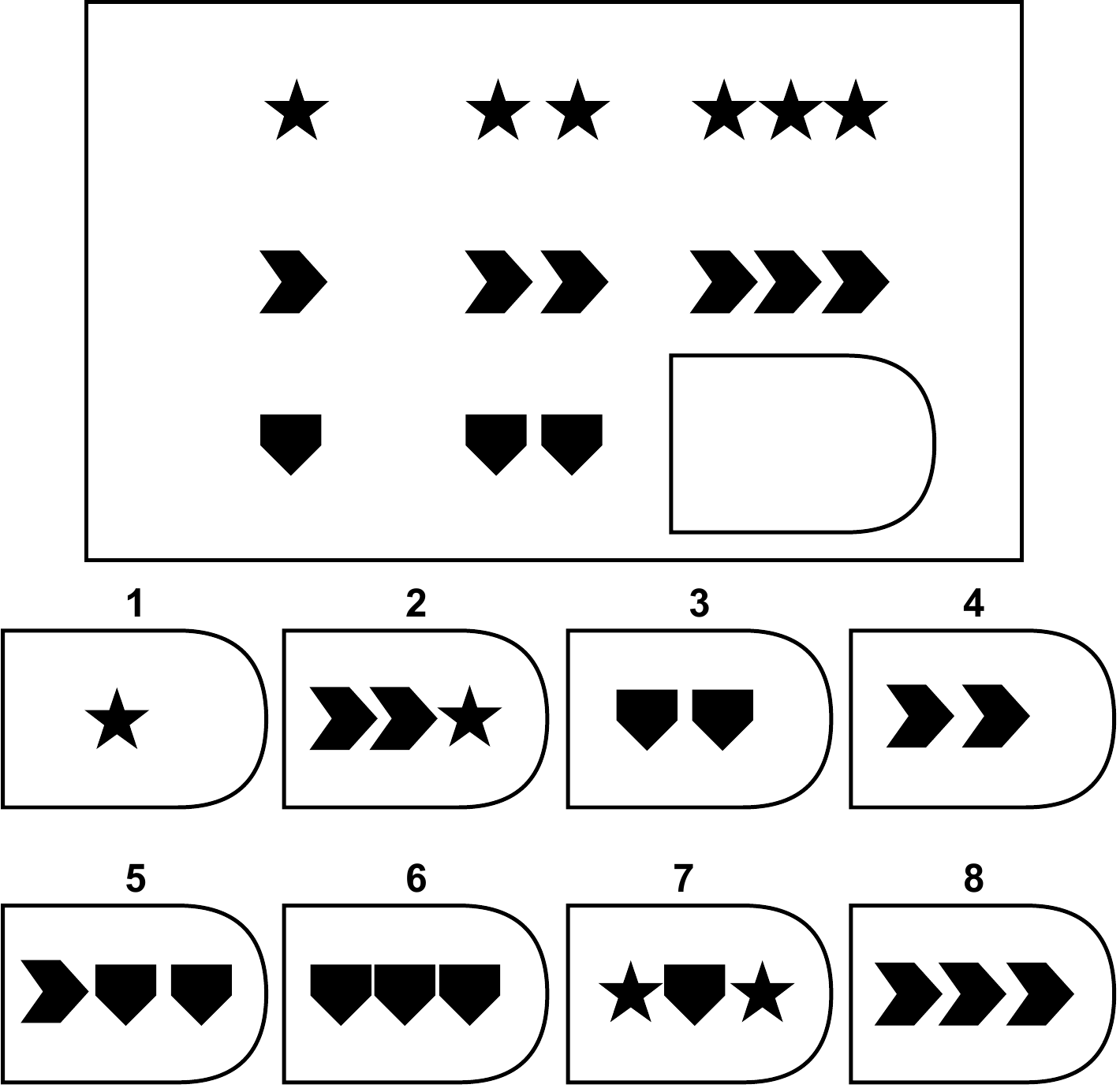}
        \caption{}
        \label{fig:spm-6}
    \end{subfigure}
    \begin{subfigure}{0.24\textwidth}
        \centering
        \includegraphics[width=\textwidth]{rpm33-logical.pdf}
        \caption{}
        \label{fig:spm-7}
    \end{subfigure}
    \begin{subfigure}{0.24\textwidth}
        \centering
        \includegraphics[width=\textwidth]{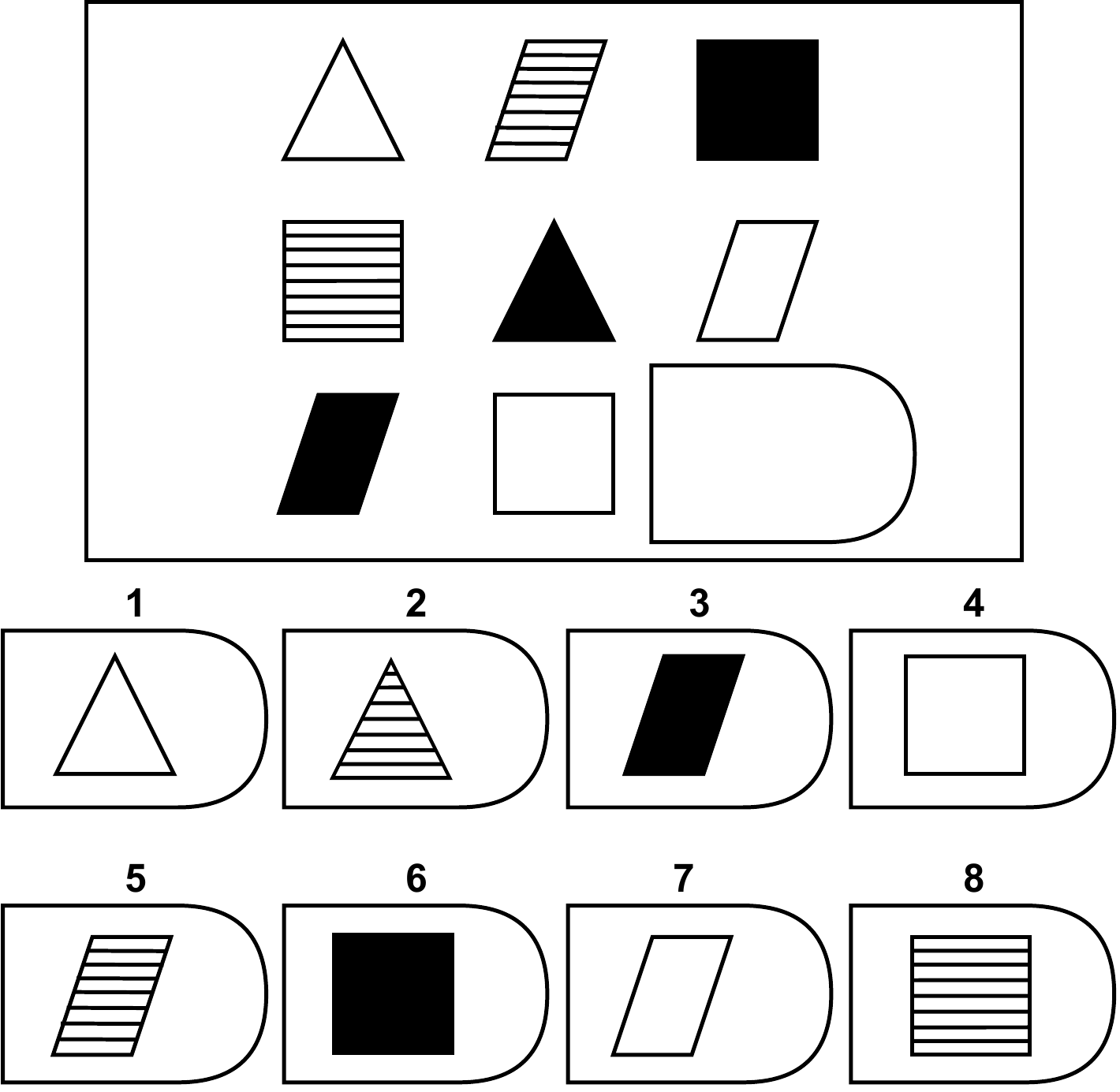}
        \caption{}
        \label{fig:spm-8}
    \end{subfigure}
\caption{Example SPM item series.}
\label{fig:spm-examples}
\end{figure}

We have indicated in previous section that the test design is the outcome of an iterative process, in which the revised tests are repeatedly administered to people so that data can be collected to further revise the test.Since RPM is also a theory-motivated test, the test design is also determined by by the theory of intelligence and how it is implemented in the test. We take SPM as an example. To protect the secrecy of RPM tests, we created several new items (Figure~\ref{fig:spm-examples}) that simulate the item series in SPM. As mentioned, there are five sets in SPM (Set A, B, C, D, and E). The eight items in Figure~\ref{fig:spm-examples} simulate the way how the item design varies from the first item of Set A to the last item of Set E. At the beginning of Set A, a participant will see an item similar to the one Figure~\ref{fig:spm-1}. The role of this item is to give the very basic idea of the test. This item is a good starting point in that no prior knowledge is required to solve the item and its solution is self-evident to almost every participant. In the standard administration procedure, this item is used for teaching trial. The administrator explicitly tells (possibly in a nonverbal way) the participant that ``only one of answer choices can complete the pattern correctly'' and which one it is correct for this item. 

Note that in every administration procedure in the manual of RPM tests (individual or group, verbal or nonverbal), the administrator only tells the participant which answer choice is correct, but never explains why it is correct or the thinking process to solve it. This point is extremely important for the testing to be valid. The teaching trials are to help the participant with the format of the test, i.e., one needs to select an answer to complete the pattern, but not the content of the text, i.e., what pattern it is and how it is completed. The content part is just what the test measures---eductive ability. An even stronger but similar argument \citep{raven2008raven} is that it is not correct to describe RPM items as ``problems to solve''. The instruction that an answer has to be selected does not means that it is a problem. Instead, only when the participant has made some meaning out of the item can the participant sees the item as a problem to solve. The meaning-making part is the core of RPM items, which measures the eductive ability.

After the items for teaching trials, the participant will see an item similar to the one in Figure~\ref{fig:spm-2}. This item takes an important transitional role that shifts the participant's attention from the test format to the test content. In particular, this item explicitly exhibits the nature of the test content---relational reasoning. That is, to solve the following items, the participant needs to consider the relations between the objects rather than, for example, repeating the raw perceptual input in the teaching trials. In addition, the transitional role also lies in the appearance of the items: the teaching-trial items and the transitional items are not presented as matrices, but the transitional items are one step closer to the matrix structure in the following items (see Figure~\ref{fig:spm-3} through \ref{fig:spm-8}), because the relations in transitional items happens in both the horizontal and vertical directions. These transitional items are necessary because they assure that the participant give valid responses to the following items based on the understanding accumulated in the previous items. 

After the transitional items, the test enters 2$\times$2 items like the one in Figure~\ref{fig:spm-3} and \ref{fig:spm-4}, in which, geometric objects are separated into the disconnected matrix entries. These 2$\times$2 matrices start with the ones that more rely on low-level perceptual processing (Figure~\ref{fig:spm-3}) and are relatively easy. After the participant is familiar with the format of 2$\times$2 matrix, it and gradually move on to the ones that involves more abstract relations (Figure~\ref{fig:spm-4}) and are thus more difficult than the perceptual items.

The four items in Figure~\ref{fig:spm-1} through \ref{fig:spm-4} represent the test design in the first two sets of SPM. The following three sets follow the same logic---each item is like a rung of a ladder that makes it possible for the participant to step on the next rung, and the maximum height the participant can reach depends on her strength for climbing the ladder. As a real ladder rung, an item cannot be two far from the previous one. For example, the participant will find an item similar to the one in Figure~\ref{fig:spm-5}, which is used to introduce the 3$\times$3 structure. This item only differs from some items in Set A and B in the matrix size but underlying perceptual processing remains the same. After the participant gets familiar with the 3$\times$3 structure, SPM moves on, as in the Set A and B, from perceptual items to the items that involve more abstract relational concepts, such as number (Figure~\ref{fig:spm-6}), binary logical operation (Figure~\ref{fig:spm-7}), and ternary permutation (Figure~\ref{fig:spm-8}). 
Moreover, the number of relations in item also gradually increases in the last three sets of SPM. For example, the items in Figure~\ref{fig:spm-5}, \ref{fig:spm-6}, and \ref{fig:spm-7} each contain only one relation; the item in Figure~\ref{fig:spm-8} contains two relations---permutation of object shape and permutation of filling texture. 

The example series in Figure~\ref{fig:spm-examples} epitomizes the design of SPM. Through this example, we can see the motivation behind the test design is to provide an ability ladder for the participant to climb. The rungs/items are distributed evenly so that the ladder is climbable. Furthermore, the ladder is climbable to participants for people at every ability level since it starts from the ``ground''---i.e., the beginning trivial items requiring no prior knowledge---and guides the participant to move in the expected direction through conceptually connected items. Once the ``field of thought'' is established, how far the participant can go depends on her ability in this field. 

In a sense, SPM is different from problem-solving tests that everyone has taken at school. Instead, SPM is a miniature that simulates a collection of all tests from elementary level to college level because one need to graduate from every level sequentially. Although the duration for these two types of testing is vastly different, both of them measure the learning potential of the participant. Note that the word ``potential'' here is more suitable than ``ability'' because the ``potential'' means a latent quality that develops under the influence of environmental factors. Since the environment factors can be better controlled in intelligence tests than in the education system, SPM is probably a better measure of learning potential. Moreover, potential is more than ability since the desire to learn knowledge and the courage to conquer new problems are also part of potential. 

In general, RPM is much more than problem solving. Even the word ``test'' is misleading because of our stereotypical impression of test. RPM tests are a system for evaluating eductive ability through measuring the learning potential. However, the common practice of using RPM or RPM-like tests as purely problem-solving tests and making extravagant claim about corresponding abilities of AI systems in many AI studies have been a big misuse of these tests.



\section{RPM-Like Tasks}
\label{sec:rpm-like}

In this section, we extend our discussion to the entire problem domain represented by RPM, which includes RPM-like tasks that inherited the basic elements of original RPM tests and implemented them in more enriched manners. Such RPM-like items can be found in almost every modern intelligence test. In contrast to the theoretical analysis in the last section, we take a more pragmatic approach in this section to describe these tasks. In particular, We surveyed four intelligence tests
\footnote{There are many more important intelligence tests, such as Kaufman, Stanford-Binet, and Wookcock-Johnson tests. But because of the limited access to these commercial tests and resemblance among their RPM-like items, we surveyed only four of them.}
that are widely used in clinical setting and/or frequently related to RPM in literature---Cattell's Culture Fair Intelligence Test (CFIT), Cognitive Assessment System-Second Edition (CAS2), Wechsler Adult Intelligence Test-Fourth Edition (WAIS-IV), and Leiter International Performance Scale-Revised (Leiter-R). Through this survey, we summarized five tasks in the problem domain---matrix reasoning, figure series, analogy making, contrastive classification, and open classification.  In addition, We further survey the methods for algorithmically generating matrix reasoning items, which are a prerequisite for the discussion in the following sections of data-driven AI models for solving RPM-like tasks. As we have mentioned, the items in intelligence tests are mostly handcrafted and thus in a very limited number, which is far below the need of current data-driven models. This section could provide options of existing RPM-like items and suggestions of algorithmically creating new RPM-like datasets for different research purposed.

\subsection{RPM-like Tasks in Intelligence Tests}


Although the theories of intelligence behind the four intelligence tests are different, the RPM-like tasks in these tests are consistent to some degree in terms of what is measured. For example,
\begin{itemize}
    \item the RPM-like tasks in CFIT measures the general cognitive ability, i.e., the $g$ factor, and stresses that the $g$ factor ``reaches its purest expression, i.e., high $g$ loading, whenever complex relationships have to be perceived'' \citep{cattell1950handbook};
    \item the RPM-like tasks in CAS2 measures the simultaneous processing ability in the PASS theory of intelligence \citep{das1994assessment}, i.e., the ability to ``integrates stimuli into (conceptually) interrelated groups or a whole'' \citep{naglieri2014cognitive};
    \item the RPM-like tasks in WAIS-IV ``involves fluid intelligence, broad visual intelligence, classification and spatial ability, knowledge of part-whole relationships, simultaneous processing, and perceptual organization'' \citep{wechsler2008technical};
    \item the RPM-like tasks in Leiter-R measure ``fluid reasoning, deductive and inductive reasoning, and the ability to perceive fragments as a whole, generate rules out of partial information, perceive sequential patterns, and form new concepts'' \citep{leiter1997manual}.
\end{itemize}
From these descriptions of RPM-like tasks in these tests, we can see that they all more or less involve measuring eductive ability or fluid intelligence. Given this internal connection between RPM-like tasks, it would be unsurprising to see common elements shared between them. For perceptual elements, to distinguish eductive ability (or fluid intelligence) with reproductive ability (or crystallized intelligence), they must not be unique to certain cultural groups. There are not too many choices satisfying this requirement, for example, elements from nature like sun and moon, human body (hand and foot), and common shapes and colors. Similarly, common conceptual elements, such as symmetry, topological relations, and number concepts, are also frequently used to create RPM-like items. Now, it is already very hard for test developers to design novel elements for RPM-like items because most of the appropriate elements have already been used in intelligence tests. If one comes up with novel perceptual and conceptual elements that can be used in RPM-like tasks, it will be a great contribution to intelligence test development. Exploration for proper perceptual and conceptual elements for RPM-like tasks is also helpful for building and evaluating AI systems working in this problem domain. 

In addition to perceptual and conceptual elements, there are different formats to present these elements. According to these formats, we classify the RPM-like tasks in the four intelligence tests surveyed into five groups---matrix reasoning, figure series, analogy making, contrastive classification, and open classification. These formats are equally interesting to the perceptual and conceptual elements, as each format is a delicate way to present the same set of elements so that they can be instantly perceived as a problem to be solve but not a trivial one. 


\subsubsection{Matrix Reasoning}

\begin{figure}[!htbp]
    \centering
    \includegraphics[width=\textwidth]{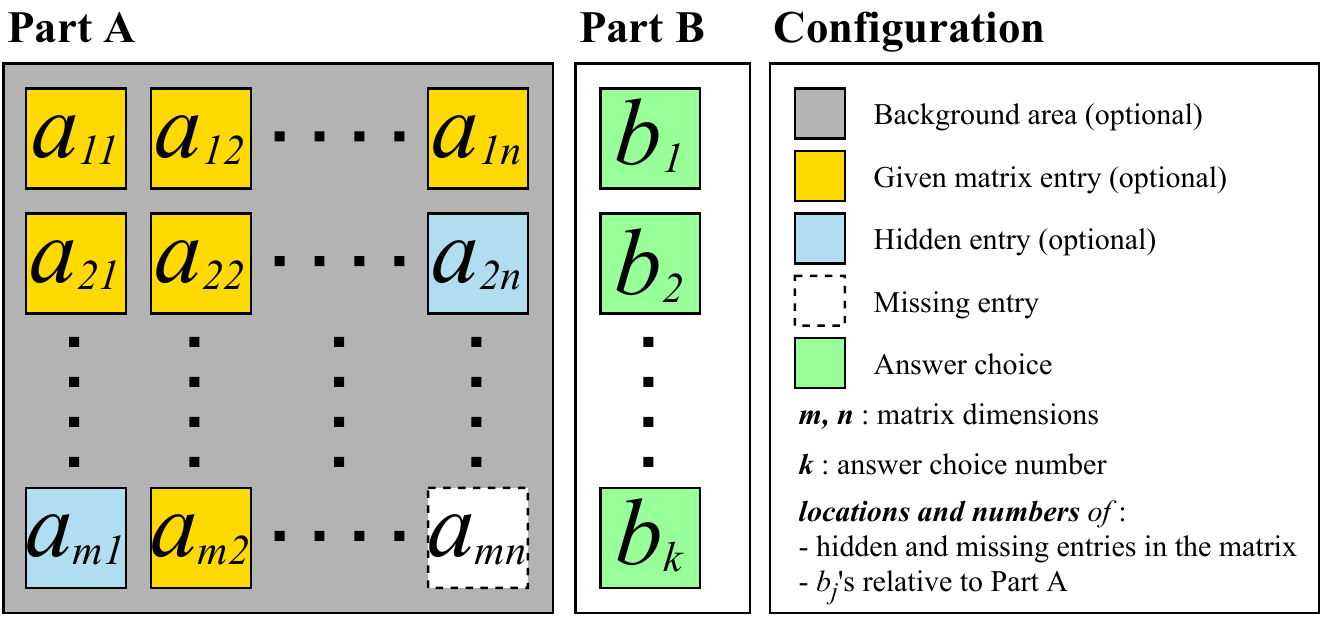}
    \caption{A diagrammatic summary of Matrix Reasoning Task}
    \label{fig:rpm-like-matrix-reasoning}
\end{figure}

Since the the four tests are battery-type tests, they all have multiple subtests, including the RPM-like subtests. Therefore, to keep the whole test in a reasonable length, the RPM-like subtests are briefer than the original RPM tests. 
In particular, these RPM-like subtests do not necessarily implement the ``ladder'' design mentioned in Section~\ref{sec:what-rpm-meansures-exactly}, which is an import feature of the original RPM tests. Nevertheless, three of the four tests surveyed contain subtests that replicate the matrix format of original RPM: Test 3 of Scale 2 and 3 of CFIT, Matrices of CAS2, and Matrix Reasoning of WAIS-IV. To distinguish them with the following RPM-like tasks that we will discuss in later sections, we refer to them as matrix reasoning. Figure~\ref{fig:rpm-like-matrix-reasoning} summarizes matrix reasoning tasks through a diagram. 

As shown in Figure~\ref{fig:rpm-like-matrix-reasoning}, there exist two parts in a matrix reasoning item---the context of this multi-choice problem (Part A) and answer choices (Part B). Part A provides the contextual information through a background and a matrix as foreground. Examples of the background can be found in the items of Figure~\ref{fig:spm-1} and \ref{fig:spm-2}. The matrix varies in size from 1$\times$1 to 4$\times$4 in most tests and has at least one missing entry. To increase the difficulty, there can be some entries, which are intentionally hidden but need not to be completed. As indicated in the Configuration in Figure~\ref{fig:rpm-like-matrix-reasoning}, the locations and numbers of these two types of entries can also be customized for each item. Part B consists of 5 to 8 answer choices in most tests. The reason that we separate answer choices from the context is not only that their functions are different but also that where answer choices are located relative to the context has an influence on the distribution of choosing each answer choices according to human experiment data. Therefore, this is a design choice that need to be considered in test development. This is also a noteworthy point when evaluating AI on RPM-like tasks. That is, it requires more investigation if AI systems behave differently when answer choices are located differently relative to the context and relative to each other.

Although the matrix reasoning tasks replicate the format of original RPM (with slight modifications such as hidden entries and different locations of missing entries), the content of them are more diverse than original RPM. For example, the difficulty of original RPM mainly lies in extracting conceptual relations, and the requirement for perceptual processing is relatively low; but, due to different underlying theories about intelligence, some RPM-like items are designed to load more on perceptual processing abilities, for example, mentally rotating complex 3D objects and the abstract conceptual relations are built upon such demanding perceptual processing.


\subsubsection{Figure Series}



\begin{figure}][!htbp]
    \centering
    \includegraphics[width=\textwidth]{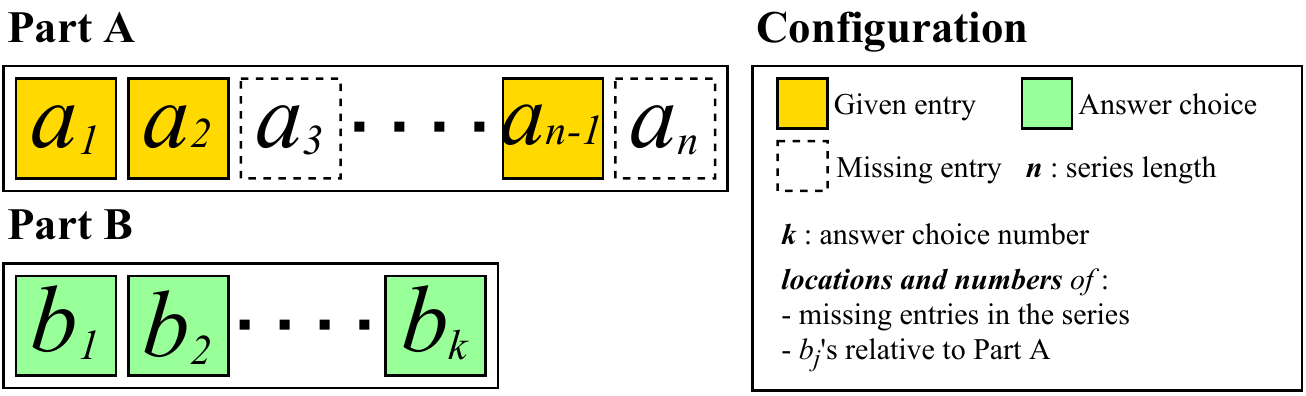}
    \caption{A diagrammatic summary of figure series task}
    \label{fig:rpm-like-figure-series}
\end{figure}

Essentially, what makes RPM items meaningful testing questions is the relations between figures and how these relations are arranged in the 2D structure of matrices. There is no particular reason for using matrix structure. That is, as long as the spatial structure makes sense to the relations, one can use any suitable spatial structures (one could use a circular structure if the relations proceeds and comes back to itself, like modulo addition $+1 \: mod \: N$ and the circle of music keys). Thus, it would not be surprising to see a more fundamental structure---series---to be used in RPM-like tasks, such as Test 1 of Scale 2 and 3 of CFIT, Sequential Order and Repeated Pattern of Leiter-R, and part of Matrix Reasoning of WAIS-IV. We refer to RPM-like items of this structure as figure series. A diagram was given to summarize figure series items in Figure~\ref{fig:rpm-like-figure-series}. 

Figure series has the several characteristics that make it different from other RPM-like tasks. First, the structure of series determines that one or more relations are repeating themselves along the series. Note that the relation is not necessarily a binary relation and it could involve more than two consecutive entries in the series. Second, to provide sufficient contextual information, the figure series are usually longer than a row or a column of matrix reasoning. Third, there could be one or more missing entries in the series. In particular, the missing entry is not necessarily the last one.

Figure series could also be considered a special case of matrix reasoning task by restricting the dimensions of the matrix, but it is also conceptual different from matrix reasoning task. In matrix reasoning, there can be multiple distinct relations along the rows and columns of the matrix. In most cases, the row relations are different from the columns one. One needs to figure out the relations in both row and and columns directions and assemble them to uniquely determine the answer. In figure series, multiple relations are repeating themselves in a single direction.

\subsubsection{Analogy Making}

\begin{figure}[!htbp]
    \centering
    \includegraphics[width=\textwidth]{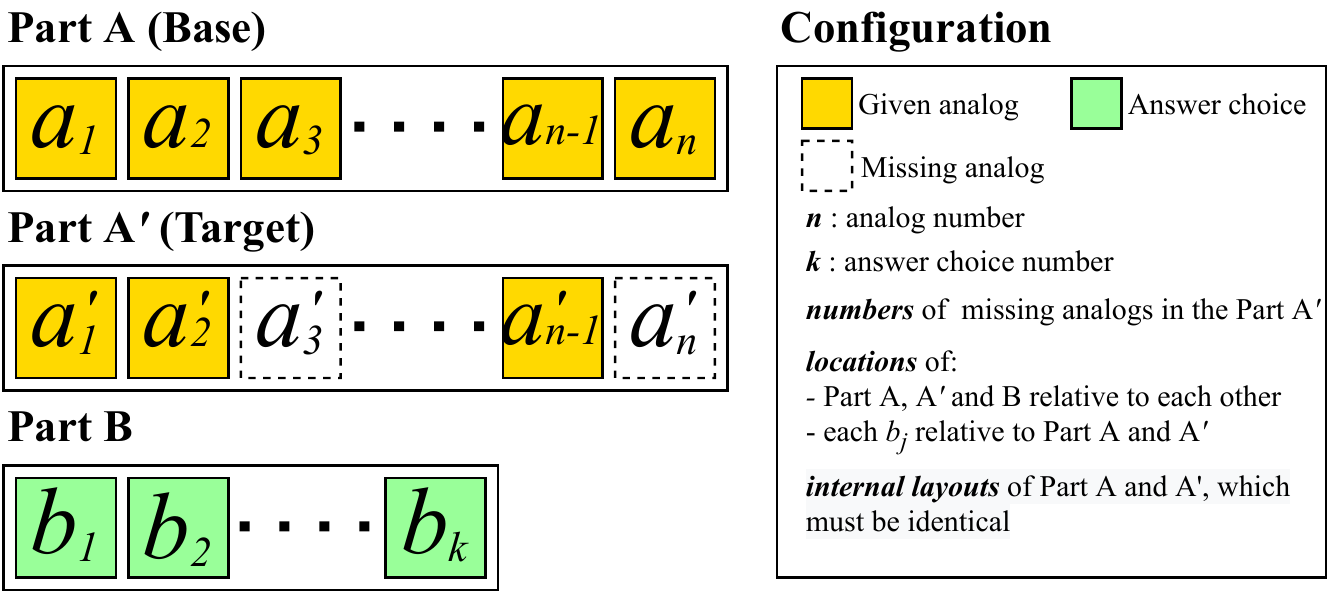}
    \caption{A diagrammatic summary of analogy making task}
    \label{fig:rpm-like-analogy-making}
\end{figure}

Besides modifying the format of original RPM (as in figure series), the context of it could also be viewed from different angles. An important view is from an important human cognitive ability---analogy making. That is, by viewing the matrix entries as analogs, analogies can be drawn between rows, between columns, or between diagonal lines. The correct answer is thus the one that makes the best analogies out of the matrix. Therefore, the nonverbal analogy-making task could be considered as a close relative of RPM. A classic example of this task is the goemetric analogy problems (find images in \citep{lovett2009solving}) published in the 1942 edition of the Psychological Test for College Freshmen of the American Council on Education. These analogy-making items can also be found in intelligence tests we surveyed, such as Design Analogy of Leiter-R and part of Matrix Reasoning of WAIS-IV. A diagrammatic summary of this task is given in Figure~\ref{fig:rpm-like-analogy-making}.   

In the analogy-making task, the context is explicitly separated into two parts, Part A and A' in Figure~\ref{fig:rpm-like-analogy-making}, which are composed of analogs from two different domains. Part A and A' correspond to the base and the target domains in general analogy making situation, where the base domain is usually a familiar one and the target domain is an unfamiliar one which is to be understood through the knowledge in the base domain. The analogy-making task simulates this situation by arranging the analogs in Part A and A' in the same way and removing one or more analogs in the Part A'. Note that, although the analogs in Figure~\ref{fig:rpm-like-analogy-making} are listed in series, this does not mean that the same relations are repeating itself in the series as in Figure series. The analogs could be arranged in any spatial layout when the layout make senses to the relations between analogs. Since the analogs are usually in two series in most intelligence tests, the analogy making task resembles the figure series task. But these two tasks are conceptually different and requires different cognitive abilities. The analogy-making task is also conceptually different from matrix reasoning task even when we artificially separate the rows or columns of a matrix into two parts. This is because, to make an ``interesting'' analogy, the base and target domains must be perceptually distant from each other and higher-order relations must be extracted from both domains. In matrix reasoning, this means that the rows (or columns) must be sufficiently perceptually different. These conditions are not always satisfied in matrix reasoning items, especially when there exist relations in both horizontal and vertical directions.

\subsubsection{Contrastive Classification}

\begin{figure}[!htbp]
    \centering
    \begin{subfigure}{\textwidth}
        \centering
        \includegraphics[width=\textwidth]{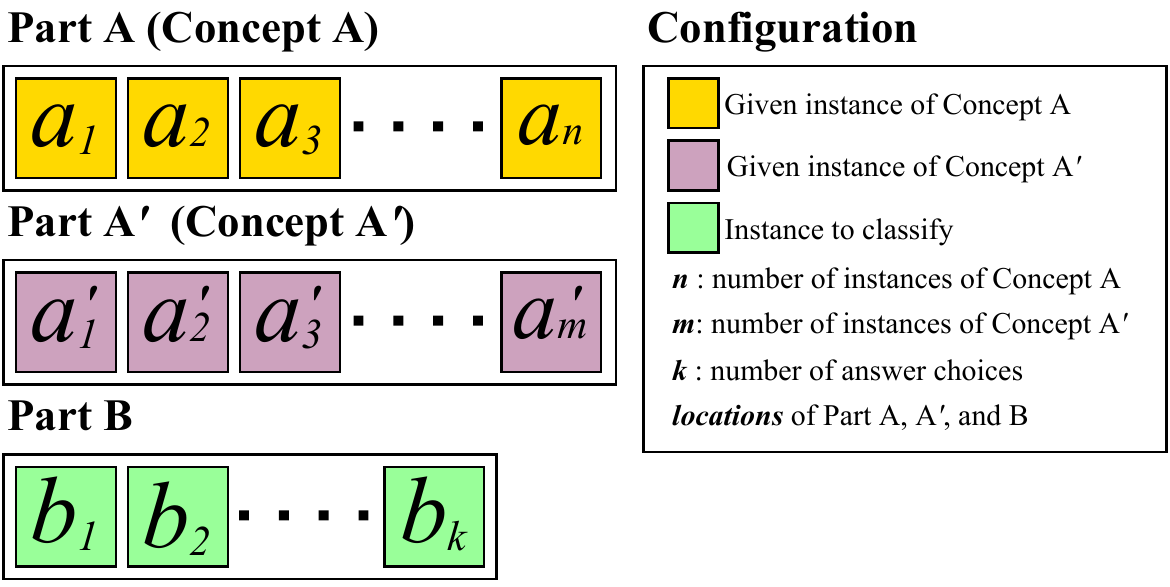} 
        \caption{Explicit contrastive classification}
        \label{fig:rpm-like-explicit-contrastive-classification}
    \end{subfigure}

    \begin{subfigure}{\textwidth}
        \centering
        \includegraphics[width=\textwidth]{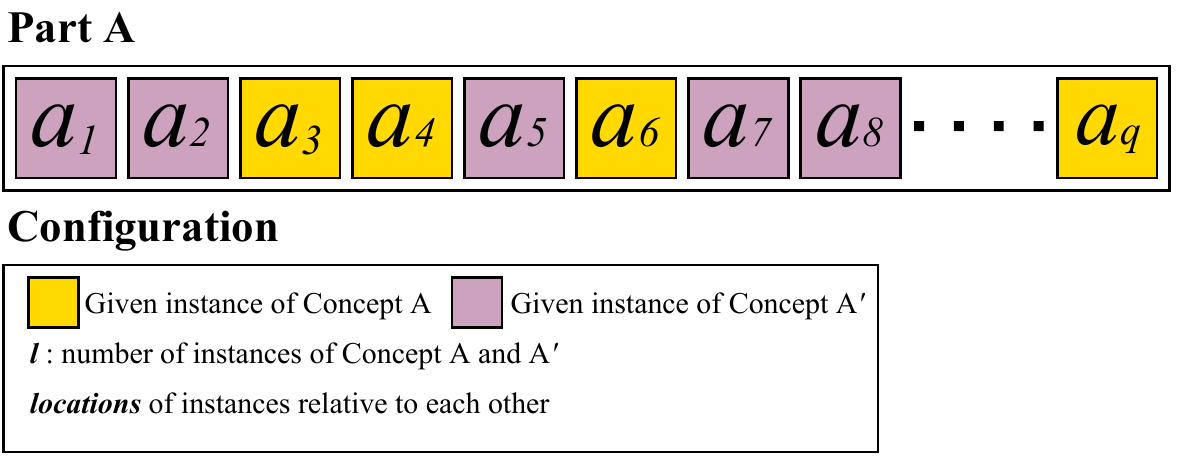}
        \caption{Implicit contrastive classification}
        \label{fig:rpm-like-implicit-contrastive-classification}
    \end{subfigure}
\caption{A diagrammatic summary of contrastive classification task}
\label{fig:rpm-like-contrastive-classfication}
\end{figure}

Classification has long been used to probe human and artificial intelligence. It requires the participant to extract an abstract concept such that the given stimuli can be classified into these concepts. When these stimuli are like the ones in RPM, classification can be regarded as an RPM-like task as they both reason about the relation between multiple visual stimuli. In intelligence tests, classification tasks can be presented in a contrastive manner. That is, two groups of stimuli are presented and the two groups represent two contrastive but related concepts, for example, large-small, concave-convex, and high-low. But note that classification is not limited to antonym pairs for it also uses concept pairs like pentagon-hexagon and more random concepts like topological structures. The advantage of being contrastive is obvious: it allows the usage of complex and diverse concepts (rather than simple concepts describing perceptual attributes) to make the test intellectually interesting to participants; meanwhile, the complex and diverse concept would not make the item too open to solve as the concept is uniquely determined by a unique difference between the two groups. 

The most representative contrastive classification is the Bongard Problems. It requires the participant to verbally describe the conceptual difference between the two groups. In most intelligence tests, contrastive classification is usually multi-choice problems, in which answer choices are selected to be a member of a conceptual group, i.e., identifying instances of the concepts drawn out of the two groups. The contrastive classification are usually presented in two manners---explicit and implicit ones. For explicit ones (Figure~\ref{fig:rpm-like-explicit-contrastive-classification}), the two stimulus groups are explicitly separated, for example, the Bongard Problems and Test 2 of Scale 1 and CFIT. Explicit contrastive classification tasks are also used to evaluate AI system, for example, the SVRT and PSVRT datasets \citep{stabinger2021evaluating}. For implicit contrastive classification (Figure~\ref{fig:rpm-like-implicit-contrastive-classification}), the stimuli from two conceptual groups are mixed together and the participant needs to separate the them into two groups, like the famous Odd-One(s)-Out tests and Test 2 of Scale 2 and 3of CFIT. Note that, in contrastive classification tasks, spatial layout of stimuli is less important compared to matrix reasoning and figure series. The only requirement is that group membership is clearly indicated in explicit contrastive classification.

\subsubsection{Open Classification}

\begin{figure}[!htbp]
    \centering
    \begin{subfigure}{\textwidth}
        \centering
        \includegraphics[width=\textwidth]{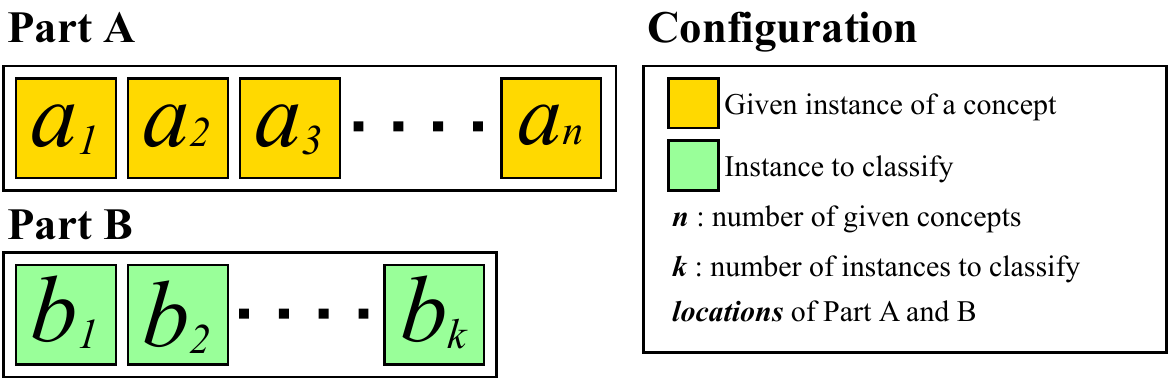} 
        \caption{Explicit Open classification}
        \label{fig:rpm-like-explicit-open-classification}
    \end{subfigure}

    \begin{subfigure}{\textwidth}
        \centering
        \includegraphics[width=\textwidth]{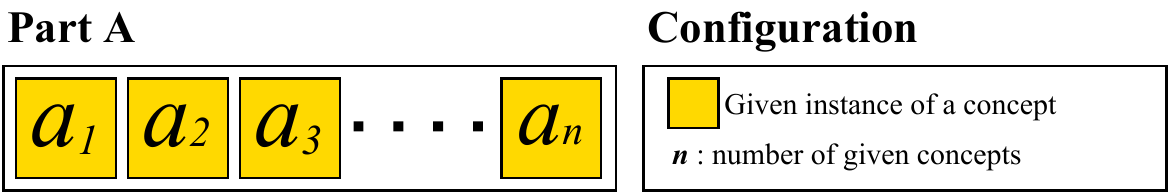}
        \caption{Implicit Open classification}
        \label{fig:rpm-like-implicit-open-classification}
    \end{subfigure}
\caption{A diagrammatic summary of open classification task}
\label{fig:rpm-like-open-classfication}
\end{figure}

Classification task is naturally not contrastive in our daily cognitive activities. That is, the object to classify is not always accompanied by instances of another contrastive concept. Instead of being contrastive, the real-life setting of classification is more based on perceptual and conceptual similarity. Thus, we referred to it as open classification. In particular, the concepts involved in an open classification item can be completely unrelated. There could be only one single concept. For example, in the verbal similarity subtest of WAIS-IV, one would see an item like ``in what way are dolphins and elephants alike'' \footnote{This item requires the participant to classify the objects into one of the many concepts that she knows.}. A possible answer is that they are both animals and a better answer is that they are both mammals. Different answers are scored differently. The more specific the answer, the higher the score. As shown by this example, verbal open classification items require a certain amount of prior knowledge to be intellectually interesting. When open classification is in nonverbal form, it could be considered as a RPM-like task. In the intelligence tests we surveyed, examples of nonverbal open classification include Test 4 of Scale 2 and 3 of CFIT and Classification Subtest of Leiter-R. 

Similar to the contrastive classification, the open classification can also be presented in explicit or implicit ones, as summarized in Figure~\ref{fig:rpm-like-open-classfication}. The explicit open classification (Figure~\ref{fig:rpm-like-explicit-open-classification}) consists of two parts. Part A provides instances of multiple concepts (not necessarily contrastive or even related) with each instance representing a distinct concept. Part B consists of instances to classify into the concepts in Part A by matching to the instances in Part A. The implicit open classification (Figure~\ref{fig:rpm-like-implicit-open-classification}) is similar to the verbal open classification example except that the dolphins and elephants are replaced by nonverbal stimuli. The response format and scoring are also similar to the dolphin-elephant example.

\subsubsection{Summary}
The five categories of RPM-like tasks that we summarized from the intelligence tests are by no means comprehensive. The purpose of them is to expand our attention to the entire problem domain represented by RPM so that the AI research is closer to the nature of the problem domain rather than focusing only solve the original RPM or specific tests. The problem domain is much more diverse and larger than the approximately 100 original RPM items. The problem domain spreads out to all visual stimuli and relations among them that are proper to test people with certain prior knowledge and experience.

In item writing of intelligence test , a good ``taste'' is extremely important. Firstly, a good item first has to be straightforward for the participant to realize that this item is a problem to solve. This point seems saying nothing because any intelligence test item is a problem to solve. The word ``problem'' here should not be understood literally. In particular, the item is a problem to solve not because the the administrator tells the participant it is so or the participant knows that a test is composed of problems. Instead, the participant should realize this by observing the item and forming a conjecture that there should be underlying patterns based on the observation. This conjecture is more of feeling rather than a complete understanding of the solution or patterns, which means that it is based on a rough idea of what should be paid attention to solve the item. This characteristic to give the participant this feeling is important because it makes the item intellectually interesting and attractive to the participants and the participant is thus motivated to solve the item. Without this characteristic, the participant would possibly give invalid responses, for example, giving random responses without thinking. 

The second point in item writing is that the scope of item content should allow a large range of difficulty. Specifically, it should allow to create rather difficult items to test highly intelligent individuals. This point in itself is not an issue because there exist a huge amount of sophisticated abstract relations and patterns if one delves into any specific field. But, when combined with the first point---straightforward as a problem, this poses a great challenge because these points are contradicting to each other in many cases. A master of item writing is one who can reconcile these two points and achieve a combined effect that when the participant sees the item, she immediately understands in what way it is problem to solve and invests effective intellectual effort to solve it, and when a correct answer is reached, it would be an aha moment that the participant strongly believes that the problem is solved. In this sense, the five categories of RPM-like items mentioned above are masterpieces of item writing. But this does not mean that the problem domain is limited to these categories, and more efforts are needed to further explore the problem domain.



\subsection{Algorithmic Item Generation of Matrix Reasoning}

Algorithmic Item Generation (AIG) refers to approaches using computer algorithms to automatically create testing items. AIG was initially introduced to address the increased demand for testing items in the special test settings:
\begin{itemize}
    \item Large-scale testing, for example, repeated tests in academic settings and longitudinal experiments, where many parallel forms are needed due to the retest effect.
    \item Adaptive testing, in which the next items are determined by the responses to previous items, which is a more efficient and reliable testing form, but also requires larger item banks.
    \item Computer-based and internet-based testing, which makes standardized tests more accessible to the public and brings the exposure control issue to a new level.
\end{itemize}
For AIG to work, test developers must have a deep understanding of what is measured and the corresponding problem domain, from which items are generated. In addition, test developers also need to examine the testing properties of generated items, such as validity and reliability, as they are examined in handcrafted tests. AIG has been studied and used in different areas, such as psychometrics, cognitive science, and education. It can be used to a wide range of testing items from domain-general tests, such as human IQ tests, to domain-specific tests, such as medical license tests \citep{gierl2012using}. 

As RPM-like tasks are more and more used in human intelligence testing and AI testing, the demand for RPM-like items has been increasing rapidly. In particular, since data-driven AI systems were applied on RPM-like tasks, the scale of this demand has been changed from hundreds to millions, which is impossible for human item writers to satisfy. Thus, AIG of RPM-like items has been receiving more and more attention. However, AIG of RPM-like items have been studied separately in different research fields. In this subsection, we aggregate these works from different fields together and systematically explore how AIG of RPM-like items works in both human intelligence testing and AI testing. To have a thorough discussion on technical details and theoretical implications, we focus on the matrix reasoning task, which is the most widely studied RPM-like task in both human intelligence and AI. In the rest of this subsection, we first review the AIG works of matrix reasoning for human testing. Then, we switch to the ones for AI testing. 



\begin{sidewaystable}
\small
\centering
\caption{The technical details of generator programs in the reviewed works.}
\label{tab:tech-details}

\begin{threeparttable}

\begin{tabularx}{\textwidth}{X X X X X X X X X X X X X}
\toprule
    \multirow{2}{*}{Generator} & \multirow{2}{1cm}{Format}   & \multirow{2}{2cm}{Analogical Direction\tnote{a}} & \multicolumn{6}{c}{Geometric Element} & \multicolumn{2}{c}{Rules} & \multirow{2}{2cm}{Per. Org.\tnote{d}} &  \multirow{2}{1cm}{Answer Set\tnote{e}} \\
    \cmidrule(lr){4-9} \cmidrule(lr){10-11}
    & & & Type\tnote{b} & Size & Color & Angle & Number & Position & Type\tnote{c} & Number & & \\
\midrule
    Rule-Based & 3$\times$3 & R, C, O & - & - & - & - & 1-2 & - & I, S, P, D & 1-2 & S, I, E & - \\
    \hline
    Cognitive & 3$\times$3 & R, C, O & - & - & - & - & - & - & I, S, P, D & 1-2 & O, F, D & 8 \\
    \hline 
    GeomGen & 3$\times$3 & R, C & S, C,T, D2, H, R & -  & - & - & - & 3$\times$3 grid & I, S, P, N & - & C, N & 5+1\\
    \hline
    Sandia & 3$\times$3 & R, C, M, S, O & O, R, T, D1, T1, T2 & 5 sizes & 5 colors & 5 angles & 1-6 & center & S, P & 1-3 & center, overlay & 8 \\
    \hline
    CSP & 3$\times$3 & R, C & - & - & - & - & - & - & S, P, D & 1-7 & - & 8 \\
    \hline
    IMak & 2$\times$2 & O & C, L, D0, T1 & 1 size & 1 color & 8 angles & 3 & - & P & 1-4 & E & 8+2 \\
    \hline
    PGM-shape & 3$\times$3 & R, C & C, T, S, P, H & 10 sizes & 10 colors & - & 0-9 & 3$\times$3 grid & S, P, D & 1-4 & 3$\times$3 grid & 8 \\
    \hline
    PGM-line & 3$\times$3 & R, C & L, D1, C & 1 size & 10 colors & 1 angle & 1-5 & center & S, P, D & 1-4 & center, overlay & 8 \\
    \hline
    RAVEN & 3$\times$3 & R & T, S, P, H, C & 6 sizes & 10 colors & 8 angles & 1-9 & fixed in configurations & I, S, P, A, D & 4 & 7 configurations & 8 \\
\bottomrule
\end{tabularx}

\begin{tablenotes}
    \item[a] R=row, C=column, M=main-diagonal, S=secondary-diagonal, O=R+C (see \citep{matzen2010recreating}).
    \item[b] S=square, C=circle, R=rectangles, T=triangles, H=hexagons, O=oval, L=line, D0=dot D1=diamond, T1=trapezoid, T2=``T'', D2=deltoid, P=pentagon..
    \item[c] I=identify, S=set/logical operations, P=progression, D=Distribution of 2 or 3 values, N=neighborhood, A = arithmetic (see \citep{carpenter1990one, arendasy2002geomgen}).
    \item[d] S=separation, I=integration, E=embedding, C=classical, N=normal, O=overlay, F=fusion, D=distortion (see \citep{hornke1986rule, embretson1998cognitive, primi2001complexity}).
    \item[e] +1=``none of the above'', +2=``none of the above'' + ``I don't know''.
\end{tablenotes}

\end{threeparttable}
\end{sidewaystable}

\subsubsection{Algorithmically Generating Matrix Reasoning Items for Human Intelligence Testing}
 
Human intelligence tests consist of items which are carefully handcrafted by strictly following the procedures of psychometrics and theories of human intelligence. In particular, Handcrafted items must go through iterations of evaluation and calibrating for good psychometric properties before being included in the final item bank. The attrition rate could be up to 50\% \citep{embretson2004measuring}. A variety of efforts in AIG have been made to free item writers from this onerousness. In the following, we discuss the typical AIG works of matrix reasoning for human intelligence testing. The title of each reviewed work is followed by a keyword of its most outstanding characteristic. 
The technical details of the works are summarized in Table \ref{tab:tech-details}.


\paragraph{Rule-Based Item Construction---Human-Based AIG}
The term ``algorithmic item generation'' is more often ``automatic item generation'' in literature. The word ``automatic'' alludes to the usage of computer. But the algorithms and the theories of what to measure that support the algorithms are the very essence of AIG, rather than the computer. As it will be shown in this first reviewed work, computer is not necessary. \citet{hornke1986rule} conducted one of the earliest studies, if not the earliest, on AIG of matrix reasoning items. They created a procedure for item generation, hired university students to manually execute the procedure, and created 648 3$\times$3 items. Each step in this procedure has finite clearly defined options so that the student can choose between them randomly. Although the diversity and complexity of these items are not comparable to ones handcrafted by human experts, no one had ever ``automatically'' created so many items before \citet{hornke1986rule}.

\begin{figure}[!htbp]
    \centering
    \includegraphics[width=\textwidth]{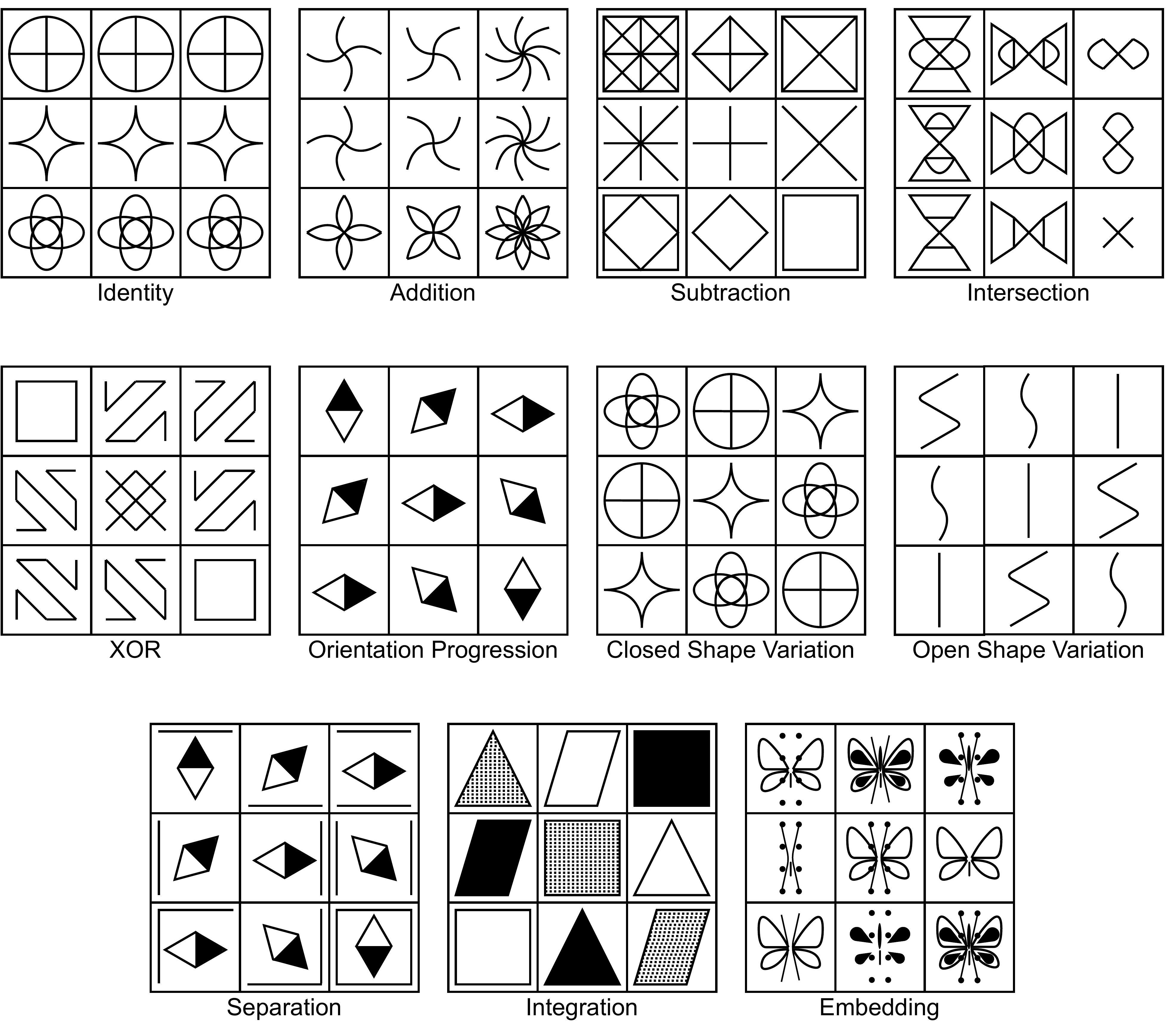}
    \caption{Example items created by following \citeauthor{hornke1986rule}'s AIG procedure.}
    \label{fig:hb}
\end{figure}

\citeauthor{hornke1986rule} considered the item writing task as the reverse of solving, which can be decomposed into three types of cognitive operations, which address three independent dimensions of the solving process. To generate items, \citeauthor{hornke1986rule} thus designed a procedure that sequentially make choices on the three dimensions by selecting from finite sets of options:
\begin{itemize}
    \item Variation rules of geometric elements: eight options are provided (see the first 8 matrices in Figure \ref{fig:hb} for examples)---identity, addition, subtraction, intersection, exclusive union (or symmetric difference), progression, variation of open/closed gestalts (i.e. permutation of three hollow/solid shapes). 
    \item Analogical directions: a variation rule proceeds in row or column direction.
    \item Perceptual organizations: this dimension addresses how multiple variation rules are combined into a stimulus in a matrix entry. Three options are provided (see the last 3 matrices in Figure \ref{fig:hb} for examples): separation, integration, and embedding. Separation means that separate geometric elements are used for different variation rule; integration means that different attributes of a single geometric element are used for different variation rules; and embedding means that different parts of a single geometric element are used for different variation rules.
\end{itemize}

In their experiment, the hired students were given a set of geometric shapes (e.g. differently sized squares and triangles) and instructed to create items by jointly sampling the 3 dimensions and geometric shapes from the given set. The students were told to create each item by combining at most two variation rules. Therefore, the resulting item bank contained only 1-rule and 2-rule items. Human experiments on this item bank showed that the cognitive operations corresponding to these 3 dimensions explained approximately 40\% of the item difficulty. As for the unexplained 60\%, other early studies \citep{mulholland1980components} indicated that the numbers of elements and rules were also major sources of difficulty. Although this ``human-based'' AIG work looks a bit primitive compared to the computational power today, the way it decomposes the generating process has a long-lasting influence on the following works.


\paragraph{Cognitive Design System Approach---Combination of Cognitive Modeling and Psychometrics}
\citet{embretson1995role, embretson1998cognitive, embretson2004measuring} introduced the Cognitive Design System Approach. Different from other AIG works that focus on generating items, this approach focuses on human testing by integrating cognitive modeling and psychometric models and theories (such as IRT theory and models) into a procedure that is similar to how human experts create and validate intelligence tests. A matrix reasoning item bank was generated as a demonstration.

This approach starts with cognitive modeling of the solving process of an existing cognitive ability test at the information-processing level. In the demonstration, \citeauthor{embretson2004measuring} reused the cognitive model proposed by \citep{carpenter1990one}, which have also been used in many other AIG works of matrix reasoning. However, \citeauthor{embretson2004measuring} also pointed out that the cognitive model did not include perceptual encoding or decision processes in the solving process. Thus, \citeauthor{embretson2004measuring} incorporated three extra binary perceptual stimulus features---object overlay, object fusion, and object distortion---in the generation procedure, which represent three different types of mental decomposition of the complete gestalt into its basic parts. Object overlay and fusion are similar to separation and embedding in Figure \ref{fig:hb}, while object distortion refers to perceptually altering the shape of corresponding elements (e.g. bending, twisting, stretching, etc.). A software---ITEMGEN---was developed based on this approach.

Once the cognitive models are determined, the stimulus features are accordingly determined. It then integrates these features into psychometric models to estimate item properties (e.g. item difficulty and item discrimination), formulated as parameterized functions of the  stimulus features. The function parameters are initially set by fitting the psychometric models to human data on the existing cognitive ability test. Thereafter, the item properties of newly generated items (by manipulating the stimulus features) can be predicted by these functions. The prediction and empirical analysis of the newly generated items are compared to further adjust the parameters. Once the functions are sufficiently predictive, the psychometric model can be integrated into an adaptive testing system to replace a fixed item bank and generate items of expected properties in real-time. To sum up, the Cognitive Design System Approach is more than constructing an item generator; it also takes into account the psychometric properties of the generated items.

\paragraph{MatrixDeveloper---4-by-4 Matrices}
MatrixDeveloper \citep{hofer2004matrixdeveloper} is an unpublished software for generating matrix reasoning items. It has been used in a series of psychometric studies of algorithmically-generated matrix reasoning items  \citep{freund2008explaining,freund2011get, freund2011retest, freund2011wants}. According to the limited description in these studies, the MatrixDeveloper is similar to the Cognitive Design System Approach in terms of variation rules (e.g. the five rules of the cognitive model of \citep{carpenter1990one}) and perceptual organizations (i.e. overlap, fusion, and distortion). The difference is that it generates 4$\times$4 matrix items, which are uncommon for matrix reasoning task. Theoretically, it can thus accommodate more variation rules than 3$\times$3 or 2$\times$2 matrices so that the differential effects of variation rules can be better studied.

\paragraph{GeomGen---Perceptual Organization}
The early cognitive modelings of solving handcrafted matrix reasoning items tend to characterize the items by the numbers of elements and rules and types of rules, for example, \citep{mulholland1980components,bethell1984adaptive,carpenter1990one}. This characterization is consistent with the firsthand experience of working on the items and direct measures of human behavior (such as accuracy, response time, verbal protocols, and eye-tracking). In addition, the rationale of this characterization could be explained through the working memory theory of Baddeley and Hitch. However, for creating new items, we need to consider at least one more factor---perceptual organization \citep{primi2001complexity}. It tells how geometric elements and rules are perceptually integrated to render the item image. For example, the third dimension in the procedure of \citet{hornke1986rule} is a specific way to deal with perceptual organization. More generally, perceptual organization involves the Gestalt grouping/mapping of elements using Gestalt principles such as proximity, similarity, and continuity. This factor is less clearly defined and no systematic description of this factor has ever been proposed. But, to create new items, one has to adopt some formalized ways to manipulate perceptual organization.

\citep{arendasy2002geomgen, arendasy2005effect} proposed a generator program---GeomGen---that adopted a binary perceptual organization, which was reused and extended in many following works.The perceptual organization in GeomGen provides two options---classical view and normal view. In classical view, the appearance of geometric elements changes while numbers and positions of them remain constant across matrix entries. In normal view, numbers and positions of elements change while the appearance of them remain constant across the matrix entries. An obvious difference between the two views is how the correspondence between elements from two matrix entries is established. And this difference is important because it leads to items that requires different cognitive processes at the very first step of correspondence finding before the rules between matrix entries are considered.

The taxonomy of perceptual organization in GeomGen is only a specific way to define perceptual organization but by no means the unique way. For example, \citep{primi2001complexity} proposed another important taxonomy---harmonic and nonharmonic, which, together with GeomGen taxonomy, forms a more comprehensive description of perceptual organization that is adopted in many following AIG works.  

\begin{figure}[!htbp]
    \centering
    \includegraphics[width=\textwidth]{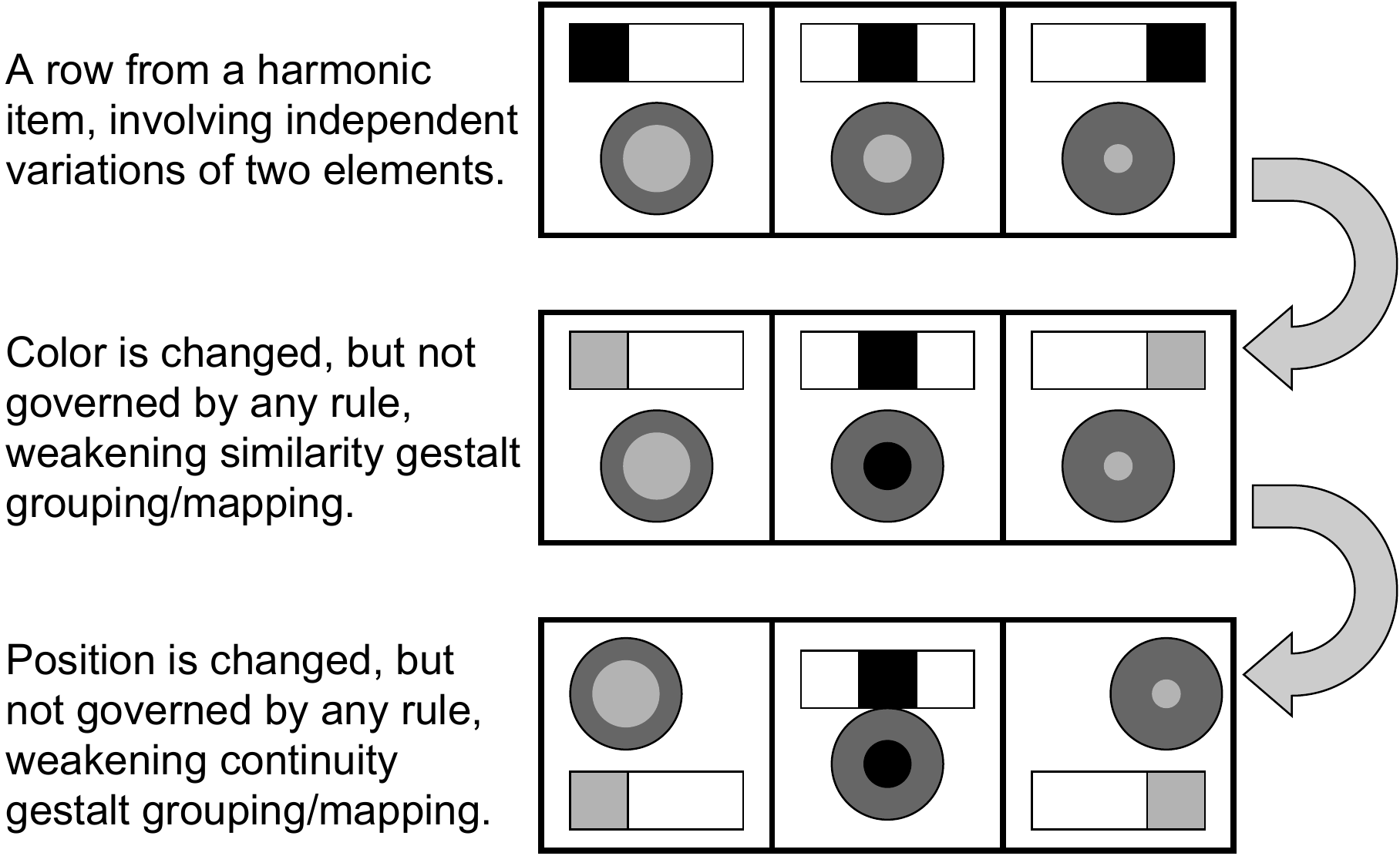}
    \caption{An example of deriving nonharmonic items from harmonic items.}
    \label{fig:harmonic}
\end{figure}

\citet{primi2001complexity} describes ``harmonic organizations as visually harmonic items display perceptual and conceptual combinations that represent congruent relationships between elements, whereas nonharmonic organizations tend to portray competitive or conflicting combinations between visual and conceptual aspects that must be dealt with in reaching a solution.'' \citet{primi2001complexity} mentioned that, in the practice of AIG, the nonharmonic items could be derived from the harmonic ones by manipulating the geometric elements to cause misleading Gestalt groupings, as shown in Figure \ref{fig:harmonic}. The correct Gestalt grouping/mapping (i.e. element correspondences) are obvious in harmonic items, whereas nonharmonic items requires extra cognitive effort to resolve the conflict between competing gestalt groupings and mappings. 

In summary, the contributions of all the aforementioned factors---the number of elements, the number of rules, the type of rules, and perceptual organization---to item complexity could be explained by their effect on the central executive component of working memory. But the ways they exert their influences are different. The number of elements and rules relate to the short-term memory management and goal (or strategy) management, whereas the type of rules and perceptual organization relate to selective encoding and short-term memory management \citet{primi2001complexity}. According to the literature of AIG of matrix reasoning, the type of rules and perceptual organization are less investigated and might be important for understanding the solving process of matrix reasoning and the item difficulty. Several human studies came to the same conclusion \citet{primi2001complexity, arendasy2005effect, meo2007element}, while other researchers might have different opinions on this \citep{embretson1998cognitive, carpenter1990one}.

\paragraph{Sandia Matrix Generation Software---High-Fidelity SPM Generator}

The previous works study AIG more from the perspective of cognitive science and psychometrics. Less details about algorithms and software development were given in the works. But, in practice, we are also interested in how these ideas are implemented and, especially, accessibility of the generator software. \citet{matzen2010recreating} provided in their work a representative example of this that could ``recreate'' the 3$\times$3 SPM with high fidelity---Sandia Matrix Generation Software. 

\citet{matzen2010recreating} identified two basic types of 3$\times$3 items in SPM--- the element transformation and the logic problems. An element transformation refers to a progressive variation of a certain attribute of the element. There could be multiple variations in different directions, for example, a color variation in the row direction and a size variation in the column direction. However, in every single direction, there is only one attribute varying. This is because, on one hand, it is so in the original SPM, on the other, multiple attributes varying in the same direction does not increase complexity of the problem (to human participants) compared to only one attribute. 
The attributes considered for transformation problems are shape, shading, orientation, size, and number, each of which takes values from an ordered categorical domain. The logic problems involve operations such as addition/subtraction, conjunction (AND), disjunction (OR), or exclusive disjunction (XOR) of elements. Each generated item is either a transformation one or a logic one, but not both.

In addition, Sandia Matrix Generator generates answer choices in a way of the original SPM problems. An incorrect answer choice could be (a) an entry in the matrix, (b) a random transformation of an entry in the matrix, (c) a random transformation of the correct answer, (d) a random transformation of an incorrect answer, (e) a combination of features sampled from the matrix, or (e) a combination of novel features that did not appear in the matrix. 

The item difficulty was studied through an item bank of 840 generated items. The problem set contained problems of 1, 2 or 3 rules (in row, column or diagonal direction). Note that the original SPM problem does not contain 3-rule problems. The generated problem set and the original SPM were given to the same group of college students. Experimental data showed that the generated items and the original SPM had very similar item difficulty. 
In particular, the data further showed that the item difficulty was strongly affected by the number of rules, analogical directions, and problem types (i.e., transformation problems versus logic problems).  

\paragraph{CSP Generator---First-Order Logic Representation}


A more important thing about AIG is to give a general formal description of the generating process, rather than developing various specific generator software. \citet{wang2015automatic} made such an effort to formalize the generating process of matrix reasoning items through the first-order logic, and turned AIG into a constraint satisfaction problem (CSP) by formulating the ``validity''\footnote{Not exactly the same definition of validity in psychometrics} of RPM items into a set of first-order logic propositions. 

\begin{figure}[ht]
    \centering
    \includegraphics[width=.8\textwidth]{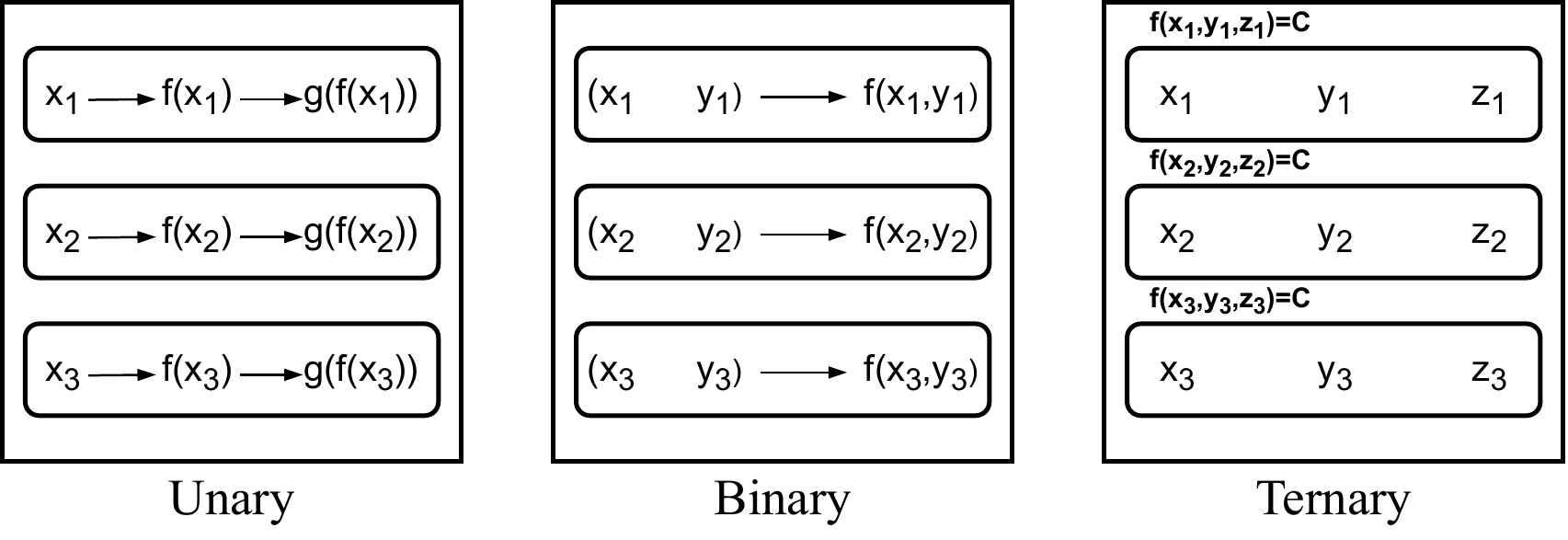}
    \caption{Three categories of relations commonly used in matrix reasoning \citep{wang2015automatic}.}
    \label{fig:3-relations}
\end{figure}

In particular, a variation rule is represented as an instantiation of Equation \eqref{eq-variation-pattern} and \eqref{eq-constraint}, 
{\begin{align}
    \exists \alpha \: \: \forall i \in \{1, 2, 3\}  \: \: \exists o_{i1}, o_{i2}, o_{i3} \: \: P (\alpha, o_{i1}, o_{i2}, o_{i3}) \label{eq-variation-pattern} \\
    P (\alpha, o_{i1}, o_{i2}, o_{i3}) = Unary (\tau (o_{i1}, \alpha), \tau (o_{i2}, \alpha), \tau (o_{i3}, \alpha)) \wedge \label{eq-constraint} \\
    Binary (\tau (o_{i1}, \alpha), \tau (o_{i2}, \alpha), \tau (o_{i3}, \alpha)) \wedge \nonumber \\
    Ternary (\tau (o_{i1}, \alpha), \tau (o_{i2}, \alpha), \tau (o_{i3}, \alpha)) \nonumber
\end{align}}
where $\alpha$ is a goemetric attribute, $o_{ij}$ is a geometric elements in the figure of Row $i$ and Column $j$, $\tau (\alpha, o_{ij})$ is the value of $\alpha$ of $o_{ij}$, and $P$ is a predicate that describes the variation pattern of attribute $\alpha$ in each row. In Equation \eqref{eq-constraint}, the predicate $P$ further equals a conjunction of three predicates---$Unary$, $Binary$, and $Ternary$---representing three categories of relations commonly used in matrix reasoning, as illustrated in Figure~\ref{fig:3-relations}.

An interesting observation of Figure \ref{fig:3-relations} is that, mathematically, the unary relation is a special case of the binary relation, which is a special of the ternary relation. That is, the ternary relation is theoretically sufficient to generate all the items. However, interpreting the same variation as unary, binary and ternary relations requires different working memory abilities and thus leads to different difficulties. Therefore, these three categories are cognitively different, and need to be separately included in a generator program to achieve a better control over psychometric properties.

Equation \eqref{eq-variation-pattern} and \eqref{eq-constraint} represent only the variation pattern of a single attribute $\alpha$. There could be multiple variation patterns of different attributes in a matrix, i.e., multiple different instantiations of Equation \eqref{eq-variation-pattern} and \eqref{eq-constraint}. Meanwhile, it is also possible that some attributes are not assigned any instantiations of Equation \eqref{eq-variation-pattern} and \eqref{eq-constraint}. In this case, they could be given either constant values or random values across matrix entries. Random values may cause distracting effects in the generated items, which is similar to the nonharmonic perceptual organizations in \citep{primi2001complexity}. 

To generate an item through Equation \eqref{eq-variation-pattern} and \eqref{eq-constraint}, the generator program samples values from finite domains to determine (a) the number of rules (i.e., the number of the instantiations of Equation \eqref{eq-variation-pattern} and \eqref{eq-constraint}), (b) the attribute $\alpha$ for each rule, (c) the values of $\tau (\alpha, o_{ij})$, (d) the specific types of $Unary$, $Binary$, and $Ternary$ relations. The matrix image is rendered from the instantiations of Equation \eqref{eq-variation-pattern} and \eqref{eq-constraint}, and each incorrect answer choice is generated by breaking an instantiation of Equation~\eqref{eq-variation-pattern} and \eqref{eq-constraint} (i.e., using values not satisfying them).

The generated items and the APM test were also given to a small group of university students. The experimental data showed that the overall difficulty and rule-wise difficulty (number of rules) were similar to the items in APM. However, as the author pointed out, their generator could not synthesize all the items in APM for some underlying transformations were hard to implement. When the items were created with distracting attributes, the generated items became much more difficult for human subjects.


\paragraph{IMak Package---Open Source}

Although there have already been many works on AIG of matrix reasoning, the generator software and the source code are usually not easily available to the public. This makes it hard to reproduce and build upon these works. \citet{blum2018automatic} realized this point and released their generator as an R package---IMak package---that is globally available via the Comprehensive R Archive Network. The source code and detailed documentation of their work come with the R package. New items could be obtained by simply three lines of R code in the R interpreter---one for downloading the package,one for importing the package,and one for generating the items.

\begin{figure}[ht]
    \centering
    \begin{subfigure}{0.32\textwidth}
        \centering
        \includegraphics[width=\textwidth]{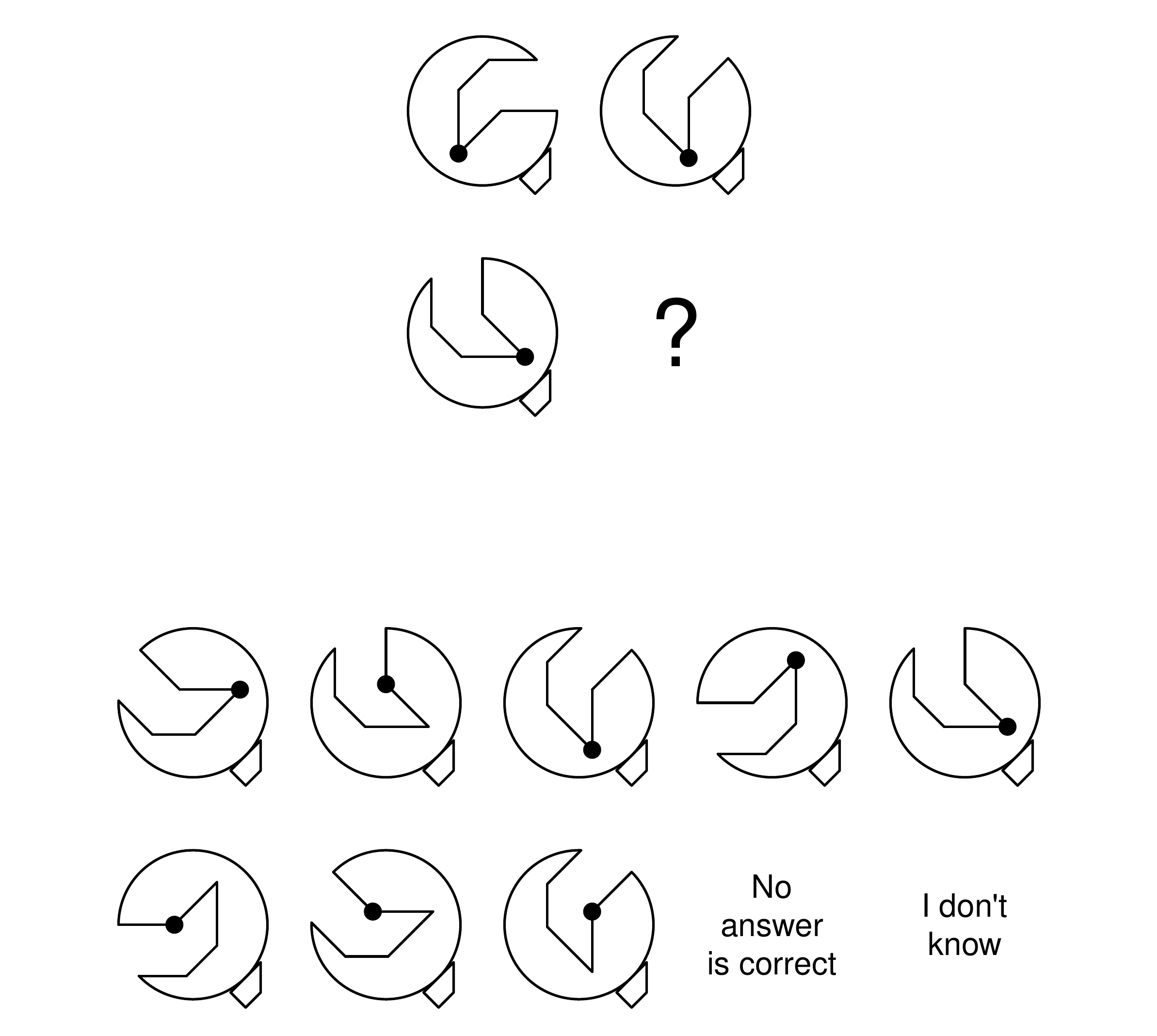} 
        \caption{Main shape rotation}
        \label{fig:imak_main_rot}
    \end{subfigure}
    \begin{subfigure}{0.32\textwidth}
        \centering
        \includegraphics[width=\textwidth]{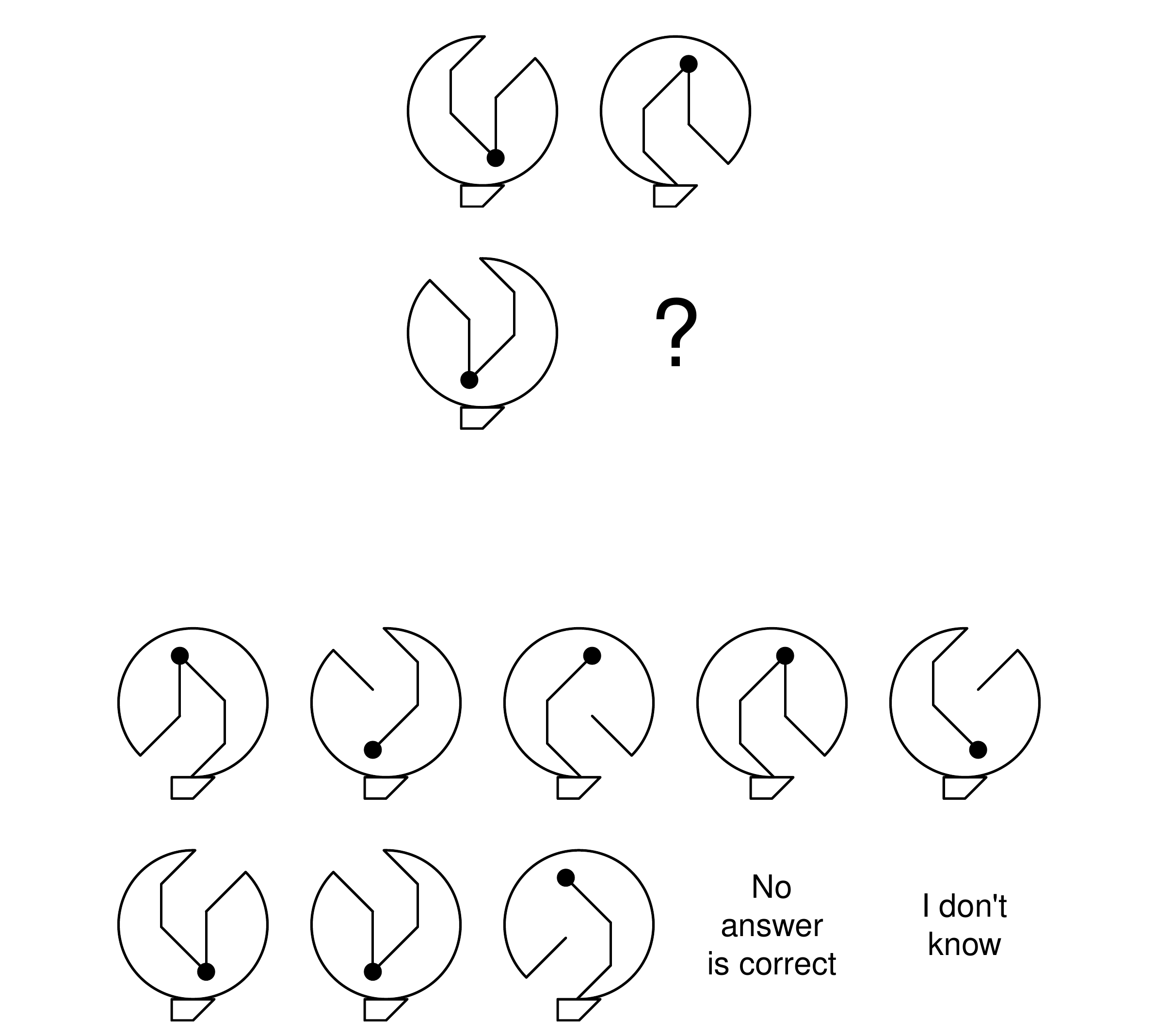}
        \caption{Main shape reflection}
        \label{fig:imak_mirror}
    \end{subfigure}
    \begin{subfigure}{0.32\textwidth}
        \centering
        \includegraphics[width=\textwidth]{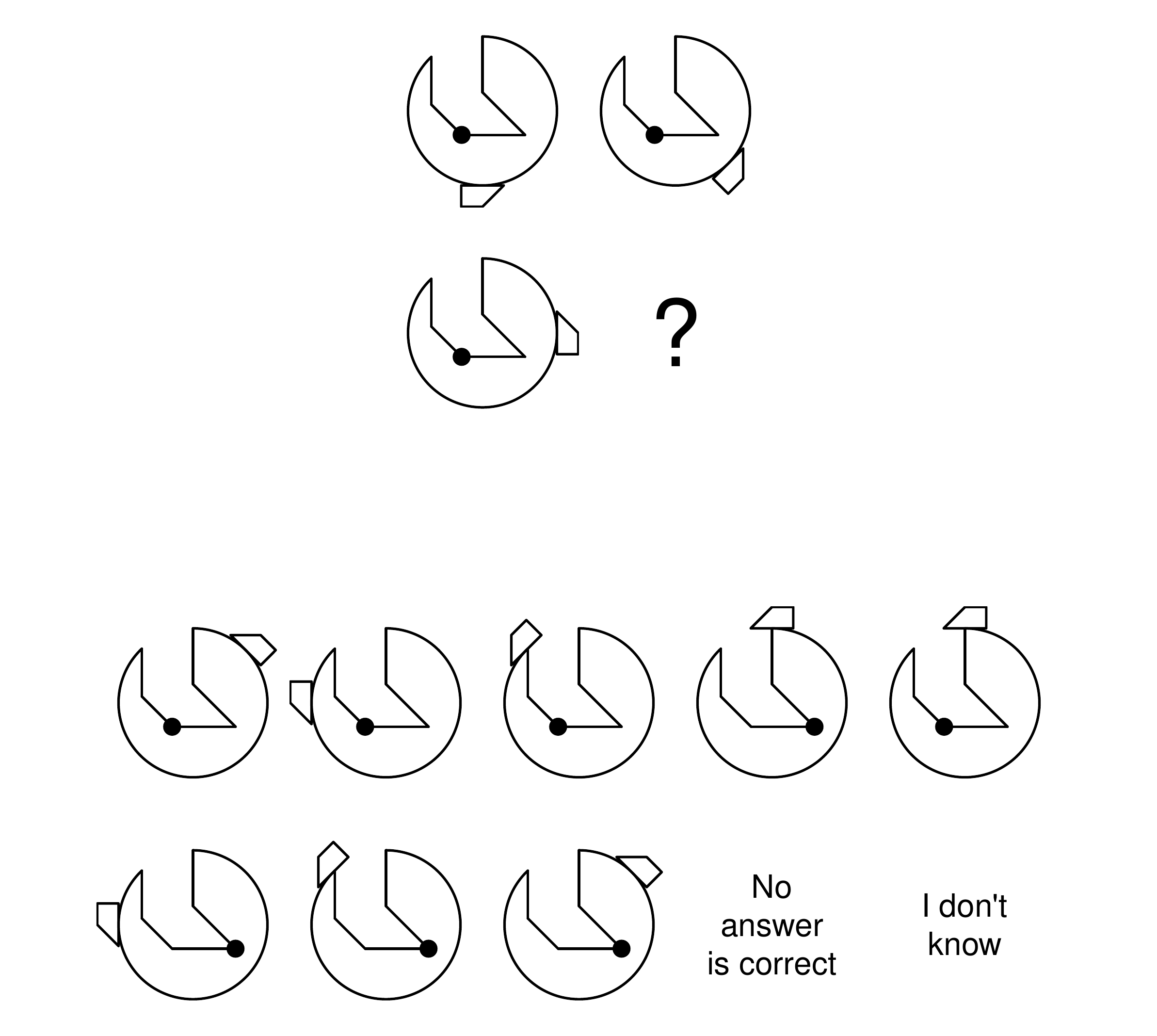}
        \caption{Trapezium rotation}
        \label{fig:imak_trap}
    \end{subfigure}
    
    \begin{subfigure}{0.32\textwidth}
        \centering
        \includegraphics[width=\textwidth]{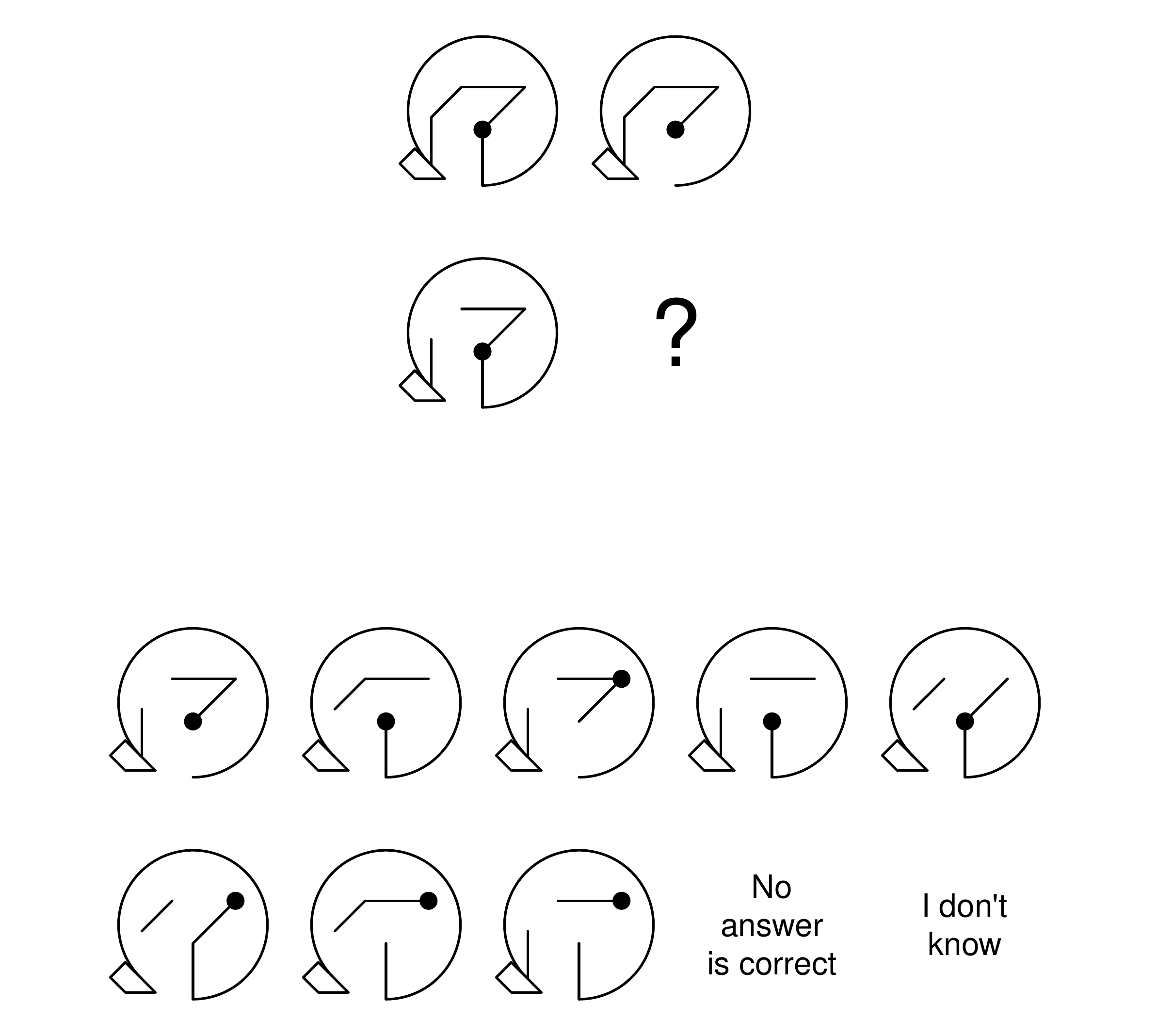}
        \caption{Line segment subtraction}
        \label{fig:imak_subtract}
    \end{subfigure}
    \begin{subfigure}{0.32\textwidth}
        \centering
        \includegraphics[width=\textwidth]{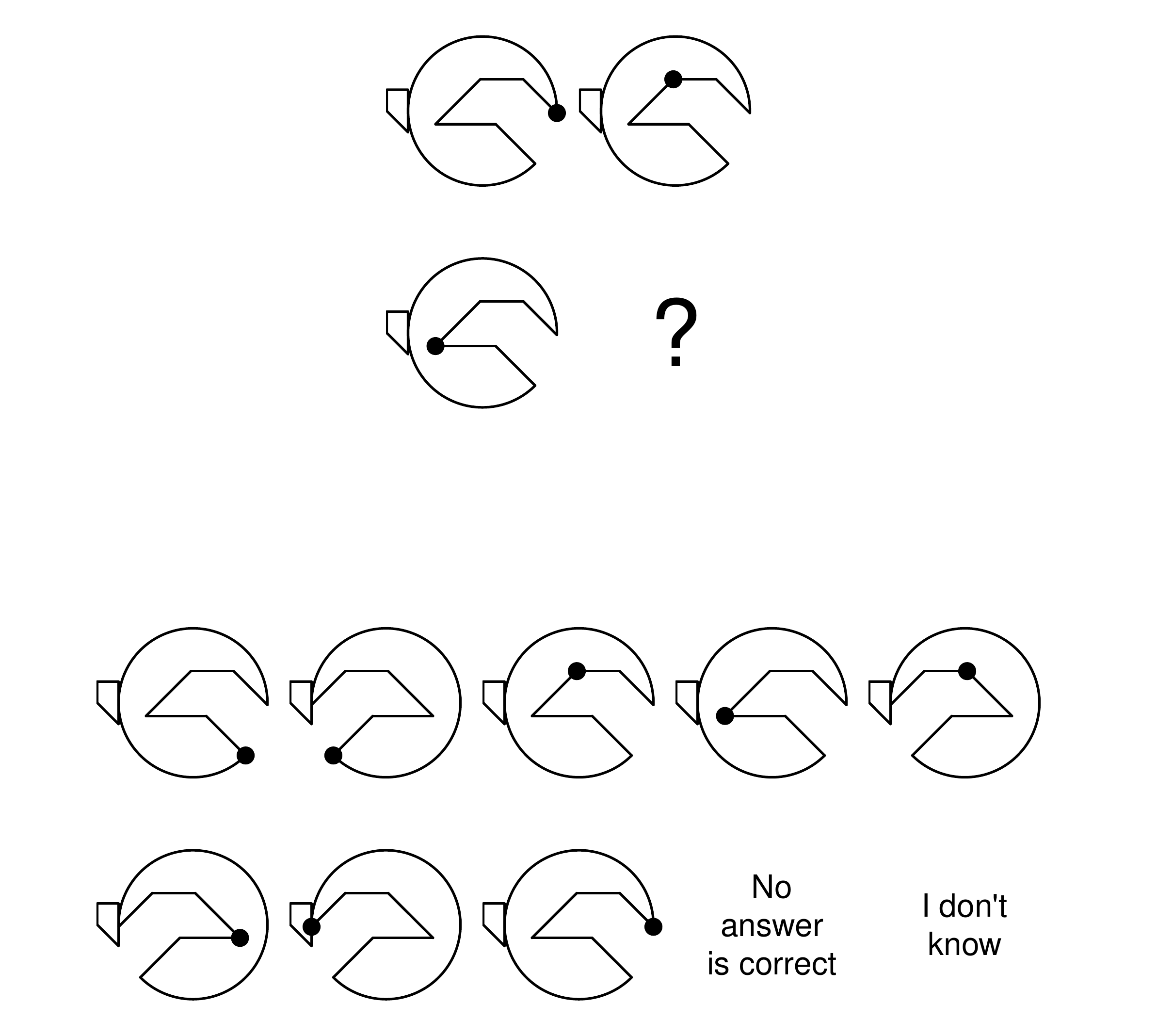}
        \caption{Dot movement}
        \label{fig:imak_dot_mov}
    \end{subfigure}

\caption{Example items generated through the IMak package. Each item exemplifies a single basic rule. The correct answer is set to the first answer choice for demonstration.}
\label{fig:imak}
\end{figure}

The author's purpose of developing the IMak package is to study the effect of types of variation rules on item difficulty.  
The generator was thus designed to manipulate the types of rules while keeping other factors constant, and, thus, the generated items look quite different from the generated items of the generators mentioned above. For example, Figure~\ref{fig:imak} shows some example items that we created through this package, each of which exemplifies a basic rule type. With the current release (version 2.0.1), the geometric elements are limited to the main shape (the broken circle plus the polyline in it), the trapezium that is tangent to the main shape, and the dot at one of the corners of the polyline. Furthermore, the size and shape of these element are fixed for all generated items, but the position, orientation and existence would vary according to 5 basic rules.

As shown in Figure \ref{fig:imak}, there are 5 basic rules in IMak. All the rules are in the outward analogical direction (i.e. row and column). For example, in Figure \ref{fig:imak_main_rot}, the main shape is rotated counterclockwise by 45 degrees in the first row; the main shape is rotated counterclockwise by 90 degrees in the first column. Then the correct answer would be a counterclockwise rotation of the main shape by 135 (45 + 90) degrees compared to the top left one. Similarly, all the other 4 examples follow the same analogical direction. Each item could contain up to 4 rules (because mains shape rotation and reflection are conflicting). This design seems to excessively simplify the RPM-like problems, but it does serve the very purpose of study the differential effect of rules by fixing other factors.


Besides open-source accessibility and the special design of geometric elements, IMak has four other distinctive features that are inspiring for following works. Firstly, IMak generates 2$\times$2 format of AIG of matrix reasoning. Being affected by the famous work of \citep{carpenter1990one} on RPM, the vast majority of AIG works would only generate 3$\times$3 matrices. The 2$\times$2 items have largely been neglected in the AIG works of matrix reasoning.
Secondly, the answer set contains two more meta-choices ``no correct answer'' and ``I don't know'', which encourage subjects to solve the items more constructively rather than eliminating responses. Thirdly, the variation of one element could depend on the variation of another element. For example, the dot's moves depend on the variation of the main shape, since the dot only moves along the polyline in the main shape. This kind of variation rule is rare in matrix reasoning items, but common in real-world problem-solving, and it represents an extra complexity factor of matrix reasoning.

Last but not least, IMak used a rule-dependent strategy to generate incorrect answer choices. For 1-rule items, 4 distinct values of the attribute of the rule are sampled, including the correct value; since all other attributes remain constant in the matrix, another random attribute is chosen and sampled for 2 values. The resulting 8 (4$\times$2) combinations make the 8 options in the answer set. For 2-rule items, 3 values are sampled for each of the 2 attributes of the 2 rules, resulting in 9 combinations, and one of them is discarded. For 3-rule items, 2$\times$2$\times$2 combinations are sampled in the same way. For 4-rule items, 2$\times$2$\times$2$\times$2 combinations were sampled in the same way, and half of them are discarded.

In a human experiment, 23 generated items were administered to 307 participants from Germany, Indonesia, and Argentina.
Reliability, validity and unidimensionality were initially verified by the experiment results. Particularly, item difficulty could be partly predicted from the number and type of rules based on psychometric models. 
As a summary, the open source software is a more recommended way to publish AIG works, especially for research purpose, as it can be shared across research groups around the world. More importantly, the studies should not be restricted to a fixed set of items but the way the generator is designed.


\subsubsection{Algorithmically Generating Matrix Reasoning Items for AI Testing}

We now review two AIG works of matrix reasoning that were specially for AI testing. The datasets generated in these two works are extremely influential on the data-driven AI models  for solving RPM-like tasks because almost all of them were tested on one or both of these two datasets. In addition, we also review the works that address the context-blind issue of the algorithmically generated datasets reviewed, which is a special and important issue for data-driven AI models.


\paragraph{Procedurally Generated Matrices}
Based on the five rules in \citep{carpenter1990one}, \citet{barrett2018measuring} continued the first-order logic approach of \citet{wang2015automatic} and created a large-scale (1.2M items) dataset of matrix reasoning items---Procedurally Generated Matrices (PGM). Since the generator program and source code are not publicly available, our discussion is based on the description in  \citep{barrett2018measuring} and our observation of the dataset.

\begin{figure}[ht]
    \centering
    \includegraphics[width=\textwidth]{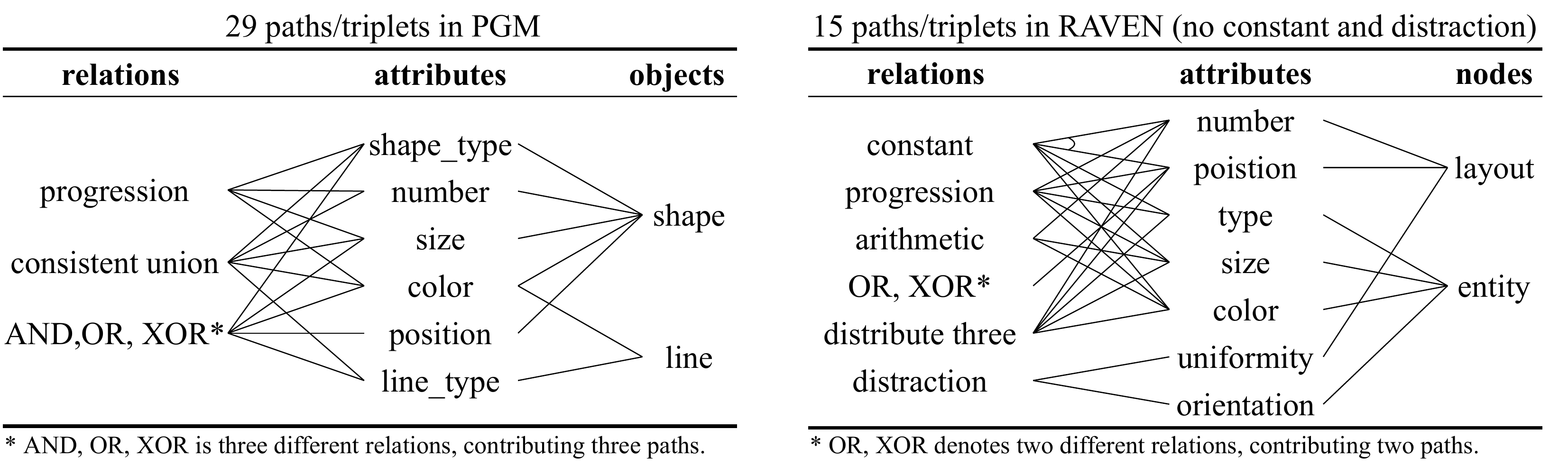}
    \caption{Left:  The dependencies among relations, objects, and attributes used to generate the Procedurally Generated Matrices (PGM) dataset \citep{barrett2018measuring}. Each path from left to right corresponds to a $[r, o, a]$ triplet representing a variation pattern in the matrices. As one can check, there are 29 paths, i.e. $[r, o, a]$ triplets, in the graph. Note that \citet{barrett2018measuring} did not differentiate between ``shape\_type'' and ``line\_type'' and referred to both of them as ``type''. But these two are treated as two distinct attributes in PGM's implementation.  Right: The dependencies among relations, nodes, and attributes used to generate the RAVEN dataset.  Note that we listed ``distraction'' as a rule in this graph to indicate that uniformity and orientation are distracting attributes. The paths from constant through number and position to layout are treated as a single rule in RAVEN's implementation. 
    Therefore, there are 15 paths/rules in the graph.}
    \label{fig:roa}
\end{figure}

In PGM, an instantiation of Equation \eqref{eq-variation-pattern} and \eqref{eq-constraint} in the first-order logic approach was denoted by a triplet $[r, o, a]$ of relation $r$, object $o$ \footnote{Geometric elements in the AIG works for human intelligence tests are commonly referred to as goemetric ``objects'', or objects for short, in AI works. We thus use the term ``object'' in the discussion of AIG works for data-driven AI models.} and attribute $a$. These three factors are not independent. Particularly, Figure~\ref{fig:roa} summarizes their dependencies in the generator of PGM. Figure~\ref{fig:roa} consists of 29 paths from the left to the right, corresponding to 29 $[r, o, a]$ triplets\footnote{This number---29---equals the number of triplets mentioned in the work of \citet{barrett2018measuring}, which, however, did not provide a list of the 29 triplets. Therefore, we could only conjecture that the 29 triplets here are the ones used in PGM.}. 

As shown in Figure \ref{fig:roa}, the objects in PGM are classified into two disjoint subsets---shape and line. In the shape subset, closed shapes are arranged in 3$\times$3 grid (fixed positions in this case) inside each matrix entry (do not mistake this with 3$\times$3 matrices). In the line subset, line drawings spans the whole area of a matrix entry and are always centered in the matrix entry. A PGM item can include both shapes and line drawing, with the shapes superimposed on the line drawings, but the reasoning about these two are completely independent. Thus, in Table \ref{tab:tech-details}, we split PGM into two rows to describe it more clearly.

The generation procedure of a PGM item could be described by 5 steps: (a) sample 1 to 4 triplets from the 29 triplets described in Figure \ref{fig:roa} (number triplets and position triplets can not be selected simultaneously);
(b) determine the analogical direction for each triplet: row or column; (c) sample attribute values for each triplet from their domains (Different sampling methods are specifically implemented for different rules and attributes); 
(d) determine the attribute values for unspecified attributes (either constant or random); and (e) render all attribute values into a pixel image of the matrix.

The relations used in the PGM dataset, which are also referred to as rules in other literature, stem from the 5 rules of APM summarized in \citep{carpenter1990one}, as follows:
\begin{itemize}[nolistsep,noitemsep]
    \item Constant in a row.
    \item Quantitative pairwise progression.
    \item Figure addition or subtraction, i.e. the set union and set diff (not arithmetic addition and subtraction), which could also be considered as the logical operator OR and XOR. 
    \item Distribution-of-three-values, i.e. the consistent union.
    \item Distribution-of-two-values, i.e. the logical operator XOR.
\end{itemize}
Comparing the PGM relations to the above rules, we found that they are almost equivalent. The ``constant in a row'' corresponds to the without-distraction mode in PGM. The ``distribution-of-three-values'' corresponds to the consistent union in PGM.
The ``figure addition or subtraction'' and ``Distribution-of-two-values'' are logical operator OR and XOR in PGM. However, the PGM has one more relation---AND---in addition to the 5 rules in \citep{carpenter1990one} to be more complete. 



\paragraph{Relational and Analogical Visual rEasoNing}

\begin{figure}[ht]
    \centering
    \includegraphics[width=\textwidth]{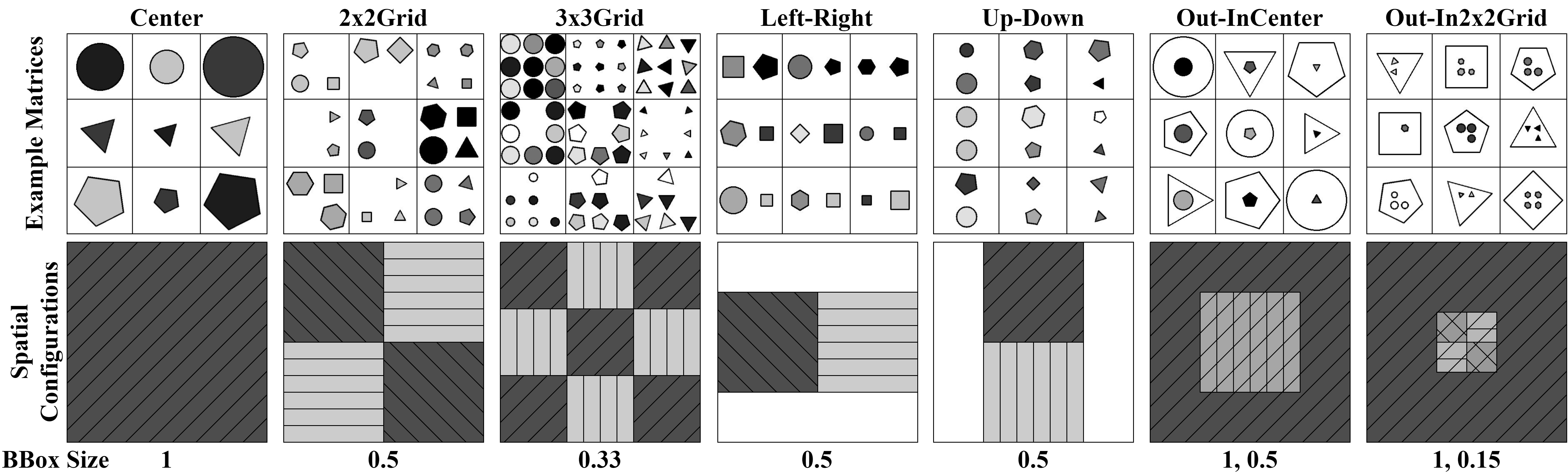}
    \caption{7 hardcoded spatial configurations---center, 2x2Grid, 3x3Grid, Left-Right, Up-Down, Out-InCenter, and Out-In2x2Grid---are used to arrange objects in each matrix entry in the RAVEN dataset. Each configuration is represented by the bounding boxes that objects could occupy. The position and size of each bounding box are hardcoded in the generator program. An example matrix for each configuration is given in the first row (image obtained by running the generator code). Note that not every bounding box has to be occupied, but every object has to be in one of the bounding boxes.}
    \label{fig:7-configs}
\end{figure}

The spatial configuration, as an important dimension of perception organization, is highly restricted in PGM---3$\times$3 grid for the shape subset, all-centered for the line subset, and superimposing a shape item on a line item. To enrich the spatial configuration of AIG of matrix reasoning, \citet{zhang2019raven} developed a new generator and generated the Relational and Analogical Visual rEasoNing (RAVEN) dataset. In particular, RAVEN includes 7 hardcoded spatial configurations, as shown in Figure \ref{fig:7-configs}. The source code of RAVEN's generator is available online\footnote{\url{https://github.com/WellyZhang/RAVEN}}. The discussion of RAVEN is thus based on the inspection of the RAVEN's generator's source code.

\begin{figure}[ht]
    \centering
    \includegraphics[width=\textwidth]{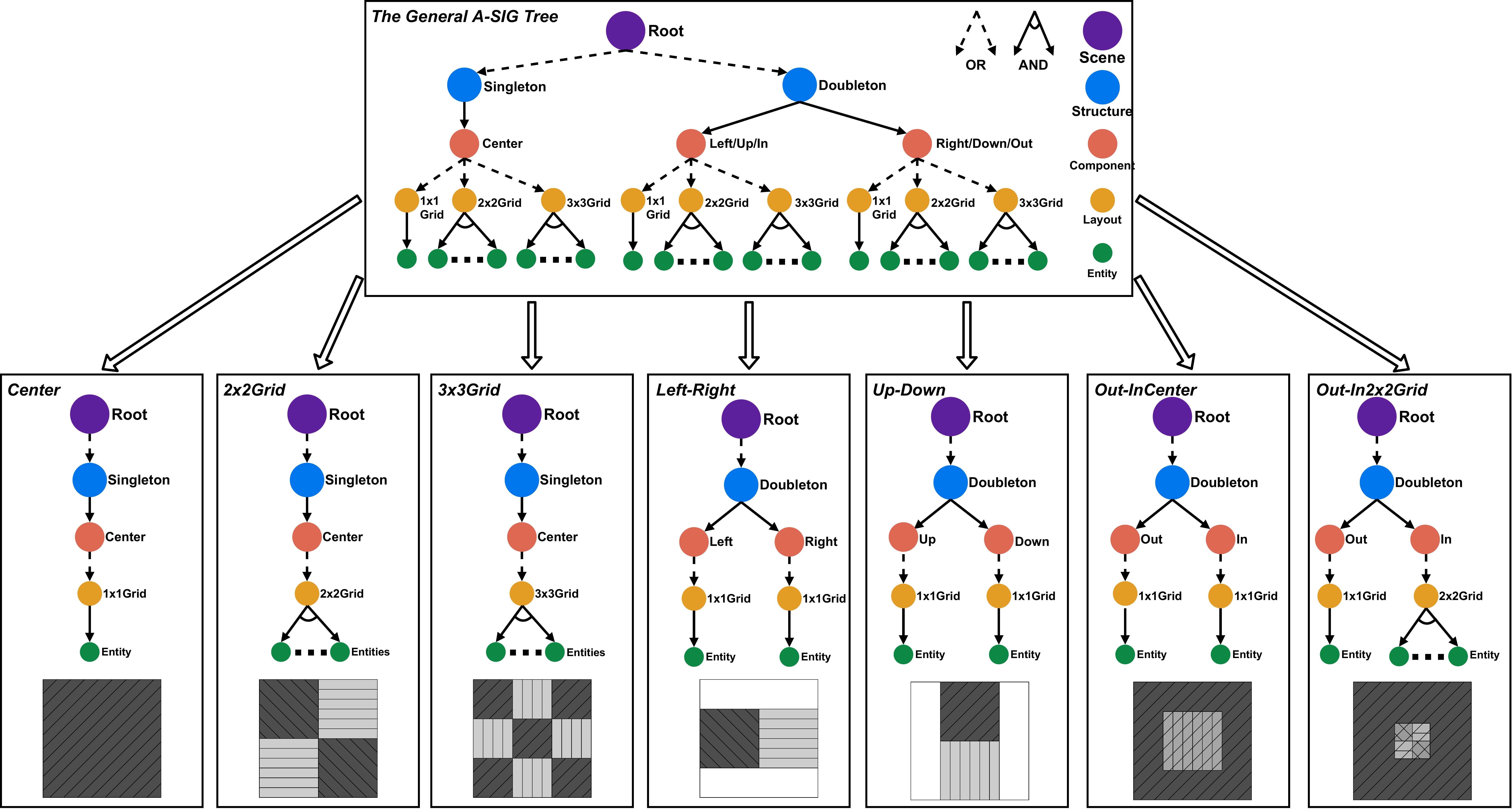}
    \caption{The general A-SIG tree and 7 specific A-SIG trees used in the RAVEN dataset (image adapted from \citep{zhang2019raven} by adding more technical details from the source code of the generator). The root node denotes the scene that the image describes. The structure nodes are the containers of different spatial structures. A structure is composed of components that could be overlaid with each other. Each component has its own layout and, more importantly, variation rules, which are independent of other components. The layout node, as its name indicated, contains the attributes specifying the number and positions of geometric objects. Entities represent geometric objects with attributes, not including number and position.}
    \label{fig:A-SIGs-RAVEN}
\end{figure}

The 7 configurations are derived from a more general symbolic representation framework for images---Attributed Stochastic Image Grammar (A-SIG). In A-SIG, an image is described by a tree structure, where the conceptual granularity becomes finer and finer from root toward leaves. To generate RAVEN, the tree structure is predefined as a general A-SIG tree as shown in Figure \ref{fig:A-SIGs-RAVEN}, which consists of 5 conceptual levels---scene, structure, component, layout, and entity---and uses a stochastic tree-traversal process to generate images. In general, the main idea of an A-SIG tree is that, while traversing the tree, if the current node has dashed edge to its child nodes, then expand a single random child node; if the current node has solid edge to its child nodes, then expand all its child nodes. Attributes and their attribute value domains are attached to nodes so that images can later be generated by sampling from these domains after the tree structure is determined. Such a stochastic traversing process from the root to leaves would generate a skeleton of a class of images---i.e. a spatial configuration. However, the 7 configurations in RAVEN were hardcoded in the language of A-SIG, rather than generated through this stochastic traversing process, which could otherwise have made RAVEN more diverse in spatial configuration.

\begin{figure}[ht]
    \centering
    \includegraphics[width=\textwidth]{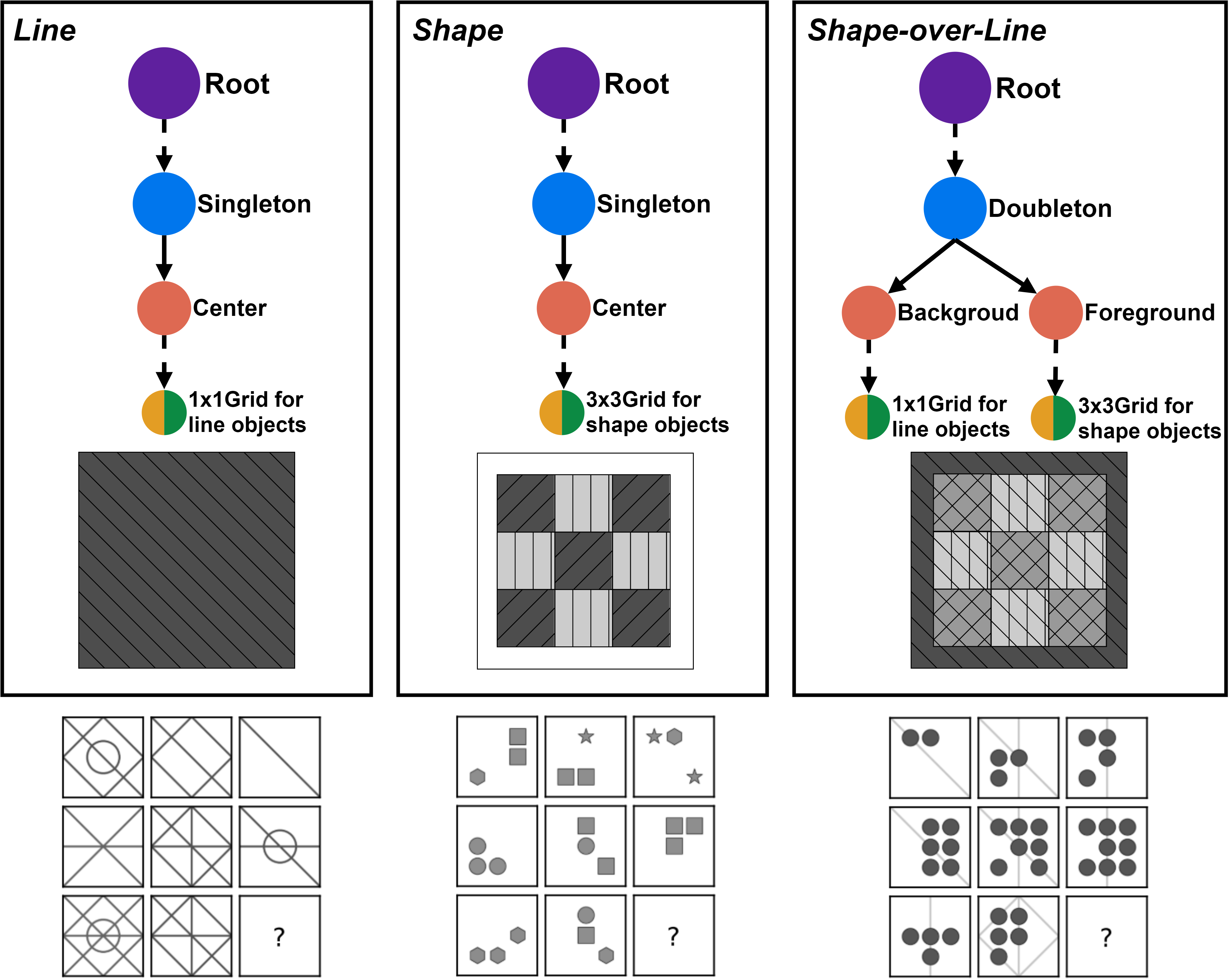}
    \caption{The spatial configurations of PGM represented in A-SIG to compare with RAVEN. There are 3 spatial configurations in PGM---line, shape, and shape-over-line---in PGM. The example matrix is given for each configuration at the bottom (images taken from PGM dataset).}
    \label{fig:A-SIGs-PGM}
\end{figure}

To compare with the PGM dataset, we represent PGM items also in A-SIG, as shown in Figure~\ref{fig:A-SIGs-PGM}. The line configuration of PGM is basically the same as the center configuration of RAVEN except that the entity types (shape) are different. The shape configuration of PGM is almost the same as the 3x3Grid configuration of RAVEN except that bounding box sizes are slightly different. The shape-over-line configuration of PGM is also conceptually similar to the double-component configurations of RAVEN. The general difference between PGM and RAVEN lies in the layout and entity nodes. As shown in Figure~\ref{fig:A-SIGs-PGM}, the PGM dataset is not able to separate the concepts of ``entity'' and ``entity layout'' by using triplets $[r, o, a]$. That is, the object $o$ takes the roles of both layout and entity nodes, but could not play the roles effectively and simultaneously.

RAVEN inherited all the five rules from \citep{carpenter1990one}. Moreover, the ``addition-and-subtraction'' rule is extended in RAVEN containing not only figure addition and subtraction (i.e., the set operations ``OR and XOR'') but also arithmetic addition and subtraction, which were not discussed in \citep{carpenter1990one}. Since these two operations are conceptually different, we refer to the arithmetic addition and subtraction as ``arithmetic'', and the figure addition and subtraction as ``OR and XOR''. In addition, the ``distribution-of-three-values and distribution-of-two-values'' from \citep{carpenter1990one} are merged into a single rule in RAVEN by considering the latter as a special case of the former with a null value for one of the three values. Therefore, RAVEN has a slightly different rule set compared to PGM. Similarly, we could represent the variation rules of RAVEN also as triplets --- $[r, n, a]$ where $n$ represents nodes (layout or entity) in A-SIG trees, and $r$ and $a$ are relations and attributes, being the same as PGM. Then Figure \ref{fig:roa} shows the dependencies among $r$, $n$ and $a$. 


PGM and RAVEN generators are similar in some aspects. In particular, they share two similarities. First, their choices of attributes, attribute domains, and rule types are similar. For example, they both forbid number-rule and position-rule from co-occurring in an item because these two attributes would probably conflict with each other. Second, although RAVEN has more spatial configurations, these configurations are not structurally different from PGM (as can be seen from the comparison of their A-SIG trees). 
Meanwhile, PGM and RAVEN are different in two aspects. First, they are different in the number of rules in an item. In PGM, 1 to 4 triplets were sampled from the 29 triplets. In contrast, in a RAVEN item, every attribute is governed by a rule except the two distracting attributes (uniformity and orientation). Thus, there are 4 rules (for number/position, type, size, and color, respectively) in each RAVEN item. Second, the rules in RAVEN are all row-wise while the rules in PGM are either row-wise or column-wise.


\paragraph{Context-Blind Issue}

The answer sets in RAVEN were generated in a similar way to the way in the first-order logic approach. That is, each incorrect answer choice is created by modifying a single attribute of the correct answer. RAVEN is slightly different from \citep{wang2015automatic} because RAVEN has only 5 attributes (not including the distracting attributes) whereas \citep{wang2015automatic} has 15 attributes. Hence, in \citep{wang2015automatic}, every incorrect answer has a unique attribute on which it differs from the correct one; but RAVEN has to reuse some of the 5 attributes to generate 7 incorrect answers, i.e. an attribute is given different values to generate multiple incorrect answers. 

This method of creating incorrect answer choices reaches the maximum level of distracting and confusing effect, because one must identify all the variation rules to solve the problem. On the contrary, ignoring any rule would lead to multiple choices. However, this design has a major drawback---it fails the context-blind test for multi-choice problems. In a matrix reasoning item, the incomplete matrix is the context of the multi-choice problem that provides information for solving the problem. Failing the context-blind test means that it is possible for human participants or computational models to solve the item while turning blind to the context.

Two works \citep{hu2021stratified, benny2021scale} separately pointed out the context-blind issue of RAVEN. They provided evidence that data-driven AI models can achieve high accuracies (from 70\%+ to 90\%+) when only given access to the answer sets of RAVEN. The context-blind performance of some data-driven AI models is even better than the normal performance with full access to the items. This implies that data-driven AI models are capable of capturing the statistical regularities in the answer sets. The reason for this context-blind issue obviously lies in the generating process of answer set. In particular, since each incorrect answer choice is a variant by modifying a single attribute of the correct answer choice, the correct answer must be the one that possesses every common feature among all the choices (or, equivalently, the one most similar to every other choice).

\begin{figure}[ht]
    \centering
    \includegraphics[width=0.7\textwidth]{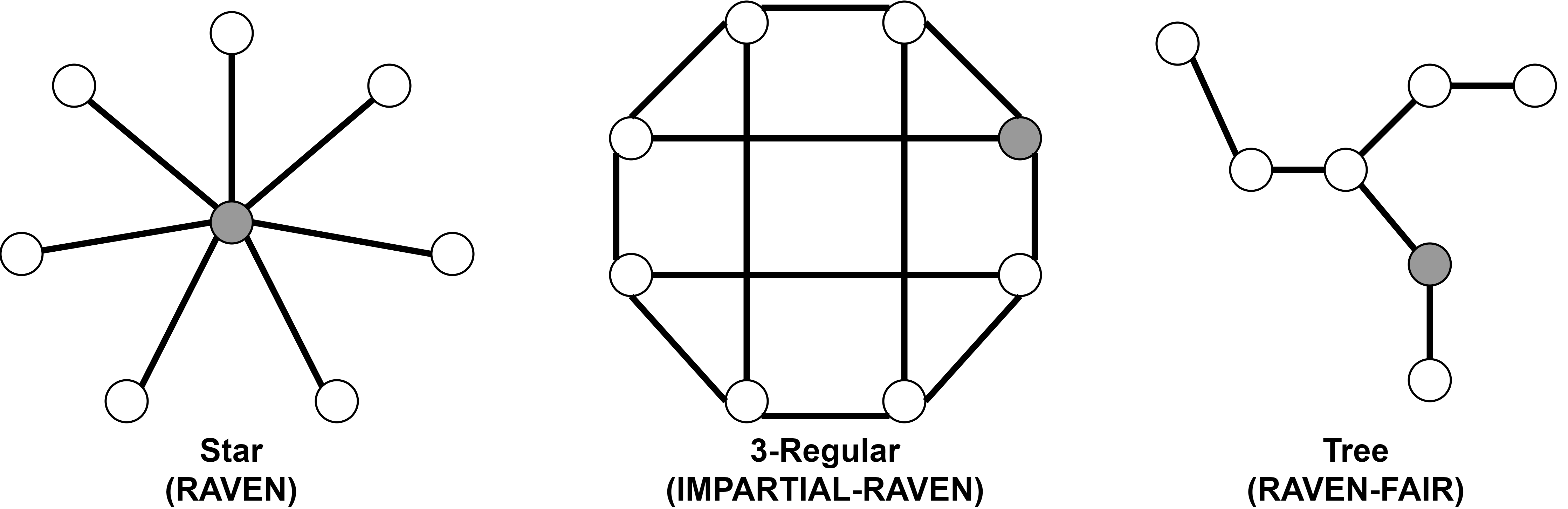}
    \caption{The answer sets of three versions of RAVEN datasets depicted in graphs. Each vertex is an answer choice and two adjacent vertices differ by one attribute.}
    \label{fig:3-RAVEN-answer-generation}
\end{figure}

Both \citet{hu2021stratified} and \citet{benny2021scale} proposed their own solutions to this issue---the Impartial-RAVEN and RAVEN-FAIR datasets. These two datasets have the same context matrices as the original RAVEN and regenerated the answer sets in different ways. The similarity and difference between these three versions can be clearly illustrated by putting them in simple graphs. If we represent each answer choice as a vertex and each modification of an attribute as an edge, then the answer sets of the three versions can be depicted by the graphs in Figure \ref{fig:3-RAVEN-answer-generation}. The answer set of the original RAVEN is created by modifying an attribute of the correct answer. Thus, its graph is a star centered at the correct answer (the solid vertex). 
And what the aforementioned computational models in the context-blind test captured was the unique center of the star structure.

\citet{hu2021stratified} proposed the Impartial-RAVEN, in which the answer set can be represented by a 3-regular graph in Figure \ref{fig:3-RAVEN-answer-generation}. To create such a graph, three independent attributes are randomly chosen from the five attributes of RAVEN, and three values of the three attributes are sampled from the three attribute value domains, respectively, so that the newly sampled values are different from the ones of the correct answer. Then, by assigning new values to these attributes combinatorially, we would have $2^3=8$ answer choices, including the correct one. The relations among these 8 answer choices form the 3-regular graph in Figure \ref{fig:3-RAVEN-answer-generation}. 

\citet{benny2021scale} proposed a less regulated procedure to generate answer sets. Starting from an initial answer set consisting of only the correct answer, an answer choice is randomly selected from the current answer set, and then an attribute of the selected answer choice is randomly altered to alter to create a new answer choice; repeat this process until we have 8 answer choices. This procedure results in tree structures similar to the one in Figure \ref{fig:3-RAVEN-answer-generation}. 

These two enhanced versions of RAVEN were tested by context-blindly training the baseline model in \citep{zhang2019raven} and the CoPINet model in \citep{zhang2019learning}. The accuracy decreased to below 20\%. Ideally, a human subject or computational model who context-blindly works on the RAVEN items should perform as well as a random guess, i.e. $1/8=12.5\%$, which implies that the answer set per se does not provide any useful information for solving the item. However, in the practice of item writing, to maintain a certain level of distracting and confusing effect of incorrect answer choices, the majority of incorrect answer choices must share some similarities among themselves, with the correct one, and the context matrix, which would raise the performance of random guess a bit. On the flip side, without this design, it would be quite easy for subjects to find the correct answer, because incorrect answers would be very much perceptually distinct from the context and other answer choices. Therefore, a reasonable context-blind performance would be slightly higher than random guess. The balance is determined by the item writer's judgment.

A subtle difference between the two enhancements of RAVEN could be found by comparing their graphs in \ref{fig:3-RAVEN-answer-generation}. If we consider a single trial (in a probabilistic sense) where we context-blindly give a participant (or an AI model) an item from Impartial-RAVEN and an item from RAVEN-FAIR, the probability that this participant solves the Impartial-RAVEN item would be almost the same as the probability of solving the RAVEN-FAIR item. However, if we repeat this with different items again and again, the performance on RAVEN-FAIR would probably exceed the performance on Impartial-RAVEN, assuming that the participant is intelligent enough to figure out the graph structures behind the answer sets, and thus makes an educated guess by selecting the ``center'' (or the max-degree vertex) of trees in a probabilistic sense. In this case, we would say that the RAVEN-FAIR is context-blind valid at the item level, but not at the dataset level.

\subsubsection{Summary}

In this subsection, we reviewed AIG works of matrix reasoning items. We classified the works into two groups by their purposes---whether it is for human intelligence testing or for AI testing. The works in the first group aim at not only generating items but also good psychometric properties. As the classical studies on intelligence tests, these works are usually based on cognitive models and psychometric models. The choices of stimulus features are thus determined by the cognitive and psychometric models. Particularly, the factors---the number of elements, the number of rules, the type of elements, types of rules, analogical directions, and perceptual organizations---are usually considered in this line of research. Among these factors, the types of elements and rules and perceptual organization are the less investigated ones due to the difficulty in defining and formalizing them.

The works in the second group can be seen as the continuation of the first group, but the psychometric aspects are less emphasized. For example, in an human experiment of PGM, in which 18 items were administered to human participants, participants without prior experience failed almost all the items, whereas participants with prior experience scored above 80\%. Such a result is definitely not what a psychometrician would expected from a test for eductive ability, fluid intelligence, or general intelligence. In contrast, the result appear to be a result from a test of reproductive ability or crystallized intelligence. Generally speaking, this result implies that the datasets for AI testing do not necessarily qualify for human intelligence testing.

More importantly, this gives rise to another interesting question---how do we assess the performance of data-driven AI models on the large datasets such as PGM and RAVEN? On one hand, some data-driven AI models indeed perform well on AIG items that pose great challenges to human subjects; on the other hand, training on the large-scale datasets specially prepares the AI models for a highly restricted subset of the problem domain, but human subjects, who are not trained at all, or just trained on several examples from this subset, could perform well in the entire problem domain.

Similar questions were asked when AI systems first entered the area of human testing \citep{detterman2011challenge}. Efforts have been made to address these questions. \citep{bringsjord2003artificial, bringsjord2011psychometric} address this issue by incorporating AI testing into a general concept---psychometric AI. \citet{hernandez2016computer} proposed that (a), instead of collecting items, we should collect item generators, and (b) the generated items should be administered to machine and human (and even other animals) alike (universal psychometrics). All these propositions are constructive and, meanwhile, suggest much higher requirements for AIG studies.

Current AIG datasets are far below the level of flexibility and diversity that human item writers can achieve. For example, the spatial configurations in PGM and RAVEN are fixed; inter-element variation, in which the variation of one element depends on the variation of another element, is also very rare; so are perceptually and conceptually ambiguous analogies. A more promising methodology for AIG of RPM-like tasks for AI testing is to study the problem domain and human cognition, rather than construct ad hoc generator programs. Huge uncharted territories lie in the complexity factors such as the types of elements and rules and perceptual organization, and how the nature of problem changes as different administration/evaluation protocols are used for human subjects and AI models.

\section{Computational Models for Solving RPM and RPM-Like Tasks}
\label{sec:com-mod}

In previous sections, we have established the basic understanding of the problem domain represented by RPM, which lays the foundation for us to discuss the core topic of this article---computational models for solving the problems. Similar to the way in the previous discussion, we start from the origin of the research, keep the prerequisite knowledge at a minimal level, and unfold our discussion in a manner that reveals the philosophy behind technical development in simplest language. 

The ultimate purpose of this section is to help our readers develop a solid understanding rather than enumerating as many previous works as possible in chronological order or in an arbitrary taxonomy. Therefore, we use a narrative, which simulates the process of how an novice's understanding of the solution to the problem domain would naturally evolve if not influenced by the external conditions (such as computational power) and other relevant research works. This narrative is not real history but specially designed to reduce the complexity for understanding. In particular, the computational models that arise late in this narrative might arise early in reality, and vice versa. 
Examples like this are common in scientific research: the original concepts behind some cutting-edge technologies might have been there for decades before these technologies are implemented, but some alternatives to the original concepts, due to being easy to implement, might have already been implemented before the cutting-edge technologies; when we look back at these concepts, we rearrange the order to make the concepts more coherent and understandable. Thus, this narrative is a conceptual chronicle for understanding rather than a real chronicle for recording.

In this conceptual chronicle, we divide the development of computational models for solving RPM into five stages---imagery-based approach, logical reasoning, neuro-symbolic reasoning, learning approach, and data manipulation. With hindsight, we found that an upward-spiral pattern is looming out of these five stages. That is, researchers are making process while visiting the same places again and again with better and better understanding. The places could be specific research questions or a type of approach to answer the research questions. The conceptual chronicle starts from a straightforward approach (imagery-based approach) which is specific to the problem domain but very effective; it then moves on to more and more general approaches (logical reasoning, neuro-symbolic reasoning, and learning approach); when these approaches is still incapable of solving the problem domain perfectly, it returns to the study of the problem domain per se and solves the problem in a similar way to the first approach, but uses completely different set of techniques. The same upward-spiral trajectory could be described differently (e.g., different methodologies are alternatively dominating the research of intelligence), but the pattern that it revisits the same places again and again until the entire problem domain is perfectly solved remains unchanged.

In the rest of this section, we will use the acronyms of computational models for simplicity and please refer to Table~\ref{tab:t1} for their full names. 


\newpage
\begin{ThreePartTable}

\begin{TableNotes}[flushleft, para]
\item [*] No Acronym was given in the original article. We created a name to clearly refer to it in our discussion.
\end{TableNotes}
\centering

\scriptsize
\begin{xltabular}{\textwidth}{p{0.2\textwidth} p{0.5\textwidth} p{0.3\textwidth}}

\caption{Computational Models for Solving RPM and RPM-like Tasks}\label{tab:t1}\\

\toprule
Acronym & Full Name & Article \\
\midrule
\endfirsthead

\caption[]{(continued from previous page)} \\
\toprule
Acronym & Full Name & Article  \\
\midrule
\endhead

\bottomrule
\multicolumn{3}{r}{\footnotesize to be continued on the next page}
\endfoot

\bottomrule
\insertTableNotes
\endlastfoot

- & Gestalt Algorithm & \citep{hunt1974quote}\\
ASTI & Affine and Set Transformation Induction & \citep{kunda2013visual} \\
ASTI+ & Affine and Set Transformation Induction Plus & \citep{yang2020not} \\
- & Fractal Model & \citep{mcgreggor2014fractals} \\
- & FAIRMAN & \citep{carpenter1990one} \\
- & BETTERMAN & \citep{carpenter1990one} \\
CogSketch+SME & CogSketch and Structual Mapping Engine & \citep{lovett2009solving} \\
- & Anthropomorphic Solver & \citep{strannegaard2013anthropomorphic} \\
- & ANALOGY & \citep{evans1964program}\\
- & Analytic Algorithm & \citep{hunt1974quote}\\
ALANS2 & ALgebra-Aware Neuro-Semi-Symbolic & \citep{zhang2020learning} \\
PrAE & Probabilistic Abduction and Execution & \citep{zhang2021abstract} \\
VAE-GPP & Variational Autoencoder and Gaussian Process Priors & \citep{shi2021raven} \\
TRIVR & Two-Stage Rule-Induction Visual Reasoning & \citep{he2021one}\\
NVSA & Neural-Vector-Symbolic Architecture & \citep{hersche2022neuro} \\
Pairwise-ADV\tnote{*} & Pairwise Attribute Difference Vector & \citep{Mekik2017deep}\\
Triple-ADV\tnote{*} & Triple Attribute Difference Vector & \citep{mekik2018similarity}\\
DeepIQ & Deep IQ & \citep{mandziuk2019deepiq}\\
CNN+MLP & - & \citep{hoshen2017iq} \\
CNN+decoder\tnote{*}	& - & \citep{hoshen2017iq} \\
ResNet+MLP & - & \citep{barrett2018measuring} \\
Wild-ResNet+MLP & - & \citep{barrett2018measuring} \\
WReN & Wild Relation Network & \citep{barrett2018measuring} \\
LEN & Logic Embedding Network & \citep{zheng2019abstract} \\
MXGNet & Multiplex Graph Network & \citep{wang2020abstract}\\
multi-layer RN & multi-layer Relation Network & \citep{, jahrens2018multi, jahrens2019multi, jahrens2020solving} \\
SRAN & Stratified Rule-Aware Network & \citep{hu2021stratified} \\
MRNet & Multi-Scale Relation Network & \citep{benny2021scale} \\
Rel-Base & Basic Relational Reasoning & \citep{spratley2020closer} \\
Rel-AIR & Attend-Infer-Repeat Relational Reasoning & \citep{spratley2020closer} \\
CNN+LSTM+MLP & - & \citep{barrett2018measuring} \\
Double-LSTM & - & \citep{sekh2020can} \\
ESBN & Emergent Symbol Binding Network & \citep{sinha2020memory} \\
NTM & Neural Turing Machine & \citep{sinha2020memory} \\
ARNe & Attention Relation Network & \citep{hahne2019attention} \\
HTR\tnote{*} & Hierarchical Transformer Reasoning & \citep{an2020hierarchical} \\
NI & Neural Interpreter & \citep{rahaman2021dynamic} \\
SCL & Scattering Compositional Learner & \citep{wu2021scattering} \\
4 VAE+WReN & 4 variants of VAE plus WReN & \citep{steenbrugge2018improving, van2019disentangled} \\
generative-MRNet\tnote{*} & - & \citep{pekar2020generating} \\
LoGe & Logic-Guided Generation & \citep{yu2021abstract} \\
MCPT & Multi-label Classification with Pseudo Target & \citep{zhuo2020solving} \\
PRD & Pairwise Relations Discriminator & \citep{kiat2020pairwise} \\
LABC & Learning Analogies by Contrasting & \citep{hill2019learning} \\
CoPINet & Contrastive Perceptual Inference Network & \citep{zhang2019learning} \\
DCNet & Dual-Contrast Network & \citep{zhuo2021effective} \\
ACL & Analogical Contrastive Learning & \citep{kim2020few} \\
Meta-ACL & Meta Analogical Contrastive Learning &  \citep{kim2020few} \\
MLCL & Multi-Label Contrastive Learning & \citep{malkinski2020multi} \\
FRAR & Feature Robust Abstract Reasoning & \citep{zheng2019abstract} \\
- & Continual Learning & \citep{hayes2021selective} \\
DRT &  Dynamic Residual Tree & \citep{zhang2019raven} \\
- & GAN & \citep{hua2019modeling} \\
- & Structural Affinity Method & \citep{shegheva2018computational}\\
PGM & Procedurally Generated Matrices & \citep{barrett2018measuring} \\
RAVEN & Relational and Analogical Visual Reasoning & \citep{zhang2019raven}
\end{xltabular}

\end{ThreePartTable}

\subsection{Stage 1: Imagery-Based Approach}

Visual mental imagery refers to mental images that play a functional role in human cognition\citep{kosslyn2006case}. The most important characteristic of mental imagery is that human can experience mental imagery in the absence of the concurrent sensory input. Try to answer this question ``how many windows are there in your house?'' when you are not home (use another building you are). Most people answer this question by imagining their houses. This imaginary house is a mental imagery. Some people count the windows by mentally walking in and around their houses, while others do so by mentally rotating their houses. Whether walking in and around the houses or rotating the horse, they inspect and manipulate on this mental representation, as they inspect and manipulate the real object. Further more, mental imagery can be unrealistic, for example, some people rotate their houses upwards or downwards without the houses falling apart. For this reason, the ability of using mental imagery is important for creativity. This point makes another important characteristic of mental imagery. 


Evidence from psychology and neuroscience \citep{kunda2013computational} suggests that mental imagery is frequently used by human participants to solve RPM items. Intuitively, a human participant would inspect objects in the matrix, compare them by mentally superimposing one on another, mentally transform the objects, and mentally estimate perceptual similarity. Without turning to more sophisticated techniques and terminology, this description is the most immediate one that one can think of to describe the solving process (although they might not use the term ``mental imagery''). For this reason, imagery-based computational models  \citep{hunt1974quote, kunda2009addressing, kunda2013computational, kunda2010taking, yang2020not, yang2022end} 
were constructed to solve RPM and RPM-like items. In general, these models represent matrix entries by pixel images, apply predefined pixel-level operations on the images (e.g., affine transformations and set operations) and calculate pixel-level similarities between the images (e.g., Jaccard index and Hausdorff distance). 

Although these systems have proven to be effective for solving the RPM items, they appear relatively late in development of computational models for solving RPM and are still an underexplored approach in the AI community. This is partly because directly working on raw perceptual input data, especially applying various operations on pixel images and computing similarity, requires the computational power that was not available at the beginning of this line of research. Another reason might be that the theory of mental imagery has been studied mostly in cognitive psychology and received less attention in the AI community. Nonetheless, it is still a promising approach for general problem solving.

Before we dive into Stage 2, it would be better to chew on the idea of imagery-based approach a bit. The imagery-based approach provides an ``in-place'' solution, i.e., solving a visual reasoning problem ``visually'' without introducing auxiliary devices such as preprocessing of raw perceptual input. There is nothing wrong for being parsimonious because being parsimonious is a general principle of problem solving (Occam's Razor). On the flip side, a tacit consensus in artificial intelligence is that certain degree of abstraction is desirable. That is, the approach must include steps that transform the raw perceptual input into a more abstract form which reduces the complexity of problem solving. Abstraction is even deemed a hallmark of valuable AI techniques---the more abstract, the more intelligent the approach is. According to this criterion, the imagery-based approach is not intelligent at all. This could be another reason that imagery-based approach received less attention in the AI community. Because experiments also show that mental imagery plays an important role in human cognition, this brings us into a dilemma of the criteria of being intelligent in problem solving. Note that the ``abstract'' end of this dilemma is not proficiency in using the abstracted information but the process or ability to abstract.

The most valuable contribution of imagery-based models is not on problem solving but bringing this dilemma into light. Being mindful of this dilemma and criteria of being intelligence would put future AI systems in a more promising direction. Although this dilemma is of vital importance to AI research, there is not simple answer to it. A possible solution is that we can try to understand imagery and abstraction as two factors, correlated or independent, rather than two options contradicting to each other. A simple analogy could be made at this point to clarify this idea: in graduate math classes, instructors are undoubtedly teaching knowledge that are quite abstract; experienced instructors are able to convey the abstract knowledge in very visual languages so that it is more accessible to students. A stronger claim is that abstract concepts are always associated with some imagery representations in human thinking. This claim might not be correct in all cases, but indeed points out an important feature of human intelligence. This solution also resembles the psychometric treatment to human intelligence. 
AI systems can be evaluated in the both dimensions of the two factors, as human intelligence is measured. And being intelligent means that the system needs to score high in both dimensions. 

Another similar solution is to view abstraction and mental imagery as two distinct and necessary cognitive processes that complement and cooperate with each other. Which of them manifests depends on the task and the subject; that a subject does not show one of them does not mean that this subject does not possess it. Conditional arguments like this is quite common in the study of human intelligence. For example, mental information processing speed (measured by special tasks) is greatly correlated with general intelligence test performance of people who score lower in the test, while processing speed is not correlated with that of people who score higher. But this does not mean the highly intelligent people cannot think fast. Either way, the dilemma is resolved by stressing that these two options are not exclusive to each other.

\subsection{Stage 2: Logical Reasoning}

Based on the discussion at the end of the last subsection, the reason why we choose logical reasoning as the second stage in this conceptual chronicle is obvious. The computational models using logical reasoning works on abstract representations of RPM-like items. For example, a entry image $A$ in a matrix could be described by a series of propositions such as ``triangle($A$)=True, triangle-large($A$)=False, triangle-on-the-left($A$)=True, square($A$)=True, square-small($A$)=True, and so on''. In this example, these representations are restricted to Boolean expression, but we can use more expressive formal logic like ``color($A$)=green, number-objects($A$)=3, texture($A$)=dotted, and so on''. The abstract representations in these models are either manually constructed or obtained through a preprocessing module. For example, the earliest computation model for solving RPM-like items---ANALOGY \citep{evans1964program}---consists of two modules and first part is for constructing such representations\footnote{Interesting anecdote about Evans' work: because the memory of computer at the time was so limited, the program had to be separated into two modules which were executed serially. But the models in this stage and following stages are designed to be so.}, whereas the influential models---FAIRMAN and BETTERMAN \citep{carpenter1990one}---use handcrafted logic representations.

Each computational model in this stage has a customized formal system for representing RPM-like items. This system is either specially designed for solving a specific problem set of interest or reusing some standard systems such as regional connection calculus and scalable vector graphics. Based on a formal representation system, the three main components of logical reasoning are implemented. In the context of RPM, the entry images in the matrix are the premise and the answer choices are possible consequences, whereas the rules are to be determined. The models in this stages split into two branches according to how rules are determined.

\subsubsection{Rule Matching}
The first branch is rule matching, in which the model hardcodes finite predefined rules and matches rows and column to each of the predefined rules. For example, a predefined rule describing the number of objects could be ``number-objects($A$)+number-objects($B$)=number-object($C$)'', in which $A, B$ and $C$ are entry images in a row or column of a 3$\times$3 matrix. If a rule applies to the first row(s) or columns(s), it is reproduced on the last row or column to generate the formal representation of the missing entry. Many computational models have been constructed this way to solve RPM-like items  \citep{hunt1974quote, carpenter1990one, ragni2012solving, ragni2014analyzing}. 
This might look amazing from the current point of view because it simply would not generalize due to the predefined rules. However, this is not true from the perspective of problem solving. The readera who are skeptical about this can make analogies to other cases, like consider how many rules one need to derive the integer field and to derive the real number field, and consider also the expressive power of these number fields. The reason why these number fields can be represented concisely and completely is that when we discuss them in math, the symbols (i.e., elements in these fields) does not need to bind with concrete entities. In computational models of logical reasoning, the formal representation systems are in charge of binding symbols with entities. Thus, it is very partial to argue that rule matching models are not generalizable without referring to the formal representation system. If the geometric visual stimuli are extremely simple (as in most general intelligence tests) or the formal representation system is extremely powerful, the rule matching models will generalize as well to the whole problem domain as several rules can produce the entire integer field and real number field. This conditional argument echoes with the discussion about abstraction and imagery at the end of last subsection as they all are dealing with the problem of the level of abstraction and the operations that can be implemented at that level.

Another observation on the rule matching models is that most of the works in this branch are for cognitive modeling rather than problem solving. For example, the models in \citep{ragni2012solving, ragni2014analyzing} are implemented on the ACT-R cognitive architecture. The purpose of cognitive modeling based on cognitive architecture lies at the information-processing level, i.e., modeling how information is exchanged between multiple cognitive function modules. But how the information is processed exactly inside each module is not really the focus of cognitive modeling. This corresponds to how a human or computation model comes up with a rule that happens to be able to solve the item at hand. Thus, the use of predefined rules would be understandable from the perspective of cognitive modelling. As a summary, the rule matching approach, though might not be able to solve all possible items in the problem domain, fulfills its duty perfectly in problem solving and in cognitive modeling.

\subsubsection{Rule Induction} 
In contrast to rule matching, the second branch---rule induction---is mainly studied for problem solving \citep{bohan2000ludi, davies2001visual, ragni2007solving, schwering2007using, strannegaard2013anthropomorphic} with an exception that the analogy-making models are closely related to both cognitive modeling and problem solving \citep{tomai2005structure, lovett2007analogy, lovett2009solving}. Rule induction means that the models need to discover the rules, i.e., how an entry is transformed to the next one or how the entries in a row or column are related, in a more open manner. In particular, the rules are represented as the identical and different parts between the logical structures of entry images, and/or how the different parts are changed to identical parts by transformations. The rules are also logical representations in nature. After the rules are discovered, the models have two options---they can either reproduce the rules on the last row or column to generate the answer and compare it to the answer choices, or insert each answer choice into the last row or column to induct a rule for the last row or column and compare it to the previously inducted rules. These two options corresponds to the two strategies---constructive matching and response elimination---commonly adopted by human participants \citep{bethell1984adaptive}. The latter one is more often adopted by analogy-making models as it is similar to how an analogy is drawn by human.

Rule induction is a larger topic that goes beyond the traditional format of logic reasoning. For example, the goemetric objects and rules can be represented in a vector-symbolic architecture \citep{rasmussen2011neural}, in which geometric objects are represented as vectors and rules are inducted and applied through operations on the vectors, such as circular convolution. If we take a closer look at the details of calculation, we would find that the calculation is a different way to implement the rule induction in the models mentioned above (of course, vector-symbolic architecture has its own advantages and purposes). Another example is that reinforcement learning methods can be used to train an agent to induct the rules in matrix reasoning items \citep{raudies2017model}, i.e., when the agent forms a correct rule (action in reinforcement learning) when it attends to certain row or column in the matrix (state in reinforcement learning), the algorithm rewards the agent. 

The boundary between rule matching and rule induction is not always so clear in practice. To what extent a model is performing rule matching or rule induction depends on how the potential rules are provided to the model: if only several specific rules are provided, it is rule matching; if a huge rule space is provided by specifying some ``bases'' or ``generators'' of it, it is rule induction; and there is a lot of places in between. One can even argue that a rule induction model is in nature a rule matching model because it matches to the whole or a subspace of the rule space in some implicit way, and, thus, there is no such distinction between rule matching and rule induction. Nonetheless, there are indeed examples of rule induction that no one will consider as rule matching. For example, consider a free group in abstract algebra, in which finding an element satisfying a specific condition could be so difficult even if the generators of this free group look so simple. Other similar examples could be find in the problems of program synthesis and inductive programming.

\subsection{Stage 2.5: Neuro-Symbolic Reasoning}

The reason why we use a decimal in the title is that this stage is an intermediate stage that shares features with both its predecessor and successor. Since the influence of its predecessor and successor is stronger than this stage, this stage is relatively short and rapidly transits to its successor.

The models of neuro-symbolic reasoning consists of two modules---a neural perception frontend and a symbolic reasoning backend. The neural perception frontend (implemented as neural networks in most cases) extracts/approximates the distributions over the values of each entry in the predefined formal representation system. The symbolic reasoning backend performs probability calculation according to a predefined set of rules. In a sense, neuro-symbolic reasoning can be considered as a special case of rule matching in logical reasoning. The probability formulae in the backend are determined by the predefined rules and the output of the reasoning, such as the probability that a rule exist in rows or columns, the missing entry contains a certain value in its representation, or a certain answer choice is correct. Similarly, different implementations of frontend and backend have been used to construct probabilistic reasoning models, such as ALANS2, PrAE, VAE-GPP, TRIVR, LoGe, and NVSA \citep{zhang2020learning, zhang2021abstract, shi2021raven, he2021two, yu2021abstract, hersche2022neuro}.

Compared to logical reasoning, neuro-symbolic reasoning clearly requires that a dedicated neural processing module is used to construct the formal abstract representation of each entry image. In addition, it also takes into account the uncertainty in perception by using probability to represent and reason. Technically speaking, neuro-symbolic reasoning is only a small step forward compared to logical reasoning. The reason why it is listed as a separate stage is that it is a natural watershed between \textbf{knowledge-based approaches} and \textbf{data-driven approaches}, because the neural perception frontend requires training data while imagery-based and logical reasoning are knowledge-based. In the next two subsections, we will elaborate on data-driven approaches.

\subsection{Stage 3: Learning Approach}

An obvious characteristic of the first three stages is that they all rely on the predefined representation systems of geometric objects and relations/rules between geometric objects. To reduce the reliance on such explicit prior knowledge, the learning approach has been introduced into the field of RPM-like tasks. This section reviews the learning models, especially deep learning, for solving RPM-like tasks. We divide the learning approach into four types according to the structures of learning models. For each type, we provide a high-level functional description that applies to all the models of the type while trying not to complicate the discussion with too much technical details. The purpose of this taxonomy is to reveal the structural evolution of the learning models (from Type 1 to Type 4), analyze the reason why it evolves this way, and, more importantly, provide a guidance for research works in this field.

\subsubsection{Type 1}

\begin{figure}[!bthp]
  \centering
  \includegraphics[width=0.25\textwidth]{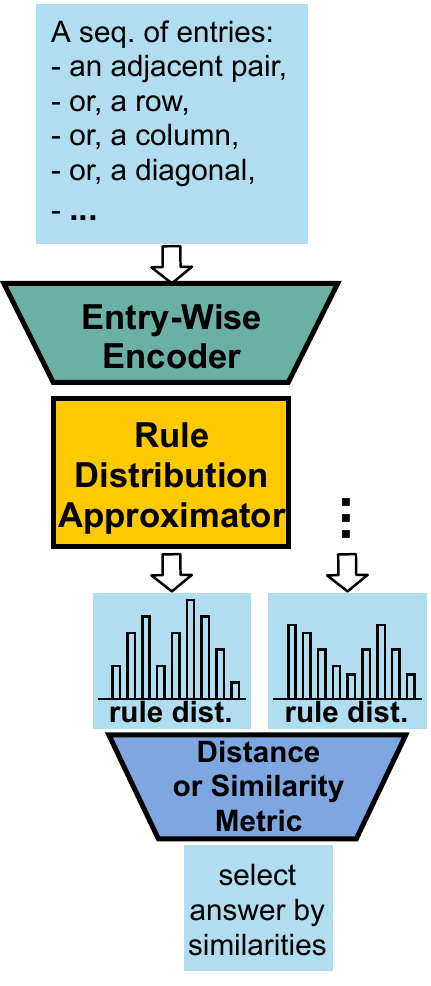}
  \caption{Type 1}
  \label{fig:lt1}
\end{figure}

A natural solution to reduce the reliance on the predefined representation system of geometric objects and rules is similar to the upgrade from the logical reasoning to the neuro-symbolic reasoning: instead of approximating the distribution of attribute values of entries, we can directly approximate the conditional distribution of possible rules given multiple matrix entries (through standard or customized neural network approximators). Therefore, the rules work only as labels to distinguish between different rules and no formal representations of geometric objects and rules are involved in computation. 

Two typical examples of Type 1 are Pairwise-ADV and Triple-ADV \citep{Mekik2017deep, mekik2018similarity} which approximate distributions of random variables of binary and ternary rules, respectively. A binary rule variable indicates whether a binary rule applies to two adjacent entries, for example, whether the objects in the two entries are of the same color, while a ternary rule variable denotes whether a ternary rule applies to three adjacent entries, such as the number of objects in Entry C equals the sum of geometric objects in Entry A and B. Another example of Type 1 is DeepIQ \citep{mandziuk2019deepiq}, in which the variable of rules between two adjacent entries is an ordered categorical variable (rather than binary), for example, the objects in the two entries differ by 3 units in their sizes. The random variables used these two examples are similar to using the different formal representation systems in the logical reasoning approach, but they are functionally equivalent.

The parallelism heuristic---spatial parallelism implies abstract conceptual parallelism--is commonly used to determine the combinations of matrix entries to present to the distribution approximator. According to the parallelism heuristic, the rule distributions in parallel rows or columns should be the same or similar; thus, probability metrics, such as KL-divergence, or general similarity metrics, like Euclidean distance, are used to measure the similarity; the answer choice is chosen so that it gives a last row/column whose rule distribution is most similar to the ones of the context rows/columns.

A diagram of Type 1 is given in Figure~\ref{fig:lt1}. Note that a entry-wise encoder is used to process each input entry individually, which is similar to the perception frontend in probabilistic reasoning. But, unlike the perception frontend, the entry-wise encoder does not necessarily output distributions over the predefined representations of geometric objects. The entry-wise encoder is to represent any latent space that can be used to approximate the rule distributions. After the entry-wise encoder encodes every entry in an input sequence, the embeddings of these entries are further aggregated and processed by the rule distribution approximator; the rule distributions of different sequences are finally compared to select the answer choice. The entry-wise encoder and rule distribution approximator can be implemented based on various neural network modules, such as CNN, ResNet and MLP. In practice, these two modules are jointly trained given the ground-truth rule labels of entry sequences.

\subsubsection{Type 2}

\begin{figure}[!bthp]
  \centering
  \includegraphics[width=0.2\textwidth]{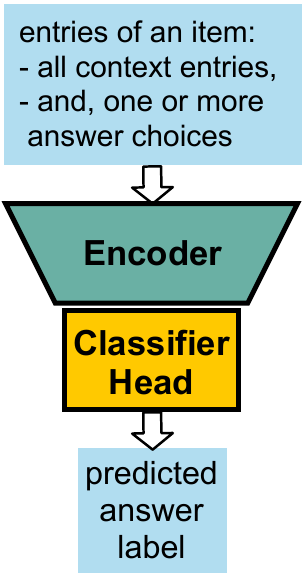}
  \caption{Type 2}
  \label{fig:lt2}
\end{figure}

Unlike the approaches in Stage 1, 2, and 3, Type 1 has avoided composing computing streams that explicitly rely on the predefined formal representation systems. But it still relies on the ground-truth rule labels and the parallelism heuristic. This issue is solved in Type 2, which is free of the reliance, as shown in Figure~\ref{fig:lt2}. This type converts an RPM into a classification problem, where the class labels are the correctness of each answer choice. In particular, when only one answer choice is included in the input, it is a binary classification problem; when all answer choices are included, it is a multi-class problem. 

Readers might have noticed a difference between Figure~\ref{fig:lt1} and Figure~\ref{fig:lt2}---the entry-wise encoder has been replaced by an encoder (not necessarily entry-wise). As the name indicates, the encoder takes as input multiple entries and, thus, the relational information between entries are thus encoded into its output. This difference gives rise to the difference between perceptual and conceptual processing. In RPM, perceptual processing is the processing of each single matrix entry, whereas conceptual processing generally involves reasoning about the relations between multiple matrix entries, i.e., the rules that govern the variation of multiple matrix entries. If one wishes to explicitly separate these two types of processing, one would have a module that attends to each entry individually and another module to aggregate the outputs of the first module, as in Type 1. This design choice is important for building computational models for visual abstract reasoning tasks. By changing the name to ``encoder'', we implies that Type 2 does not necessarily require an explicit separation of perceptual and conceptual processing. 

\begin{figure}[!bthp]
  \centering
  \includegraphics[width=0.25\textwidth]{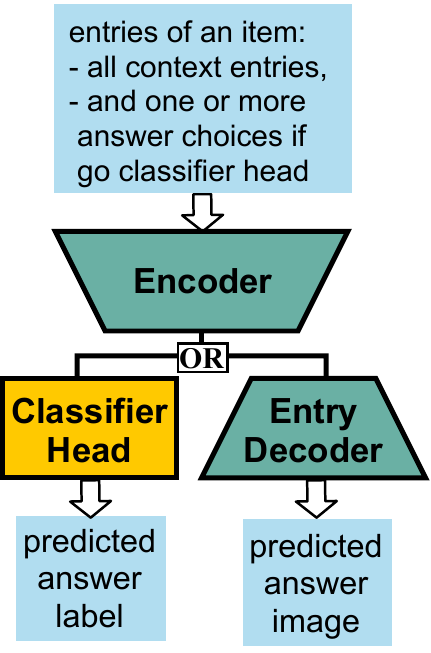}
  \caption{Type 2+}
  \label{fig:lt2-1}
\end{figure}

\citet{hoshen2017iq} implemented the first Type-2 model using a CNN encoder and an MLP classifier, and tested it on simple figural series and RPM-like tasks. This CNN+MLP model has since been used as a baseline to evaluate later works. Being influenced by popular works in image classification, the early attempts to solve RPM-like tasks by learning models mostly follows the structure of Type 2, for example, the Wild-ResNet+MLP model \citep{barrett2018measuring} and the ResNet+MLP model \citep{zhang2019raven}, respectively representing the binary and multi-class versions of Type 2. In the work of \citet{hoshen2017iq}, they also proposed the generative counterpart of the CNN+MLP model, by replacing the MLP classifier with a deconvolutional module to generate the predicted answer image (no answer choice is provided as input in this case). We include this modification in Type 2 by upgrading Type 2 to Type 2+, as shown in Figure~\ref{fig:lt2-1}. 


\subsubsection{Type 3}

\begin{figure}[!bthp]
  \centering
  \includegraphics[width=0.23\textwidth]{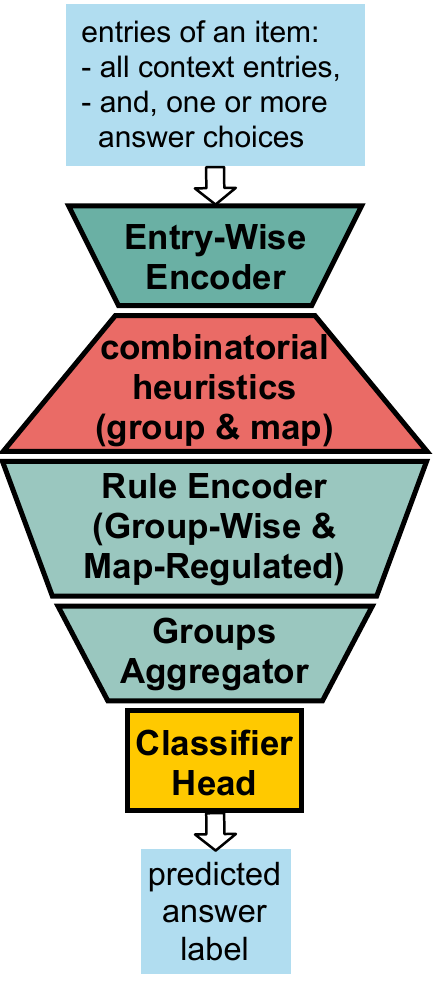}
  \caption{Type 3}
  \label{fig:lt3}
\end{figure}

By following the formulation of image classification, Type 2 eliminates the reliance on the ground-truth rule labels and the parallelism heuristic. However, visual abstract reasoning is a conceptually different tasks from image classification. In particular, abstract high-order relations need to be built upon raw perceptual visual input. The reason why a human participant thinks a visual abstract reasoning item difficulty is not because she cannot recognize the simple geometric objects in the item, but because the abstract concepts and relations could be complex, diverse, and hard to be extract from the simple geometric objects. In the latter case, concrete concepts are built upon complex visual stimuli, for example, recognizing daily objects in various backgrounds. Therefore, without further customization, the standard learning models for image classification is not able to give a satisfying solution to visual abstract reasoning. 

By comparing the Type 2 with its predecessors, which performs well on RPM (but rely on predefined formal representation systems of goemetric objects and rules), we find that an unnecessary design in Type 2 is that it does not separate perceptual and conceptual processing, which has been proved to be beneficial for visual abstract reasoning in many later works. This observation leads us to Type 3, as shown in Figure~\ref{fig:lt3}. Note that one can argue that Type 3 as a special case of Type 2, by regarding everything before the classifier head as a single module. But models based on this specification generally perform better than the typical models of Type 2.

After the entry-wise encoder encodes every entry, these entry embeddings go through a combinatorial process, in which subsets of these entry embeddings are selected and fed into next module subset by subset. In Figure~\ref{fig:lt3}, we use two trapezoids of opposite orientations for the entry-wise encoder and this combinatorial process to indicate that the amount of information is compressed and decompressed (i.e., the number of combinations is more than what are combined). As the name ``combinatorial heuristics'' indicates, Type 3 explicitly relies on some heuristics to take combinations, which include but not limited to the aforementioned parallelism heuristic. Essentially, these heuristics inform the model of which entry embeddings, together as a group, would make an instance of a rule. Each group is individually processed by a singleton rule encoder to produce a rule embedding for the group. At last, all rule embeddings are aggregated for classification.

A typical example of Type 3 is the WReN model \citep{barrett2018measuring}. WReN takes as input all context entries and one answer choice (thus solving binary classification). The entry-wise encoder is a small CNN (plus tagging the entry embeddings with one-hot position vectors indicating the entries' positions in the matrix). For combinatorial heuristics, WReN considers all binary rules (i.e., relations between every two entries). Note that WReN does not use the parallelism heuristic, which is commonly used in other models; but the position-tagged entry embedding compensates this, because the rule encoder can easily find the non-parallel sequences through position tags and output a specific rule-embedding to indicate this for the following processing. The groups aggregator in WReN is simply a summation.

\begin{figure}[!htbp]
  \centering
  \includegraphics[width=0.3\textwidth]{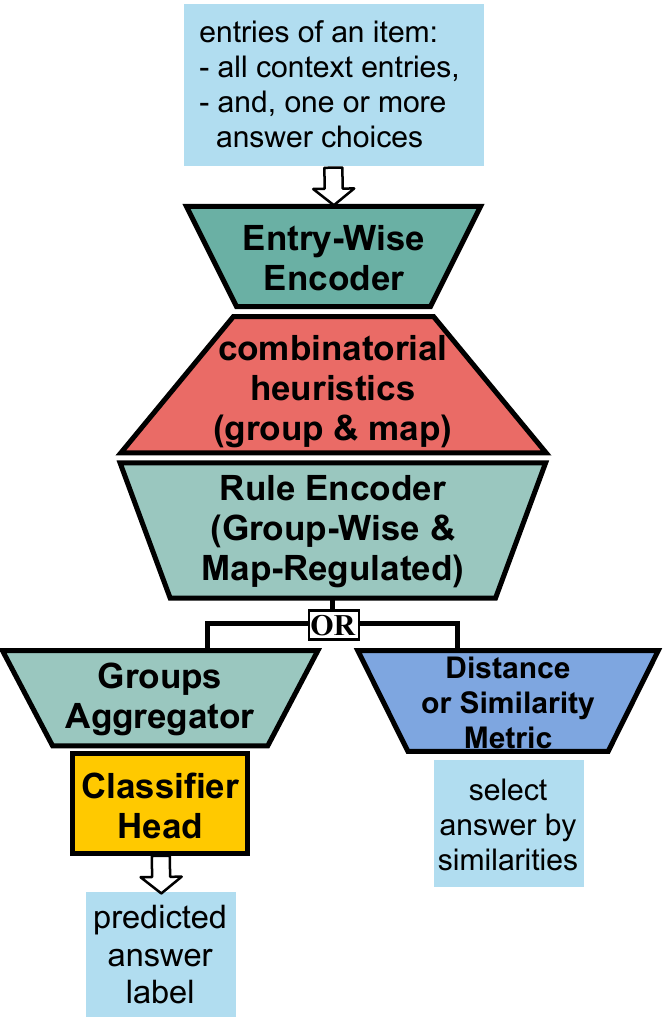}
  \caption{Type 3+}
  \label{fig:lt3-1}
\end{figure}

Following WReN, a series of models of Type 3 have been created, using different entry-wise encoders, combinatorial heuristics, rule encoders and groups aggregators. For example, LEN \citep{zheng2019abstract} considers only ternary rules for combinatorial heuristics, i.e., groups every three entries together for rule encoding, and applies gating variables to each groups in aggregator instead of tagging positions of entries (unsurprisingly, the experiment results showed that all gating variables but the ones of rows and columns were zeroed);
MXGNet \citep{wang2020abstract} also considers ternary rules, uses CNN or R-CNN as entry-wise encoder, relies on parallelism heuristic for combinatorial heuristics (instead of gating variables), and employs a graph-learning-based rule encoder that regards the 3 entries as a graph and computes the graph embedding as the rule embedding.

Different from the previous Type-3 models, multi-layer RN \citep{jahrens2020solving, jahrens2019multi, jahrens2018multi} extends the relation encoding in WReN into a multi-layer form. This is, the relation embeddings of entry groups are not aggregated into a single embedding for classification, but into multiple embeddings, which are further fed into another combinatorial module and rule encoder. Therefore, one could visualize multi-layer RN as a Type-3 model, repeating the middle three modules as many times as needed. Intuitively, higher-order relations can better be extracted through this multi-layer design.

The SRAN model \citep{hu2021stratified} adopts a more complicated encoding scheme by using multiple encoders and multiple rule encoders, where entries of two context rows/columns of 3$\times$3 matrices (6 in total) are encoded entry-wise, 3-entries-wise, 6-entries-wise by three different encoders, and the resulting entry-embeddings, 3-entry-embeddings and 6-entry-embeddings are sequentially integrated by three rule encoders into a single rule embedding, representing the rule of these two context rows/columns. The encoding scheme of SRAN, though complicated, does not deviate too much from Type 3. But, in stead of using rule embeddings to solve the item as an classification problem, SRAN directly uses similarity metrics of rule embeddings to select the answer, as in Type 1, which is also a common practice (just a different way to present the same supervising signal). Thus, it gives us a more complete Type 3+, as shown in Figure~\ref{fig:lt3-1}.

MRNet \citep{benny2021scale} is another Type-3 model using multiple entry encoders and multiple rule encoders, which process the input at multiple resolutions, determined by different layers' output in a CNN entry-wise encoder. The computational streams of different resolutions proceed separately and are aggregated at the end for classification. 

Both these two models---SRAN and MRNet---are examples of using multiple entry encoders and multiple rule encoders. Another model---NSM \citep{shekhar2021neural}---would be a better example to show the flexibility of Type 3. In particular, NSM solves the analogy-making task through two different rule encoders---a LSTM rule encoder and a modular network encoder---for the base domain and the target domain, respectively. Moreover, the structure of the modular network depends on the output of the LSTM rule encoder. These examples imply that, to build a Type-3 model, one can use not only multiple encoders but also different types of encoders, and that the multiple encoders can be assembled in more complex ways, rather than being parallel.

Readers might have noticed the words ``group'' and ``map'' in the diagrams of Figure~\ref{fig:lt3} and \ref{fig:lt3-1}. By these words, we intend to call attention to a mechanism that is pervasive in information processing for visual abstract reasoning, i.e., which pieces of information should be grouped together and thus to be aggregated later, and which pieces of information should be mapped
\footnote{or aligned, or corresponded; we use ``map'' to resonate with structure-mapping theory of analogy making; i.e., if two entities in the base and target domains are mapped to each other, then they are analogous to each other.} 
and thus to be processed in the same way\footnote{i.e., processed by the same module to force the analogical relation between them.}. These two types of decisions are interdependent on each other; more precisely, they are better to be viewed as two aspects of the same cognitive process. These decisions have to be made repeatedly at every level in information processing. Unfortunately, there might not be a centralized or universal theory for this grouping-mapping mechanism. As one can see in these Type-3 models, they all resort to some specific heuristics, which might not be always incorrect for visual abstract reasoning tasks.

\subsubsection{Type 4}

\begin{figure}[!htbp]
  \centering
  \includegraphics[width=0.2\textwidth]{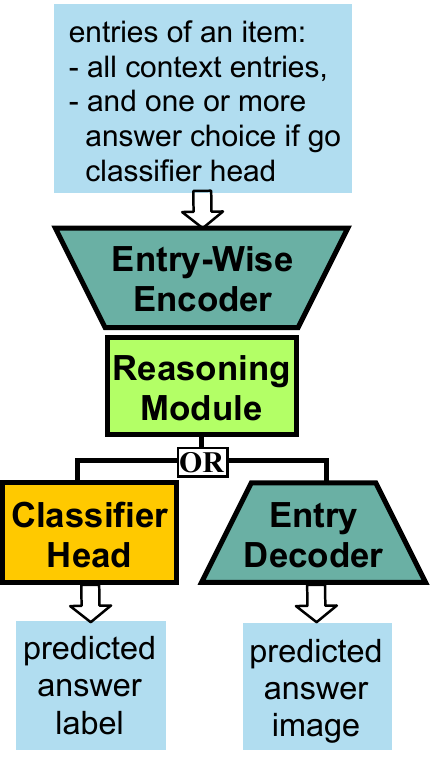}
  \caption{Type 4}
  \label{fig:lt4}
\end{figure}

Now, it is a good time to look back at the path that we have walked down for reviewing data-driven approaches and summarize how we come here:
\begin{itemize}
    \item From neuro-symbolic reasoning to Type 1: we eliminate the need of predefined representation systems of geometric objects and rules, but introduce the need of ground-truth rule labels. Parallelism heuristics is also inherited. 
    \item From Type 1 to Type 2: we eliminate the need of the ground-truth rule labels and parallelism heuristics, but the models do not perform well, because we use models for image classification, which is a fundamentally different task from visual abstract reasoning.
    The problem is solved in an image-classification way. 
    \item From Type 2 to Type 3: we separate perceptual and conceptual processing to make the model more suitable for visual abstract reasoning. Although the models perform reasonably well, specific grouping-mapping mechanism (or combinatorial heuristics) are needed for solving different RPM-like tasks. 
\end{itemize}
In this path, every time we want to eliminate the need of some prior knowledge, we introduce one or more neural networks modules to learn it from annotated data. This general solution in Stage 3 makes the procedural aspect, i.e., the process of computing, of solving RPM less of problem as the procedure can be interpolated from the input and expected output, through learning. The critical research point of the learning approach that determines the outcome of learning is thus shifted to the structural aspect. In visual abstract reasoning, the structural aspect includes hierarchical structure of processing, for example, the separation of perceptual and conceptual processing. With another layer of analogical processing (i.e., higher-order processing involving multiple relations), it would make a more complete hierarchy. In another dimension, the structural aspect also includes grouping-mapping mechanism we mentioned above. If one cannot abstract the task into these factors in the structural aspect and identify the ``atomic'' ones that can be easily solved through learning, the resulting learning model would not be effective and generalizable in the entire problem domain. 

As indicated above, a remaining factor unsolved in Type 3 is the combinatorial heuristics. Type 4 attempts to solve it by regarding grouping-mapping mechanism and rule encoding as a single ``atomic'' factor that can be learned through a single module---reasoning module, as shown in Figure~\ref{fig:lt4}. The reason for combining them is empirical and pragmatic, because they are interwoven and it is hard to say the former determines the latter, or the other way around. Since the reasoning module of Type 4 contains no grouping-mapping heuristics, its output does not necessarily indicate various rule among entries, and thus cannot be processed as in Type 3/3+. Thus, supervising signals are directly applied on this output. If we go back to see the structure of Type 2, you will find that Type 4 resembles Type 2 in appearance. Nonetheless, Type 4 is much more effective than Type 2 on RPM and it takes many trials and errors to settle on this solution. It has now become a relatively stable solution to visual abstract reasoning, and different core techniques have been used to implement the reasoning module. We summarize the works into four categories using distinct reasoning kernels.


\paragraph{Reasoning Kernel 1: CNN}
CNN has been a basic tool to extract features from raw perceptual input, and the extracted features are not only relevant for solving specific downstream tasks, but also representing correlations in the input. Solving visual abstract reasoning tasks is also to process correlations among raw inputs. Theoretically, CNN would have been an effective solution to RPM-like tasks. However, several early influential works \citep{barrett2018measuring, zhang2019raven, zhang2019learning} argued that CNN and CNN-based models are not capable of solving RPM-like tasks. 
\footnote{This is also why the CNN+MLP and ResNet+MLP models have been constantly used as baselines.}.
Since then, the research has been mainly focusing on other solutions. Ironically, after several years of exploration, \citet{spratley2020closer} proposed two Type-4 models---Rel-Base and Rel-AIR---which are all CNN-based models and perform well on both PGM and RAVEN. After comparing these two models with the previous CNN models, we found that the difference is whether the conceptual processing and the perceptual processing are separated. Taking Rel-Base as an example, its entry-wise encoder is a CNN module and its reasoning module is also a CNN module; all the entry embeddings are first stacked together and then convolved with convolution kernels in the reasoning module. But the baseline CNN-based models do not have this artificial separation. Therefore, we conjecture that the outstanding performance of many non-CNN models is not because they found better solutions than CNN, but because they separate perceptual and conceptual processing. On the flip side, another implication is that when using a single CNN module for both perceptual and conceptual processing, it is an extremely difficult task to learn this separation from data, i.e., learn the hierarchical structure of the task and how the information at each level is correlated. However, from the perspective of general problem solving, it would be impossible for us to know when the perceptual-conceptual separation lies for every possible task; in this case, we would have to use a single huge monolithic model; and how such a model can be trained effectively would be an important future research question.

\paragraph{Reasoning Kernel 2: LSTM}
A typical Type-4 model is the CNN+LSTM+MLP model \citep{barrett2018measuring}. This model takes as input all context entries and one or more answer choices. Each entry embedding is sequentially processed by an LSTM reasoning module, and the final state of LSTM is fed into an MLP classifier to predict the answer. This model is also used as a common baseline in many later works. LSTM has also been combined with other modules: Double-LSTM \citep{sekh2020can} uses two LSTM modules, which each specialize in different rule types and are coordinated by an extra module trained to predict the rule type\footnote{The reliance on ground-truth rule labels slightly deviates from our definition of Type 4.}; 
ESBN and NTM \citep{sinha2020memory, webb2020emergent}, combining LSTM with external memory modules, can also be used as the reasoning kernels in Type 4.

\paragraph{Reasoning Kernel 3: Self-Attention}
Another commonly used reasoning kernel is the self-attention module, which is composed of a multi-head attention and a feed-forward network (with residual connections and normalization). The most typical example of this reasoning kernel is the ARNe model \citep{hahne2019attention}. It extends the Type-3 model, WReN, by inserting between the entry-wise encoder and the combinatorial heuristics a self-attention module. Note that although ARNe inherits the combinatorial heuristic of WReN, it is no longer a Type-3 model because the self-attended embeddings no longer represent individual entries. Instead, each self-attended embedding contains information about all the matrix entries, and should better be considered as summaries of the whole matrix from different angles. Therefore, the inherited combinatorial heuristics module and the following modules of WReN can be considered similar to other general classifier heads, 
simply aggregating the input and predicting the answer. With hindsight, a reasonable order should have been first testing the self-attention module with a simpler classifier head rather than WReN. 

A similar example is the HTR model \citep{an2020hierarchical}, where an R-CNN entry-wise encoder is used to extract all geometric objects in each entry and two self-attention-based sub-modules are used to move the reasoning from object-level to entry-level and from entry-level to matrix-level. The first sub-module takes as input the object embeddings in a single entry and sums up the self-attended object embeddings as the entry embedding. Unlike ARNe and WReN solving RPM as binary classification, HTR solves it as multi-classification. Therefore, the output of the second sub-module contains 8 embeddings corresponding to the 8 answer choice. These 8 embeddings are fed into a contrastive classifier head \citep{zhang2020learning} to predict the answer label.

A more general example is the Neural Interpreter model \citep{rahaman2021dynamic}, which implements its most basic building block ``function'' as a self-attention module associated with two learnable vectors, which affects the module's computation and its access to input data, respectively. The self-attention modules are analogous to functions in programming language (as the term ``interpreter'' indicates), with one vector defining the function body and the other defining the function signature (type-matching particularly). A neural interpreter is composed of multiple iterations of a finite set of functions. As the original self-attention in Transformer, it converts a set of embeddings into a set of corresponding embeddings decorated with relational information. Neural interpreter was tested on RPM as binary classification. A CNN entry-wise encoder is used to produce entry embeddings. As in BERT \citep{devlin2018bert}, a classification token is included in the input embeddings, whose corresponding output embedding was fed into a linear classifier head.

\paragraph{Reasoning Kernel 4: Multi-Head Rule Detector}
The last reasoning kernel is closely related to the rule encoder of Type 3. Recall that the combinatorial heuristics module in Type 3 groups the entry embeddings into multiple groups, and each group is separately processed by the rule encoder to obtain a rule embedding for this group. Although this rule encoder has 1-in and 1-out, it is responsible for recognizing and encoding all the possible rules that might occur in the input. Recall that, by moving from Type 3 to Type 4, we intended to eliminate the reliance on combinatorial heuristics. An natural alternative solution could be an ``all-in-all-out'' rule encoder (rather than 1-in-1-out), which takes as input all the entry embeddings of a matrix (no grouping) and outputs all the possible rules. The relationship between ``1-in-1-out'' and ``all-in-all-out'' is analogous to the relationship between image classification versus object detection, where multiple objects exist in the image. Particularly, the new rule encoder can have multiple output heads, where later supervising pressure can be applied to force each head to represent a specific rule or specific rules. Therefore, we refer to this reasoning kernel as multi-head Rule detector. This kernel is underrepresented because we found only one model using this kernel---the SCL model \citep{wu2021scattering}, but it is very efficient for visual abstract reasoning. 


\subsection{Stage 4: Data Manipulation}
\label{supp:training}


The reported performance of some learning models in Stage 3 has already surpassed human performance under certain circumstances. However, the unreported or non-highlighted performance is far from satisfactory. A serious issue is that abstract concepts are not learned by these models because they do not generalize well when the abstract concepts are presented in different perceptual stimuli. This type of generalization is fundamental to visual abstract reasoning and also a hallmark of human intelligence. 
Therefore, the exploration has never stopped. Since the four types of learning models in the last stage explored many structural possibilities for building learning models, we have observed more and more efforts on studying the problem domain per se and how it is solved by human. This is perfectly understandable because when one realizes that all the existing tools do not work, she will naturally scrutinize the problem per se and try to understand why it is different from previously solved problems. These efforts result in the works of Stage 4, which utilize the features of visual abstract reasoning task that do not necessarily exist in other tasks. These efforts also resonate with the upward-spiral pattern we mentioned at the beginning of this conceptual chronicle as these task-specific features are also heavily used in the approaches in Stage 1 and 2, though in different ways. In particular, datasets of RPM-like items are delicately manipulated to present the task to learning models in a similar way of how human perceive and conceptualize RPM items. This way, the works in Stage 4 could force the models to learn abstract concepts and specific visual stimuli, distinguish between them, generalize the abstract concepts to the entire domain, and, finally, build the ability on the entire problem domain.

\subsubsection{Auxiliary Training}
For the models of Type 2, 3 and 4, an extra classifier head can be attached to exactly where the existing classifier head is attached to predict the meta-target of the input RPM-like item, which is a multi-hot vector indicating the attributes of geometric objects and rules in this item. These meta-targets are usually accessible in algorithmically-generated datasets, such PGM and RAVEN. The learning models can thus be trained on the answer labels and meta-targets simultaneously. The training on meta-targets is often referred to as auxiliary training in literature. 

Intuitively, this extra supervising signal can boost the accuracy of the answer-label classifier head. Auxiliary training was first tried with the WReN model on the PGM dataset and indeed showed a approximately 10\% boost (in IID generalization regime). The contribution of auxiliary training was verified by a high correlation between the two classifier heads' accuracies \citep{barrett2018measuring}. Similar observations on PGM were also found in other studies \citep{pekar2020generating, hahne2019attention}. In particular, the ARNe model would not even converge without auxiliary training. 

However, the effect of auxiliary training is still inconclusive. \citet{benny2021scale} showed that auxiliary training on PGM could only increase the accuracy of 1-rule items but decrease the accuracy of multi-rule items. This could cause the decrease of the overall accuracy when the dataset is composed of complex RPM-like items. Besides being affect by rules, the effect also differs between datasets. It has been reported that the auxiliary training would generally decrease the performance on the RAVEN dataset \citep{zhang2019raven, zhang2019learning, zheng2019abstract, wang2020abstract}, with one exception \citep{kim2020few}, which used a special contrastive loss and will be discussed later. Besides, \citet{malkinski2020multi} also showed contradictory results that when the meta-target is encoded in a sparse manner (the above works are all dense-encoding), the auxiliary training can increase the performance on RAVEN. Therefore, we can only say that the effect of auxiliary training is jointly determined by model, loss function, dataset, and meta-target encoding.

\subsubsection{Disentangled and Generative Representations}
The neuro-symbolic reasoning in Stage 3 has been frequently using standard neural networks, such as autoencoder and CNN, as the perception frontend to construct representations of entry image with explicit symbolic meaning. In contrast, as we mentioned in Type 1 of Stage 4, the symbolic meaning of encoders' output is not guaranteed. In addition to representations with symbolic meaning, disentangled and generative representations are used in Stage 4. For example, the Type-1 model, DeepIQ \citep{mandziuk2019deepiq}, uses a variational auto-encoder (VAE) as its encoder, which is pretrained on entry images of the Sandia dataset and kept frozen when the rule approximator is trained later.

Several advantages of disentangled and generative representations in RPM have been reported, such as data efficiency \citep{van2019disentangled}, robustness to distracting attributes \citep{zheng2019abstract} and better OOD generalization \citep{steenbrugge2018improving}. Disentangled and generative representations of entry images are usually obtained through VAE or its variants. For examples, in Type 3, $\beta$-VAE, FactorVAE, $\beta$-TCVAE and DIP-VAE were pretrained on entry images and the frozen encoders were combined with WReN \citep{steenbrugge2018improving, van2019disentangled}; a reduced version of MRNet was jointly trained with a VAE to simultaneously predict the answer label and generate the answer image (thus we call it generative-MRNet) \citep{pekar2020generating}. For Type-4 models, the VAE is usually jointly trained with the reasoning module, for example, the aforementioned ESBN model \citep{sinha2020memory}, and the LoGe model \citep{hersche2022neuro}, which uses vq-VAE as its encoder and decoder. Another special example of Type 4 is the Rel-AIR model \citep{spratley2020closer}, which integrates into its encoder an Attend-Infer-Repeat model \citep{eslami2016attend}---a model that can bee thought of as iterative VAE.

\subsubsection{Contrastive learning and Manipulating Data}
In addition to supervised learning, contrastive learning has also been used for solving RPM. We need to point out that the techniques of contrastive learning have been highly adapted to employ the structural and analogical characteristics of RPM and thus might not strictly follow the paradigms of contrastive learning. Particularly, the characteristics of RPM provides more options to manipulate data, such as decomposing matrices into rows and columns and regrouping them, and regrouping answer choices and even RPM problems, and various supervising signals can be applied to contrast the decomposed and regrouped data. 

\paragraph{Intra-Item Contrasting: Row/Column Contrasting}
The minimum structure that can be contrasted is rows/columns of a matrix. This type of contrasting was first attempted in the MCPT model \citep{zhuo2020solving}, where 8 answer choices are inserted into the 3$\times$3 matrix to obtain 10 rows/columns (2 context rows/columns and 8 answer choice rows/columns). The context rows/columns are assigned pseudo-label 1 and answer choice rows/columns are assigned pseudo-label 0; and this newly constructed pseudo-dataset of row/columns is learned by a Type-2 model, assuming that only one ``mis-assigned'' pseudo-label for the correct choice row/columns does not affect the final result of learning. To solve RPM, the answer choice row/column with the highest predicted output (between 0 and 1) is selected.

The intuition behind MCPT is to capture any characteristic that distinguishes between the correct and incorrect choices when they are embedded into the third row/column. In particular, it checks whether the third row/column has a meaningful variation that is similar to any context row/column in the dataset. The PRD model \citep{kiat2020pairwise} enhanced this type of single-row/column contrasting by including the parallelism heuristic. As in standard contrastive learning, positive and negative pairs are constructed from rows/columns, where the first two rows/columns in an RPM matrix make a positive pair. The negative pair could be constructed in different ways, such as rows/columns from different RPM-like items, randomly shuffled rows/columns of the same RPM, or filling the third row/column with a random non-choice entries. In PRD, a Type-2 model is used to learn a metric to measure the similarity between the two rows/columns in a pair. To solve an RPM, the choice row/column that is most similar to the first two rows/columns is selected. Compared to the single-row/column contrasting, the double-row/column contrasting is more common, which could be found in many other works. For example, the aforementioned generative-MRNet \citep{pekar2020generating} contrasts the answer choice rows/columns completed by the generated answer to the answer choice rows/columns completed by the given answer choices.

The rationale of moving from single-row/column to double-row/column contrasting was also exemplified by the LABC training/testing regime \citep{hill2019learning}, which makes the contrasting more accurate and complete through the meta-targets used in auxiliary training. Different from the single-row/column and double-row/column contrasting, where the effect of contrasting is applied through extra contrastive loss functions, LABC, as a training/testing regime, requires models to learn adapted datasets, which will force the model to contrast the rows/columns. In particular, an RPM-like item is adapted by muting some digits of its meta-target vector and regenerating the incorrect answer choices according to the muted meta-target vector.
Since meta-targets represent the rules and geometric objects that are used to generate RPM items, the newly-generated answer choices are partially correct. This way, the model will have to compare such answer choice rows/columns and the context rows/columns to find the correct answer, instead of only seeking meaningful variations in the answer choice row/column as in the single-row/column contrasting. LABC makes this idea more systematic by introducing the concepts of semantically and perceptually plausible answer choice corresponding to muting different subsets of mete-target digits and using distracting objects and rules.

\paragraph{Intra-Item Contrasting: Matrix Contrasting}
Instead of contrasting rows/columns, we can also contrasting the matrices completed by each answer choice. This is essentially contrasting the answer choices in the context of context entries. The Type-2 model, CoPINet \citep{zhang2019learning}, is the first model performing such contrasting. The contrasting in CoPINet is two-fold---contrastive representation and contrastive loss. First, for an RPM-like item, the embeddings of the matrices completed by each answer choice are aggregated into a ``central" embedding, and their differences to the ``central" embedding are used in the following processing. Second, given the interweaving of these matrix embeddings, it naturally leads to a contrastive loss function that incorporates matrices completed by correct and incorrect answer choices and increases the gap between their predicted values. This contrastive loss function could be easily embedded into models of parallel computation streams, for example, the aforementioned HTR model \citep{an2020hierarchical}.

We need to point out that row/column contrasting and matrix contrasting are not exclusive. For example, the DCNet model \citep{zhuo2021effective} first uses row/col contrasting to compute the matrix embeddings and then uses the matrix contrasting to predict the answer.

\paragraph{Inter-Item Contrasting: Single-Label Contrasting}
The above contrasting has been restricted within a single RPM-like item. The contrasting can also be between multiple items. The ACL and Meta-ACL \citep{kim2020few} are the first two inter-item contrasting models. The relation between ACL and Meta-ACL is similar to that between single-row/column and double-row/column contrasting. Given an RPM, let $X$ be its incomplete context matrix (regarding the missing entry as an empty image), $X_i$ an incomplete matrix obtained by replacing the $i$-th entry with a white-noise image, and $X'$ an incomplete matrix obtained by randomly reordering the entries of $X$. The ACL model contrasts the positive pair $(X, X_i)$ with the negative pair $(X, X')$. The Meta-ACL resorts to meta-targets to compose positive and negative pairs. In particular, two incomplete matrices of two items of the same meta-target form a positive pair $(X_S, X_T)$, and the corresponding negative pair is $(X_S, X'_S)$. In both ACL and Meta-ACL, the contrasting effect is applied through an extra standard contrastive loss function.

The MLCL model \citep{malkinski2020multi} formalizes the idea of Meta-ACL in a multi-label setting by regarding multi-hot meta-targets as multi-labels. Therefore, instead of requiring positive pairs to have exactly the same meta-targets, MLCL regards pairs of intersecting meta-targets as positive pairs. Different from Meta-ACL, the completed matrices are used. In particular, the correctly completed matrices are used for inter-item contrasting, and the intra-item contrasting between the correctly completed matrix and its incorrectly completed matrices is performed as in CoPINet. These two types of contrasting losses are jointly optimized.

\subsubsection{Other Dimensions of Manipulating Data}
Besides contrasting, there are also other dimensions of manipulating data. For example, the FRAR model \citep{zheng2019abstract} utilizes a reinforcement learning teacher model to select items from an RPM-like item back to train a student model. The items in the bank are characterized by their meta-targets and the reward is the increase in accuracy of the student model. The models solving RPM-like datasets have also been examined in the setting of continual learning. For example, the RAVEN dataset can be divided into 7 batches according to its spatial configurations and the models are trained with different methods to mitigate forgetting when sequentially learning the 7 batches in different orders \citep{hayes2021selective}.

\subsection{Summary}
The food for thought to share with the readers is that the study of the problem domain and the exploration for general solutions are both important for the overall advance in this field, as indicated by the upward-spiral pattern in the conceptual chronicle of computational models reviewed above. On one hand, the technical development always explores new methods, on the other hand, it inevitably revisits the old ideas again and again until the problem is perfectly solved. Therefore, the most recent models are not necessarily superior to the traditional ones in nature, and the early approaches, like the imagery-based approach, might trigger the next cycle of technical development in future research.


\section{Discussion}
\label{sec:dis}

After a historical overview of RPM and the problem domain represented by RPM in Section~\ref{sec:rpm} and \ref{sec:rpm-like} and a conceptual chronicle of computational models for solving this problem domain in Section~\ref{sec:com-mod}, we will zoom away in this section to discuss more general topics related to intelligence testing and AI systems. A good introduction to these topics is through a fundamental cognitive process---analogy making. In particular, we list the following analogies about intelligence tests and AI system:
\begin{itemize} 
    \item Analogy A---Intelligence Test : Human :: Intelligence Test : AI system 
    \item Analogy B---Intelligence Test : Human :: AI Test : Human
    \item Analogy C---Intelligence Test : Human :: AI Test: AI System 
    \item Analogy D---Intelligence Test : AI System :: AI Test : Human
    \item Analogy E---Intelligence Test : AI System :: AI Test : AI system
    \item Analogy F---AI Test : Human :: AI Test : AI system
\end{itemize}
The AI tests in the analogies above specifically means the tests that are inspired by human intelligence tests and specially designed to evaluating AI systems, for example, PGM and RAVEN datasets. These AI tests represent the motivation of testing AI systems in a similar way of human intelligence testing. To be rigorous, we enumerate all the possibilities of permutating tests and test-takers in the above analogies. These analogies represent research questions in different fields. For example, cognitive scientists might be interested in A; test developers might be interested in B and E; AI researchers, might be interested in A, C, E, and F; and some people might be interested in D simply for exploration purpose. Many of works reviewed above allude to one or more of these analogies. But most of them did not take one more step to examine whether these analogies hold or under what conditions they holds. In this case, the result of these works should be interpreted with caution. When it comes to AI testing, We are particularly interested in Analogy C. It describes a situation where human intelligence testing and AI testing are similar and common test theories could possibly apply to both cases. This analogy further gives rise to two general dual topics that are important for building and testing AI systems, respectively:
\begin{itemize}
    \item How tests measure subjects: the validity of measuring AI in a similar way human intelligence is measured;
    \item How subjects solve tests: the implication of human intelligence for building AI systems.
\end{itemize}



\subsection{The Validity of AI Testing}
\label{subsec:validity}
Analogy C---Intelligence Test : Human ::AI Test: AI System---calls attention to the connection between human intelligence testing and AI testing. It describes a situation where AI tests based on human intelligence tests are used to evaluate AI systems, as human intelligence tests are used to measure human intelligence. However, whether this analogy holds remain largely unknown to us.  
If they are, conclusions about human intelligence can be translated to AI systems. For example, one can claim that an AI system has the ability of visual abstract reasoning if the system passes the tests of the algorithmically-generated datasets mentioned above. Analogy C is best represented by the learning models in Stage 3 because the learning models are mainly evaluated through specially designed AI tests, such as PGM, RAVEN, and Sandia. Most of the works discuss their AI systems and contributions in the background of human cognitive abilities, and attempt to draw the conclusions that are comparable to human intelligence when the AI systems perform well. Unfortunately, when we are enjoying the acclamation, an elephant in the room is still in the room---the analogy simply does not hold and there is no validity in building and evaluating these models in the way they are currently built and evaluated. Note that the word ``validity'' is two-fold: on one hand, it is the validity in psychometrics; on the other, it is practically meaningless. We will now elaborate on this using learning models as an example.


To prove the idea that the AI testing in the reviewed works is psychometrically invalid, we check if the determinants of validity of human intelligence testing hold for AI testing. 
\begin{itemize}
    \item The first determinant is that human intelligence tests, as other psychological tests, is to measure individual difference on some tasks. Statistical evidence show that the performances on many tasks are correlated, and experts use the word ``intelligence'' to denote the latent factor or factors that cause the correlation. In other words, it is humans' behavior that comes first; then the word ``intelligence'' is abductively defined to explain humans' behavior. When an AI system shows similar behavior on the tasks which are comparable to human performance on these tasks, it is not necessarily the same factor(s), i.e., human's intelligence factor(s), that is behind the behavior of the AI system. 
    To satisfy the first determinant in AI testing, we needs to show that the underlying mechanisms are the same or equivalent in all cases. Otherwise, we need to be more cautious when we are describing the AI system's ability and explicitly distinguish it from human cognitive abilities.
    \item The second determinant is the requirements for designing human intelligence tests: human intelligence tests are usually short to prevent the participant from being exhausted; the stimuli in intelligence tests are diverse and there is usually no repeating stimulus in a single test; meanwhile, the stimuli in intelligence tests are also concise so that it does not introduce confounding factors; the items need to be evenly spread on the spectrum of difficulty so that people at different ability levels can be measured; and so on. All these requirements contribute to the validity of intelligence tests and are not easily satisfied in AI tests. An exception is the Cognitive Design System Approach by \citet{embretson2004measuring}, but this approach has not used to develop any test for AI systems.
\end{itemize}
The determinants listed here are by no means complete given the complex nature of human intelligence testing, but are sufficient to break the analogy between human intelligence testing and AI testing.

Given the fundamental distinction between human intelligence testing and AI testing, we might simply abandon the idea of establishing the validity by comparing AI testing to human intelligence testing. Instead, as most works in AI, we analyze AI systems for solving intelligence tests and intelligence-test-like datasets purely from the perspective of problem solving, and claim that these AI systems are more capable of solving the tests or datasets than human participants. However, this brings us back to an old issue: the AI systems are specially prepared or trained on the items that are similar to the one used for testing, whereas testing items are kept secret from human participants, let alone training. For visual abstract reasoning, no AI system has shown performance that is comparable to human, especially when generalizing an abstract concept to new visual stimuli that were not associated with this concept before.

Nonetheless, we can still argue that these AI systems are useful because they can at least act as automatic tools to free humans from simple repeating tasks in our daily life. However, this is also not true because intelligence tests, especially general intelligence tests, are designed to distant from the our daily activities so that the result is not affected by one's previous experience. Thus, the ability to solve intelligence test items would not be able to assist human in most cases. Moreover, a cognitive ability or general intelligence does not correspond to a specific clearly defined task that is constantly repeating in certain scenario. Instead, it is abstracted from various daily activities. That is, it is common but also very sparse across various daily activities, and, more importantly, deeply interwoven with other abilities. There is simply no such simple clearly-defined repeating tasks where these AI systems can be applied. For other complex ill-defined tasks, these AI systems also need to be integrated with various other AI systems of different abilities. This kind research, though valuable, is still infeasible at the current stage of AI.


We can try to continue this debate by proposing more contributions and purposes of building AI systems for solving intelligence tests or intelligence-test-like tests. As long as the contribution is relative to human intelligence, we can always come up with a reason to 
refute it (except that the contribution is pure scientific exploration). Unfortunately, comparing to human intelligence is unavoidable on our way to implementing human-level AI. It seems that we have come to a dead end. 

The solution lies in the theory of analogy making and the origin of intelligence tests. Let us first check the analogy-making aspect of Analogy C to see if we interpret the analogy correctly. One of the most important theories of analogy-making is the structure mapping theory by \citet{gentner1983structure}. It emphasizes the similarity between the relations in the base and the target domains, rather than the literal similarity between objects in the base and target domains. In particular, the corresponding objects can be starkly different in a literal sense without compromising the strength of the analogy, when the corresponding relationships are similar. This seems trivial to humans who know how to make analogies. But people indeed make mistakes by relying on literal similarity rather than relational similarity when interpreting analogy. In fact, we did in interpreting Analogy C above. We started from corresponding human intelligence tests with AI tests by literal similarity, i.e., they are items to solve. We then took a simple relation ``human solves intelligence tests'' in the base domain and translated it into the target domain. After a thorough analysis, we found everything went wrong. We just made the very mistake that is just pointed by structure-mapping theory. Thus, interpreting analogy correctly might not be trivial at all in practice.

The correct interpretation starts from studying the relations in the base domain, which can be clarified by a revisit to the origin of intelligence tests. Modern schooling is actually a new manner of eduction compared to the whole history of education. It does not exist until the 20th century. At the beginning, educators found that some children had a great deal of trouble learning in this manner. In order to select the students who were suitable for modern schooling, the French Education Ministry hired Alfred Binet. The solution Binet provided was to test children's ability to solve problems that could be commonly solved by children at certain ages, determining the children's mental ages. The ratio of mental age to chronological age was used as an index to select students for school education. This index is the prototype of today's intelligence quotient. Therefore, the origin of intelligence tests tells us that intelligence tests were developed to measure individual difference of learning ability under a certain circumstance (school education) relative to the average of a certain group of people (peers). This definition echoes our discussion of RPM in Section~\ref{sec:rpm}.

While this definition of intelligence tests seems complicated, it does accurately describe the relations in the base domain of Analogy C. Now, let us check the target domain for a similar relational structure. The general idea of the target domain is undoubtedly to test AI systems. We can try to extract from the target domain the counterparts of the concepts in the definition of intelligence tests. The most important two concepts in the definition of intelligence tests is definitely ``learning ability'' and ``individual difference''. ``Learning ability'' of AI systems is a clear concept because it is native to the learning models. ``Learning ability'' has been considered as an integral part of AI systems (though the ``learning ability'' of AI systems might be the different from human learning ability). Thus, ``learning ability'' does not pose any problem to us. ``Individual difference'' of ``learning ability'' of AI system is less clearly defined because of the heterogeneous nature of various AI systems. Note that, in contrast to human intelligence testing, the inherent ``learning ability'' cannot be sufficiently reflected in the final outcome of learning. 
This problem can be solved if we considered the dual concept of ability---difficulty. 
Put simply, if we have items at various levels of difficulty, we can use human ability test items like a ruler to measure people's ability. On the flip side, if we know people at different levels of ability, we can use these people's response to these items to determine the difficulty of these items. That is, ability and difficulty are defined relative to each other. We are so familiar with difficulty in AI research because we have experienced so much of it. In particular, when evaluating AI systems' learning ability, the concept of difficulty is reified as learning tasks. We would say that a learning task is difficult to a specific AI systems or to a class of AI systems. In practice, learning tasks can be defined differently, such as different datasets, different ways to present datasets, and access to other resources. A good example of learning tasks is the different generalization regimes of PGM and RAVEN datasets \citeauthor{barrett2018measuring,zhang2019raven}, which correspond to different conceptual distances between the abstract concepts in training and testing. The more distant, the harder the learning task. Now, we can look back at the the ruler to measure human intelligence, on which the marks are individual test items. Therefore, to interpret Analogy C, we can make the correspondence between human intelligence test items and learning tasks of AI systems. In contrast to previous interpretation of Analogy C, this correspondence is not based on literal similarity but derived from the relational structures in the base and the target domain. This correspondence is extremely important for us to establish the general testing theory of AI systems, but might not be obvious from literal meaning of Analogy C. We now can interpret Analogy C as human intelligence tests measure human intelligence as AI tests of learning tasks measure AI systems.

It is important to point out that this interpretation of Analogy C is not just a rhetoric or an arbitrary makeshift. It calls attention to two basic factors that one needs to consider to establish a test theory---what is being measured? what is used to measure it? For the first question, we definitely want to measure the ``learning ability'' of AI systems. For the second question, we have a great many existing learning tasks for AI systems. The context to answer the second question is subtly different from human testing and more complex. First, when we are evaluating an AI system on a learning task, we are interested in the overall performance rather than the response on a specific instance of this task. For example, for an image classification task, we would compare the overall accuracies of two AI systems to conclude that one is more capable than the other. We would not make such conclusion because one system gives a correct prediction for a specific image while the other does not, unless this instance (the image) is fundamentally different from other instances and possibly posing more demands for processing. In that case, this instance would make a separate learning task. In both cases, the correspondence between human intelligence test items and learning tasks for AI systems remains unchanged. The context of AI testing is more complex than human intelligence testing because there exist various learning tasks and various AI systems to solve them, but, for now, not every AI system is designed to solve every learning task. And for practical purposes, we need these specialized AI systems in our society rather only pursuing the ultimate goal of human-level AI. For human intelligence tests, although people might perform extremely well on some subtests but terribly on the others, the tests are valid measure for all human beings. But, currently, one cannot design an AI test that applies to all AI systems. What we can do now is to identify problem domains and fundamentally different learning tasks in the domain, which can be used to compose tests for AI systems. When AI technology enters the era of Artificial General Intelligence (AGI) in the future, we can design AI tests using learning tasks across multiple problem domains.

In general, this interpretation of Analogy C allows us to establish a testing framework for AI systems, which is similar to the testing theories in human intelligence testing. This framework requires extra efforts to study problem domains and, more importantly, study cognitive information processing to identify various learning tasks in the problem domain. Therefore, it is naturally a interdisciplinary research direction. this framework proposes a much higher standard than how AI systems are tested now. Although it requires extra efforts to implement, it will make sure that we are making concrete progress.

\subsection{The Implication of Human Intelligence for Building AI systems}

Although the history of human intelligence testing is much shorter (approximately 100 years) compared to the time intelligence exists, humans' intelligence test scores have shown a substantial increase (Flynn effect). Many efforts have been made to find what is responsible for this increase. These efforts are important not only for human development but also for AI systems from the perspective of AI testing. Specific social changes have been used to explain Flynn effect, such as television, computer games, changes in school education and so on. Most of these explanations do not hold up because these social changes are not accompanied by the changes in intelligence test scores. Interestingly, the change in testing scores does correlate with to the changes in human's height, birth weight, and infant mortality in a more than general sense. Thus, the increase in intelligence test scores might be attributed to the same factors responsible for height, birth weight and infant mortality---i.e., improved living conditions such as food and medical care\citep{raven2000raven}.

When we are reviewing the development of AI, we are facing the same meta-question---what causes the development---that is not well answered. either. We could conclude that the recent improvement of AI is due to the increase of computational power and massive amount of data generated through internet. This explanation is not so different from attributing the increase of human intelligence to improvement of living conditions, which is not very operable for theoretical AI research. Apart from computational power and data, most of knowledge in basic science that are used in the cutting-edge AI technologies have been there for decades. Therefore, it is hard to find a theoretical factor that promoted the development of AI. 

A hypothesis from the social studies that was proposed to explain humans' cognitive development can better explain the development of AI than other explanations. \textbf{The Challenge Hypothesis} \citep{hunt2010human}:
\begin{quote}
    Intelligence is developed by engaging in cognitively challenging activities. Environments vary in the extent to which they support such challenges, and individuals vary in the extent to which they seek them out.
\end{quote}
The statement of the hypothesis is, though concise, but full of wisdom. In the last decades, the development of AI have been definitely accompanied by tasks that were initially challenging to AI systems, such as facial recognition and spam filtering, and later solved. These tasks did not exist before the era of AI. This argument echoes the emphasis in the last subsection on identifying and collating learning tasks for AI testing.

The second half of the challenge hypothesis---``individuals vary in the extent to which they seek them out''---is even more interesting. In the studies of human cognitive development, there is a somewhat surprising empirical result---eductive ability is more easily influenced by appropriate educational and developmental experience than reproductive ability. In particular, researcher found that educational self-direction, in which students are responsible for deciding what they need to learn, how they learn it, and what are goals, and complex educational activities (e.g., challenge and reasonable learning tasks) give rise to a cyclical development in cognitive ability \citep{raven2000raven}. These studies shed light on a possible promising future trend in AI research, in which AI systems take the initiative to seek out learning tasks in the challenging environment that provide the most efficient development. This trend implies a fundamental change to the paradigm of AI systems by shifting from learning specific tasks to interacting with the environment\citep{laird2017interactive}


\bibliography{main}

\begin{thebibliography}{109}
\expandafter\ifx\csname natexlab\endcsname\relax\def\natexlab#1{#1}\fi
\providecommand{\url}[1]{\texttt{#1}}
\providecommand{\href}[2]{#2}
\providecommand{\path}[1]{#1}
\providecommand{\DOIprefix}{doi:}
\providecommand{\ArXivprefix}{arXiv:}
\providecommand{\URLprefix}{URL: }
\providecommand{\Pubmedprefix}{pmid:}
\providecommand{\doi}[1]{\href{http://dx.doi.org/#1}{\path{#1}}}
\providecommand{\Pubmed}[1]{\href{pmid:#1}{\path{#1}}}
\providecommand{\bibinfo}[2]{#2}
\ifx\xfnm\relax \def\xfnm[#1]{\unskip,\space#1}\fi
\bibitem[{An and Cho(2020)}]{an2020hierarchical}
\bibinfo{author}{An, J.}, \bibinfo{author}{Cho, S.}, \bibinfo{year}{2020}.
\newblock \bibinfo{title}{Hierarchical transformer encoder with structured
  representation for abstract reasoning}.
\newblock \bibinfo{journal}{IEEE Access} \bibinfo{volume}{8},
  \bibinfo{pages}{200229--200236}.
\bibitem[{Arendasy(2002)}]{arendasy2002geomgen}
\bibinfo{author}{Arendasy, M.}, \bibinfo{year}{2002}.
\newblock \bibinfo{title}{Geomgen—ein itemgenerator f{\"u}r
  matrizentestaufgaben}.
\newblock \bibinfo{journal}{Wien: Eigenverlag} .
\bibitem[{Arendasy and Sommer(2005)}]{arendasy2005effect}
\bibinfo{author}{Arendasy, M.}, \bibinfo{author}{Sommer, M.},
  \bibinfo{year}{2005}.
\newblock \bibinfo{title}{The effect of different types of perceptual
  manipulations on the dimensionality of automatically generated figural
  matrices}.
\newblock \bibinfo{journal}{Intelligence} \bibinfo{volume}{33},
  \bibinfo{pages}{307--324}.
\bibitem[{Barrett et~al.(2018)Barrett, Hill, Santoro, Morcos and
  Lillicrap}]{barrett2018measuring}
\bibinfo{author}{Barrett, D.}, \bibinfo{author}{Hill, F.},
  \bibinfo{author}{Santoro, A.}, \bibinfo{author}{Morcos, A.},
  \bibinfo{author}{Lillicrap, T.}, \bibinfo{year}{2018}.
\newblock \bibinfo{title}{Measuring abstract reasoning in neural networks}, in:
  \bibinfo{booktitle}{International conference on machine learning},
  \bibinfo{organization}{PMLR}. pp. \bibinfo{pages}{511--520}.
\bibitem[{Benny et~al.(2021)Benny, Pekar and Wolf}]{benny2021scale}
\bibinfo{author}{Benny, Y.}, \bibinfo{author}{Pekar, N.},
  \bibinfo{author}{Wolf, L.}, \bibinfo{year}{2021}.
\newblock \bibinfo{title}{Scale-localized abstract reasoning}, in:
  \bibinfo{booktitle}{Proceedings of the IEEE/CVF Conference on Computer Vision
  and Pattern Recognition}, pp. \bibinfo{pages}{12557--12565}.
\bibitem[{Bethell-Fox et~al.(1984)Bethell-Fox, Lohman and
  Snow}]{bethell1984adaptive}
\bibinfo{author}{Bethell-Fox, C.E.}, \bibinfo{author}{Lohman, D.F.},
  \bibinfo{author}{Snow, R.E.}, \bibinfo{year}{1984}.
\newblock \bibinfo{title}{Adaptive reasoning: Componential and eye movement
  analysis of geometric analogy performance}.
\newblock \bibinfo{journal}{Intelligence} \bibinfo{volume}{8},
  \bibinfo{pages}{205--238}.
\bibitem[{Blum and Holling(2018)}]{blum2018automatic}
\bibinfo{author}{Blum, D.}, \bibinfo{author}{Holling, H.},
  \bibinfo{year}{2018}.
\newblock \bibinfo{title}{Automatic generation of figural analogies with the
  imak package}.
\newblock \bibinfo{journal}{Frontiers in psychology} \bibinfo{volume}{9},
  \bibinfo{pages}{1286}.
\bibitem[{Bohan and O'Donoghue(2000)}]{bohan2000ludi}
\bibinfo{author}{Bohan, A.}, \bibinfo{author}{O'Donoghue, D.},
  \bibinfo{year}{2000}.
\newblock \bibinfo{title}{Ludi: A model for geometric analogies using attribute
  matching}, pp. \bibinfo{pages}{110--119}.
\bibitem[{Bringsjord(2011)}]{bringsjord2011psychometric}
\bibinfo{author}{Bringsjord, S.}, \bibinfo{year}{2011}.
\newblock \bibinfo{title}{Psychometric artificial intelligence}.
\newblock \bibinfo{journal}{Journal of Experimental \& Theoretical Artificial
  Intelligence} \bibinfo{volume}{23}, \bibinfo{pages}{271--277}.
\bibitem[{Bringsjord and Schimanski(2003)}]{bringsjord2003artificial}
\bibinfo{author}{Bringsjord, S.}, \bibinfo{author}{Schimanski, B.},
  \bibinfo{year}{2003}.
\newblock \bibinfo{title}{What is artificial intelligence? psychometric ai as
  an answer}, in: \bibinfo{booktitle}{IJCAI}, \bibinfo{organization}{Citeseer}.
  pp. \bibinfo{pages}{887--893}.
\bibitem[{Burke(1958)}]{burke1958raven}
\bibinfo{author}{Burke, H.R.}, \bibinfo{year}{1958}.
\newblock \bibinfo{title}{Raven's progressive matrices: A review and critical
  evaluation}.
\newblock \bibinfo{journal}{The Journal of Genetic Psychology}
  \bibinfo{volume}{93}, \bibinfo{pages}{199--228}.
\bibitem[{Carpenter et~al.(1990)Carpenter, Just and Shell}]{carpenter1990one}
\bibinfo{author}{Carpenter, P.A.}, \bibinfo{author}{Just, M.A.},
  \bibinfo{author}{Shell, P.}, \bibinfo{year}{1990}.
\newblock \bibinfo{title}{What one intelligence test measures: a theoretical
  account of the processing in the raven progressive matrices test.}
\newblock \bibinfo{journal}{Psychological review} \bibinfo{volume}{97},
  \bibinfo{pages}{404}.
\bibitem[{Cattell(1941)}]{cattell1941some}
\bibinfo{author}{Cattell, R.B.}, \bibinfo{year}{1941}.
\newblock \bibinfo{title}{Some theoretical issues in adult intelligence
  testing}.
\newblock \bibinfo{journal}{Psychological Bulletin} \bibinfo{volume}{38},
  \bibinfo{pages}{592}.
\bibitem[{Cattell(1943)}]{cattell1943measurement}
\bibinfo{author}{Cattell, R.B.}, \bibinfo{year}{1943}.
\newblock \bibinfo{title}{The measurement of adult intelligence.}
\newblock \bibinfo{journal}{Psychological bulletin} \bibinfo{volume}{40},
  \bibinfo{pages}{153}.
\bibitem[{Cattell(1950)}]{cattell1950handbook}
\bibinfo{author}{Cattell, R.B.}, \bibinfo{year}{1950}.
\newblock \bibinfo{title}{Handbook for the individual or group culture fair
  intelligence test}.
\newblock \bibinfo{publisher}{Institute for Personality and Ability Testing}.
\bibitem[{Cattell(1963)}]{cattell1963theory}
\bibinfo{author}{Cattell, R.B.}, \bibinfo{year}{1963}.
\newblock \bibinfo{title}{Theory of fluid and crystallized intelligence: A
  critical experiment.}
\newblock \bibinfo{journal}{Journal of educational psychology}
  \bibinfo{volume}{54}, \bibinfo{pages}{1}.
\bibitem[{Cattell(1987)}]{cattell1987intelligence}
\bibinfo{author}{Cattell, R.B.}, \bibinfo{year}{1987}.
\newblock \bibinfo{title}{Intelligence: Its structure, growth and action}.
\newblock \bibinfo{publisher}{Elsevier}.
\bibitem[{Das et~al.(1994)Das, Naglieri and Kirby}]{das1994assessment}
\bibinfo{author}{Das, J.P.}, \bibinfo{author}{Naglieri, J.A.},
  \bibinfo{author}{Kirby, J.R.}, \bibinfo{year}{1994}.
\newblock \bibinfo{title}{Assessment of cognitive processes: The PASS theory of
  intelligence.}
\newblock \bibinfo{publisher}{Allyn \& Bacon}.
\bibitem[{Davies and Goel(2001)}]{davies2001visual}
\bibinfo{author}{Davies, J.}, \bibinfo{author}{Goel, A.K.},
  \bibinfo{year}{2001}.
\newblock \bibinfo{title}{Visual analogy in problem solving}, in:
  \bibinfo{booktitle}{IJCAI}, pp. \bibinfo{pages}{377--384}.
\bibitem[{Detterman(2011)}]{detterman2011challenge}
\bibinfo{author}{Detterman, D.K.}, \bibinfo{year}{2011}.
\newblock \bibinfo{title}{A challenge to watson}.
\bibitem[{Devlin et~al.(2018)Devlin, Chang, Lee and Toutanova}]{devlin2018bert}
\bibinfo{author}{Devlin, J.}, \bibinfo{author}{Chang, M.W.},
  \bibinfo{author}{Lee, K.}, \bibinfo{author}{Toutanova, K.},
  \bibinfo{year}{2018}.
\newblock \bibinfo{title}{Bert: Pre-training of deep bidirectional transformers
  for language understanding}.
\newblock \bibinfo{journal}{arXiv preprint arXiv:1810.04805} .
\bibitem[{Embretson(1995)}]{embretson1995role}
\bibinfo{author}{Embretson, S.E.}, \bibinfo{year}{1995}.
\newblock \bibinfo{title}{The role of working memory capacity and general
  control processes in intelligence}.
\newblock \bibinfo{journal}{Intelligence} \bibinfo{volume}{20},
  \bibinfo{pages}{169--189}.
\bibitem[{Embretson(1998)}]{embretson1998cognitive}
\bibinfo{author}{Embretson, S.E.}, \bibinfo{year}{1998}.
\newblock \bibinfo{title}{A cognitive design system approach to generating
  valid tests: Application to abstract reasoning.}
\newblock \bibinfo{journal}{Psychological methods} \bibinfo{volume}{3},
  \bibinfo{pages}{380}.
\bibitem[{Embretson(2004)}]{embretson2004measuring}
\bibinfo{author}{Embretson, S.E.}, \bibinfo{year}{2004}.
\newblock \bibinfo{title}{Measuring human intelligence with artificial
  intelligence: Adaptive item generation}, \bibinfo{publisher}{Georgia
  Institute of Technology}.
\bibitem[{Eslami et~al.(2016)Eslami, Heess, Weber, Tassa, Szepesvari, Hinton
  et~al.}]{eslami2016attend}
\bibinfo{author}{Eslami, S.}, \bibinfo{author}{Heess, N.},
  \bibinfo{author}{Weber, T.}, \bibinfo{author}{Tassa, Y.},
  \bibinfo{author}{Szepesvari, D.}, \bibinfo{author}{Hinton, G.E.}, et~al.,
  \bibinfo{year}{2016}.
\newblock \bibinfo{title}{Attend, infer, repeat: Fast scene understanding with
  generative models}.
\newblock \bibinfo{journal}{Advances in Neural Information Processing Systems}
  \bibinfo{volume}{29}.
\bibitem[{Evans(1964)}]{evans1964program}
\bibinfo{author}{Evans, T.G.}, \bibinfo{year}{1964}.
\newblock \bibinfo{title}{A program for the solution of a class of
  geometric-analogy intelligence-test questions.}
\newblock \bibinfo{type}{Technical Report}. AIR FORCE CAMBRIDGE RESEARCH LABS
  LG HANSCOM FIELD MASS.
\bibitem[{Freund and Holling(2011a)}]{freund2011get}
\bibinfo{author}{Freund, P.}, \bibinfo{author}{Holling, H.},
  \bibinfo{year}{2011}a.
\newblock \bibinfo{title}{How to get really smart: Modeling retest and training
  effects in ability testing using computer-generated figural matrix items}.
\newblock \bibinfo{journal}{Intelligence} \bibinfo{volume}{39},
  \bibinfo{pages}{233--243}.
\bibitem[{Freund and Holling(2011b)}]{freund2011retest}
\bibinfo{author}{Freund, P.}, \bibinfo{author}{Holling, H.},
  \bibinfo{year}{2011}b.
\newblock \bibinfo{title}{Retest effects in matrix test performance:
  Differential impact of predictors at different hierarchy levels in an
  educational setting}.
\newblock \bibinfo{journal}{Learning \& ind. diff.} \bibinfo{volume}{21},
  \bibinfo{pages}{597--601}.
\bibitem[{Freund and Holling(2011c)}]{freund2011wants}
\bibinfo{author}{Freund, P.}, \bibinfo{author}{Holling, H.},
  \bibinfo{year}{2011}c.
\newblock \bibinfo{title}{Who wants to take an intelligence test? personality
  and achievement motivation in the context of ability testing}.
\newblock \bibinfo{journal}{Personality \& Ind. Diff.} \bibinfo{volume}{50},
  \bibinfo{pages}{723--728}.
\bibitem[{Freund et~al.(2008)Freund, Hofer and Holling}]{freund2008explaining}
\bibinfo{author}{Freund, P.A.}, \bibinfo{author}{Hofer, S.},
  \bibinfo{author}{Holling, H.}, \bibinfo{year}{2008}.
\newblock \bibinfo{title}{Explaining and controlling for the psychometric
  properties of computer-generated figural matrix items}.
\newblock \bibinfo{journal}{Applied Psychological Measurement}
  \bibinfo{volume}{32}, \bibinfo{pages}{195--210}.
\bibitem[{Gentner(1983)}]{gentner1983structure}
\bibinfo{author}{Gentner, D.}, \bibinfo{year}{1983}.
\newblock \bibinfo{title}{Structure-mapping: A theoretical framework for
  analogy}.
\newblock \bibinfo{journal}{Cognitive science} \bibinfo{volume}{7},
  \bibinfo{pages}{155--170}.
\bibitem[{Gierl et~al.(2012)Gierl, Lai and Turner}]{gierl2012using}
\bibinfo{author}{Gierl, M.J.}, \bibinfo{author}{Lai, H.},
  \bibinfo{author}{Turner, S.R.}, \bibinfo{year}{2012}.
\newblock \bibinfo{title}{Using automatic item generation to create
  multiple-choice test items}.
\newblock \bibinfo{journal}{Medical education} \bibinfo{volume}{46},
  \bibinfo{pages}{757--765}.
\bibitem[{Hahne et~al.(2019)Hahne, L{\"u}ddecke, W{\"o}rg{\"o}tter and
  Kappel}]{hahne2019attention}
\bibinfo{author}{Hahne, L.}, \bibinfo{author}{L{\"u}ddecke, T.},
  \bibinfo{author}{W{\"o}rg{\"o}tter, F.}, \bibinfo{author}{Kappel, D.},
  \bibinfo{year}{2019}.
\newblock \bibinfo{title}{Attention on abstract visual reasoning}.
\newblock \bibinfo{journal}{arXiv preprint arXiv:1911.05990} .
\bibitem[{Hayes and Kanan(2021)}]{hayes2021selective}
\bibinfo{author}{Hayes, T.L.}, \bibinfo{author}{Kanan, C.},
  \bibinfo{year}{2021}.
\newblock \bibinfo{title}{Selective replay enhances learning in online
  continual analogical reasoning}, in: \bibinfo{booktitle}{Proceedings of the
  IEEE/CVF Conference on Computer Vision and Pattern Recognition}, pp.
  \bibinfo{pages}{3502--3512}.
\bibitem[{He et~al.(2021a)He, Ren and Bai}]{he2021one}
\bibinfo{author}{He, W.}, \bibinfo{author}{Ren, J.}, \bibinfo{author}{Bai, R.},
  \bibinfo{year}{2021}a.
\newblock \bibinfo{title}{One-shot visual reasoning on rpms with an application
  to video frame prediction}.
\newblock \bibinfo{journal}{arXiv preprint arXiv:2111.12301} .
\bibitem[{He et~al.(2021b)He, Ren and Bai}]{he2021two}
\bibinfo{author}{He, W.}, \bibinfo{author}{Ren, J.}, \bibinfo{author}{Bai, R.},
  \bibinfo{year}{2021}b.
\newblock \bibinfo{title}{Two-stage rule-induction visual reasoning on rpms
  with an application to video prediction}.
\newblock \bibinfo{journal}{arXiv preprint arXiv:2111.12301} .
\bibitem[{Hern{\'a}ndez-Orallo et~al.(2016)Hern{\'a}ndez-Orallo,
  Mart{\'\i}nez-Plumed, Schmid, Siebers and Dowe}]{hernandez2016computer}
\bibinfo{author}{Hern{\'a}ndez-Orallo, J.},
  \bibinfo{author}{Mart{\'\i}nez-Plumed, F.}, \bibinfo{author}{Schmid, U.},
  \bibinfo{author}{Siebers, M.}, \bibinfo{author}{Dowe, D.L.},
  \bibinfo{year}{2016}.
\newblock \bibinfo{title}{Computer models solving intelligence test problems:
  Progress and implications}.
\newblock \bibinfo{journal}{Artificial Intelligence} \bibinfo{volume}{230},
  \bibinfo{pages}{74--107}.
\bibitem[{Hersche et~al.(2022)Hersche, Zeqiri, Benini, Sebastian and
  Rahimi}]{hersche2022neuro}
\bibinfo{author}{Hersche, M.}, \bibinfo{author}{Zeqiri, M.},
  \bibinfo{author}{Benini, L.}, \bibinfo{author}{Sebastian, A.},
  \bibinfo{author}{Rahimi, A.}, \bibinfo{year}{2022}.
\newblock \bibinfo{title}{A neuro-vector-symbolic architecture for solving
  raven's progressive matrices}.
\newblock \bibinfo{journal}{arXiv preprint arXiv:2203.04571} .
\bibitem[{Hill et~al.(2019)Hill, Santoro, Barrett, Morcos and
  Lillicrap}]{hill2019learning}
\bibinfo{author}{Hill, F.}, \bibinfo{author}{Santoro, A.},
  \bibinfo{author}{Barrett, D.G.}, \bibinfo{author}{Morcos, A.S.},
  \bibinfo{author}{Lillicrap, T.}, \bibinfo{year}{2019}.
\newblock \bibinfo{title}{Learning to make analogies by contrasting abstract
  relational structure}.
\newblock \bibinfo{journal}{arXiv preprint arXiv:1902.00120} .
\bibitem[{Hofer(2004)}]{hofer2004matrixdeveloper}
\bibinfo{author}{Hofer, S.}, \bibinfo{year}{2004}.
\newblock \bibinfo{title}{Matrixdeveloper [unpublished computer software]}.
\newblock \bibinfo{note}{Unpublished software}.
\bibitem[{Hornke and Habon(1986)}]{hornke1986rule}
\bibinfo{author}{Hornke, L.F.}, \bibinfo{author}{Habon, M.W.},
  \bibinfo{year}{1986}.
\newblock \bibinfo{title}{Rule-based item bank construction and evaluation
  within the linear logistic framework}.
\newblock \bibinfo{journal}{Applied psychological measurement}
  \bibinfo{volume}{10}, \bibinfo{pages}{369--380}.
\bibitem[{Hoshen and Werman(2017)}]{hoshen2017iq}
\bibinfo{author}{Hoshen, D.}, \bibinfo{author}{Werman, M.},
  \bibinfo{year}{2017}.
\newblock \bibinfo{title}{Iq of neural networks}.
\newblock \bibinfo{journal}{arXiv preprint arXiv:1710.01692} .
\bibitem[{Hu et~al.(2021)Hu, Ma, Liu, Wei and Bai}]{hu2021stratified}
\bibinfo{author}{Hu, S.}, \bibinfo{author}{Ma, Y.}, \bibinfo{author}{Liu, X.},
  \bibinfo{author}{Wei, Y.}, \bibinfo{author}{Bai, S.}, \bibinfo{year}{2021}.
\newblock \bibinfo{title}{Stratified rule-aware network for abstract visual
  reasoning}, in: \bibinfo{booktitle}{Proceedings of the AAAI Conference on
  Artificial Intelligence}, pp. \bibinfo{pages}{1567--1574}.
\bibitem[{Hua and Kunda(2019)}]{hua2019modeling}
\bibinfo{author}{Hua, T.}, \bibinfo{author}{Kunda, M.}, \bibinfo{year}{2019}.
\newblock \bibinfo{title}{Modeling gestalt visual reasoning on the raven's
  progressive matrices intelligence test using generative image inpainting
  techniques}.
\newblock \bibinfo{journal}{arXiv preprint arXiv:1911.07736} .
\bibitem[{Hunt(1974)}]{hunt1974quote}
\bibinfo{author}{Hunt, E.}, \bibinfo{year}{1974}.
\newblock \bibinfo{title}{Quote the raven? nevermore.} .
\bibitem[{Hunt(2010)}]{hunt2010human}
\bibinfo{author}{Hunt, E.}, \bibinfo{year}{2010}.
\newblock \bibinfo{title}{Human intelligence}.
\newblock \bibinfo{publisher}{Cambridge University Press}.
\bibitem[{Jahrens and Martinetz(2018)}]{jahrens2018multi}
\bibinfo{author}{Jahrens, M.}, \bibinfo{author}{Martinetz, T.},
  \bibinfo{year}{2018}.
\newblock \bibinfo{title}{Multi-layer relation networks}.
\newblock \bibinfo{journal}{arXiv preprint arXiv:1811.01838} .
\bibitem[{Jahrens and Martinetz(2019)}]{jahrens2019multi}
\bibinfo{author}{Jahrens, M.}, \bibinfo{author}{Martinetz, T.},
  \bibinfo{year}{2019}.
\newblock \bibinfo{title}{Multi-layer relation networks for relational
  reasoning}, in: \bibinfo{booktitle}{Proceedings of the 2nd International
  Conference on Applications of Intelligent Systems}, pp.
  \bibinfo{pages}{1--5}.
\bibitem[{Jahrens and Martinetz(2020)}]{jahrens2020solving}
\bibinfo{author}{Jahrens, M.}, \bibinfo{author}{Martinetz, T.},
  \bibinfo{year}{2020}.
\newblock \bibinfo{title}{Solving raven’s progressive matrices with
  multi-layer relation networks}, in: \bibinfo{booktitle}{International Joint
  Conference on Neural Networks}, \bibinfo{organization}{IEEE}.
\bibitem[{Kiat et~al.(2020)Kiat, Wang and Jamnik}]{kiat2020pairwise}
\bibinfo{author}{Kiat, N.Q.W.}, \bibinfo{author}{Wang, D.},
  \bibinfo{author}{Jamnik, M.}, \bibinfo{year}{2020}.
\newblock \bibinfo{title}{Pairwise relations discriminator for unsupervised
  raven's progressive matrices}.
\newblock \bibinfo{journal}{arXiv preprint arXiv:2011.01306} .
\bibitem[{Kim et~al.(2020)Kim, Shin, Yang and Hwang}]{kim2020few}
\bibinfo{author}{Kim, Y.}, \bibinfo{author}{Shin, J.}, \bibinfo{author}{Yang,
  E.}, \bibinfo{author}{Hwang, S.J.}, \bibinfo{year}{2020}.
\newblock \bibinfo{title}{Few-shot visual reasoning with meta-analogical
  contrastive learning}.
\newblock \bibinfo{journal}{Advances in Neural Information Processing Systems}
  \bibinfo{volume}{33}, \bibinfo{pages}{16846--16856}.
\bibitem[{Kosslyn et~al.(2006)Kosslyn, Thompson and Ganis}]{kosslyn2006case}
\bibinfo{author}{Kosslyn, S.M.}, \bibinfo{author}{Thompson, W.L.},
  \bibinfo{author}{Ganis, G.}, \bibinfo{year}{2006}.
\newblock \bibinfo{title}{The case for mental imagery}.
\newblock \bibinfo{edition}{1st.} ed., \bibinfo{publisher}{Oxford University
  Press}, \bibinfo{address}{New York, NY}.
\bibitem[{Kunda(2013)}]{kunda2013visual}
\bibinfo{author}{Kunda, M.}, \bibinfo{year}{2013}.
\newblock \bibinfo{title}{Visual problem solving in autism, psychometrics, and
  AI: The case of the Raven's progressive matrices intelligence test}.
\newblock Ph.D. thesis. Georgia Institute of Technology.
  \bibinfo{address}{Atlanta, GA}.
\bibitem[{Kunda et~al.(2009)Kunda, McGreggor and Goel}]{kunda2009addressing}
\bibinfo{author}{Kunda, M.}, \bibinfo{author}{McGreggor, K.},
  \bibinfo{author}{Goel, A.}, \bibinfo{year}{2009}.
\newblock \bibinfo{title}{Addressing the raven’s progressive matrices test of
  “general” intelligence}, in: \bibinfo{booktitle}{2009 AAAI Fall Symposium
  Series}.
\bibitem[{Kunda et~al.(2010)Kunda, McGreggor and Goel}]{kunda2010taking}
\bibinfo{author}{Kunda, M.}, \bibinfo{author}{McGreggor, K.},
  \bibinfo{author}{Goel, A.}, \bibinfo{year}{2010}.
\newblock \bibinfo{title}{Taking a look (literally!) at the raven’s
  intelligence test: Two visual solution strategies}, in:
  \bibinfo{booktitle}{Proceedings of the Annual Meeting of the Cognitive
  Science Society}.
\bibitem[{Kunda et~al.(2013)Kunda, McGreggor and Goel}]{kunda2013computational}
\bibinfo{author}{Kunda, M.}, \bibinfo{author}{McGreggor, K.},
  \bibinfo{author}{Goel, A.K.}, \bibinfo{year}{2013}.
\newblock \bibinfo{title}{A computational model for solving problems from the
  raven’s progressive matrices intelligence test using iconic visual
  representations}.
\newblock \bibinfo{journal}{Cognitive Systems Research} \bibinfo{volume}{22},
  \bibinfo{pages}{47--66}.
\bibitem[{Laird et~al.(2017)Laird, Gluck, Anderson, Forbus, Jenkins, Lebiere,
  Salvucci, Scheutz, Thomaz, Trafton et~al.}]{laird2017interactive}
\bibinfo{author}{Laird, J.E.}, \bibinfo{author}{Gluck, K.},
  \bibinfo{author}{Anderson, J.}, \bibinfo{author}{Forbus, K.D.},
  \bibinfo{author}{Jenkins, O.C.}, \bibinfo{author}{Lebiere, C.},
  \bibinfo{author}{Salvucci, D.}, \bibinfo{author}{Scheutz, M.},
  \bibinfo{author}{Thomaz, A.}, \bibinfo{author}{Trafton, G.}, et~al.,
  \bibinfo{year}{2017}.
\newblock \bibinfo{title}{Interactive task learning}.
\newblock \bibinfo{journal}{IEEE Intelligent Systems} \bibinfo{volume}{32},
  \bibinfo{pages}{6--21}.
\bibitem[{Lovett et~al.(2007)Lovett, Forbus and Usher}]{lovett2007analogy}
\bibinfo{author}{Lovett, A.}, \bibinfo{author}{Forbus, K.},
  \bibinfo{author}{Usher, J.}, \bibinfo{year}{2007}.
\newblock \bibinfo{title}{Analogy with qualitative spatial representations can
  simulate solving raven's progressive matrices}, in:
  \bibinfo{booktitle}{Proceedings of the Annual Meeting of the Cognitive
  Science Society}.
\bibitem[{Lovett et~al.(2009)Lovett, Tomai, Forbus and
  Usher}]{lovett2009solving}
\bibinfo{author}{Lovett, A.}, \bibinfo{author}{Tomai, E.},
  \bibinfo{author}{Forbus, K.}, \bibinfo{author}{Usher, J.},
  \bibinfo{year}{2009}.
\newblock \bibinfo{title}{Solving geometric analogy problems through two-stage
  analogical mapping}.
\newblock \bibinfo{journal}{Cognitive science} \bibinfo{volume}{33},
  \bibinfo{pages}{1192--1231}.
\bibitem[{Ma{\l}ki{\'n}ski and Ma{\'n}dziuk(2020)}]{malkinski2020multi}
\bibinfo{author}{Ma{\l}ki{\'n}ski, M.}, \bibinfo{author}{Ma{\'n}dziuk, J.},
  \bibinfo{year}{2020}.
\newblock \bibinfo{title}{Multi-label contrastive learning for abstract visual
  reasoning}.
\newblock \bibinfo{journal}{arXiv preprint arXiv:2012.01944} .
\bibitem[{Ma{\'n}dziuk and {\.Z}ychowski(2019)}]{mandziuk2019deepiq}
\bibinfo{author}{Ma{\'n}dziuk, J.}, \bibinfo{author}{{\.Z}ychowski, A.},
  \bibinfo{year}{2019}.
\newblock \bibinfo{title}{Deepiq: A human-inspired ai system for solving iq
  test problems}, in: \bibinfo{booktitle}{2019 International Joint Conference
  on Neural Networks (IJCNN)}, \bibinfo{organization}{IEEE}. pp.
  \bibinfo{pages}{1--8}.
\bibitem[{Matzen et~al.(2010)Matzen, Benz, Dixon, Posey, Kroger and
  Speed}]{matzen2010recreating}
\bibinfo{author}{Matzen, L.E.}, \bibinfo{author}{Benz, Z.O.},
  \bibinfo{author}{Dixon, K.R.}, \bibinfo{author}{Posey, J.},
  \bibinfo{author}{Kroger, J.K.}, \bibinfo{author}{Speed, A.E.},
  \bibinfo{year}{2010}.
\newblock \bibinfo{title}{Recreating raven’s: Software for systematically
  generating large numbers of raven-like matrix problems with normed
  properties}.
\newblock \bibinfo{journal}{Behavior research methods} \bibinfo{volume}{42},
  \bibinfo{pages}{525--541}.
\bibitem[{McGreggor et~al.(2014)McGreggor, Kunda and
  Goel}]{mcgreggor2014fractals}
\bibinfo{author}{McGreggor, K.}, \bibinfo{author}{Kunda, M.},
  \bibinfo{author}{Goel, A.}, \bibinfo{year}{2014}.
\newblock \bibinfo{title}{Fractals and ravens}.
\newblock \bibinfo{journal}{Artificial Intelligence} \bibinfo{volume}{215},
  \bibinfo{pages}{1--23}.
\bibitem[{Mekik et~al.(2017)Mekik, Sun and Dai}]{Mekik2017deep}
\bibinfo{author}{Mekik, C.S.}, \bibinfo{author}{Sun, R.}, \bibinfo{author}{Dai,
  D.Y.}, \bibinfo{year}{2017}.
\newblock \bibinfo{title}{Deep learning of raven's matrices}, in:
  \bibinfo{booktitle}{Proceedings of the fifth Annual Conference on Advances in
  Cognitive Systems}.
\bibitem[{Mekik et~al.(2018)Mekik, Sun and Dai}]{mekik2018similarity}
\bibinfo{author}{Mekik, C.S.}, \bibinfo{author}{Sun, R.}, \bibinfo{author}{Dai,
  D.Y.}, \bibinfo{year}{2018}.
\newblock \bibinfo{title}{Similarity-based reasoning, raven's matrices, and
  general intelligence}, in: \bibinfo{booktitle}{Proceedings of the 27th
  International Joint Conference on Artificial Intelligence},
  \bibinfo{publisher}{AAAI Press}. p. \bibinfo{pages}{1576–1582}.
\bibitem[{Meo et~al.(2007)Meo, Roberts and Marucci}]{meo2007element}
\bibinfo{author}{Meo, M.}, \bibinfo{author}{Roberts, M.J.},
  \bibinfo{author}{Marucci, F.S.}, \bibinfo{year}{2007}.
\newblock \bibinfo{title}{Element salience as a predictor of item difficulty
  for raven's progressive matrices}.
\newblock \bibinfo{journal}{Intelligence} \bibinfo{volume}{35},
  \bibinfo{pages}{359--368}.
\bibitem[{Mulholland et~al.(1980)Mulholland, Pellegrino and
  Glaser}]{mulholland1980components}
\bibinfo{author}{Mulholland, T.M.}, \bibinfo{author}{Pellegrino, J.W.},
  \bibinfo{author}{Glaser, R.}, \bibinfo{year}{1980}.
\newblock \bibinfo{title}{Components of geometric analogy solution}.
\newblock \bibinfo{journal}{Cognitive psychology} \bibinfo{volume}{12},
  \bibinfo{pages}{252--284}.
\bibitem[{Naglieri et~al.(2014)Naglieri, Das and
  Goldstein}]{naglieri2014cognitive}
\bibinfo{author}{Naglieri, J.A.}, \bibinfo{author}{Das, J.P.},
  \bibinfo{author}{Goldstein, S.}, \bibinfo{year}{2014}.
\newblock \bibinfo{title}{Cognitive assessment system: Interpretive and
  technical manual}.
\newblock \bibinfo{publisher}{Pro-ed}.
\bibitem[{Pekar et~al.(2020)Pekar, Benny and Wolf}]{pekar2020generating}
\bibinfo{author}{Pekar, N.}, \bibinfo{author}{Benny, Y.},
  \bibinfo{author}{Wolf, L.}, \bibinfo{year}{2020}.
\newblock \bibinfo{title}{Generating correct answers for progressive matrices
  intelligence tests}.
\newblock \bibinfo{journal}{Advances in Neural Information Processing Systems}
  \bibinfo{volume}{33}, \bibinfo{pages}{7390--7400}.
\bibitem[{Primi(2001)}]{primi2001complexity}
\bibinfo{author}{Primi, R.}, \bibinfo{year}{2001}.
\newblock \bibinfo{title}{Complexity of geometric inductive reasoning tasks:
  Contribution to the understanding of fluid intelligence}.
\newblock \bibinfo{journal}{Intelligence} \bibinfo{volume}{30},
  \bibinfo{pages}{41--70}.
\bibitem[{Ragni and Neubert(2012)}]{ragni2012solving}
\bibinfo{author}{Ragni, M.}, \bibinfo{author}{Neubert, S.},
  \bibinfo{year}{2012}.
\newblock \bibinfo{title}{Solving raven's iq-tests: an ai and cognitive
  modeling approach}, in: \bibinfo{booktitle}{ECAI 2012}.
  \bibinfo{publisher}{IOS Press}, pp. \bibinfo{pages}{666--671}.
\bibitem[{Ragni and Neubert(2014)}]{ragni2014analyzing}
\bibinfo{author}{Ragni, M.}, \bibinfo{author}{Neubert, S.},
  \bibinfo{year}{2014}.
\newblock \bibinfo{title}{Analyzing raven’s intelligence test: Cognitive
  model, demand, and complexity}, in: \bibinfo{booktitle}{Computational
  Approaches to Analogical Reasoning: Current Trends}.
  \bibinfo{publisher}{Springer}, pp. \bibinfo{pages}{351--370}.
\bibitem[{Ragni et~al.(2007)Ragni, Schleipen and
  Steffenhagen}]{ragni2007solving}
\bibinfo{author}{Ragni, M.}, \bibinfo{author}{Schleipen, S.},
  \bibinfo{author}{Steffenhagen, F.}, \bibinfo{year}{2007}.
\newblock \bibinfo{title}{Solving proportional analogies: A computational
  model}.
\newblock \bibinfo{journal}{Analogies: Integrating Multiple Cognitive
  Abilities} \bibinfo{volume}{5}, \bibinfo{pages}{51}.
\bibitem[{Rahaman et~al.(2021)Rahaman, Gondal, Joshi, Gehler, Bengio, Locatello
  and Sch{\"o}lkopf}]{rahaman2021dynamic}
\bibinfo{author}{Rahaman, N.}, \bibinfo{author}{Gondal, M.W.},
  \bibinfo{author}{Joshi, S.}, \bibinfo{author}{Gehler, P.},
  \bibinfo{author}{Bengio, Y.}, \bibinfo{author}{Locatello, F.},
  \bibinfo{author}{Sch{\"o}lkopf, B.}, \bibinfo{year}{2021}.
\newblock \bibinfo{title}{Dynamic inference with neural interpreters}.
\newblock \bibinfo{journal}{Advances in Neural Information Processing Systems}
  \bibinfo{volume}{34}.
\bibitem[{Rasmussen and Eliasmith(2011)}]{rasmussen2011neural}
\bibinfo{author}{Rasmussen, D.}, \bibinfo{author}{Eliasmith, C.},
  \bibinfo{year}{2011}.
\newblock \bibinfo{title}{A neural model of rule generation in inductive
  reasoning}.
\newblock \bibinfo{journal}{Topics in Cognitive Science} \bibinfo{volume}{3},
  \bibinfo{pages}{140--153}.
\bibitem[{Raudies and Hasselmo(2017)}]{raudies2017model}
\bibinfo{author}{Raudies, F.}, \bibinfo{author}{Hasselmo, M.E.},
  \bibinfo{year}{2017}.
\newblock \bibinfo{title}{A model of symbolic processing in raven’s
  progressive matrices}.
\newblock \bibinfo{journal}{Biologically Inspired Cognitive Architectures}
  \bibinfo{volume}{21}, \bibinfo{pages}{47--58}.
\bibitem[{Raven(2000)}]{raven2000raven}
\bibinfo{author}{Raven, J.}, \bibinfo{year}{2000}.
\newblock \bibinfo{title}{The raven's progressive matrices: change and
  stability over culture and time}.
\newblock \bibinfo{journal}{Cognitive psychology} \bibinfo{volume}{41},
  \bibinfo{pages}{1--48}.
\bibitem[{Raven(2008)}]{raven2008raven}
\bibinfo{author}{Raven, J.}, \bibinfo{year}{2008}.
\newblock \bibinfo{title}{The raven progressive matrices tests: their
  theoretical basis and measurement model} .
\bibitem[{Raven(1936)}]{raven1936mental}
\bibinfo{author}{Raven, J.C.}, \bibinfo{year}{1936}.
\newblock \bibinfo{title}{Mental tests used in genetic studies: The performance
  of related individuals on tests mainly educative and mainly reproductive}.
\newblock \bibinfo{journal}{Unpublished master’s thesis, University of
  London} .
\bibitem[{Raven(1941)}]{raven1941standardization}
\bibinfo{author}{Raven, J.C.}, \bibinfo{year}{1941}.
\newblock \bibinfo{title}{Standardization of progressive matrices, 1938}.
\newblock \bibinfo{journal}{British Journal of Medical Psychology}
  \bibinfo{volume}{19}, \bibinfo{pages}{137--150}.
\bibitem[{Roid and Miller(1997)}]{leiter1997manual}
\bibinfo{author}{Roid, G.H.}, \bibinfo{author}{Miller, L.J.},
  \bibinfo{year}{1997}.
\newblock \bibinfo{title}{Leiter International Performance Scale-Revised:
  Examiners Manual}.
\newblock \bibinfo{publisher}{Stoelting}.
\bibitem[{Schwering et~al.(2007)Schwering, Krumnack, Kuhnberger and
  Gust}]{schwering2007using}
\bibinfo{author}{Schwering, A.}, \bibinfo{author}{Krumnack, U.},
  \bibinfo{author}{Kuhnberger, K.U.}, \bibinfo{author}{Gust, H.},
  \bibinfo{year}{2007}.
\newblock \bibinfo{title}{Using gestalt principles to compute analogies of
  geometric figures}, in: \bibinfo{booktitle}{Proceedings of the Annual Meeting
  of the Cognitive Science Society}.
\bibitem[{Sekh et~al.(2020)Sekh, Dogra, Kar, Roy and Prasad}]{sekh2020can}
\bibinfo{author}{Sekh, A.A.}, \bibinfo{author}{Dogra, D.P.},
  \bibinfo{author}{Kar, S.}, \bibinfo{author}{Roy, P.P.},
  \bibinfo{author}{Prasad, D.K.}, \bibinfo{year}{2020}.
\newblock \bibinfo{title}{Can we automate diagrammatic reasoning?}
\newblock \bibinfo{journal}{Pattern Recognition} \bibinfo{volume}{106},
  \bibinfo{pages}{107412}.
\bibitem[{Shegheva(2018)}]{shegheva2018computational}
\bibinfo{author}{Shegheva, S.}, \bibinfo{year}{2018}.
\newblock \bibinfo{title}{A computational model for solving raven’s
  progressive matrices intelligence test}.
\newblock Master's thesis. Georgia Institute of Technology.
  \bibinfo{address}{Atlanta, GA}.
\bibitem[{Shekhar and Taylor(2021)}]{shekhar2021neural}
\bibinfo{author}{Shekhar, S.}, \bibinfo{author}{Taylor, G.W.},
  \bibinfo{year}{2021}.
\newblock \bibinfo{title}{Neural structure mapping for learning abstract visual
  analogies}, in: \bibinfo{booktitle}{SVRHM 2021 Workshop@ NeurIPS}.
\bibitem[{Shi et~al.(2021)Shi, Li and Xue}]{shi2021raven}
\bibinfo{author}{Shi, F.}, \bibinfo{author}{Li, B.}, \bibinfo{author}{Xue, X.},
  \bibinfo{year}{2021}.
\newblock \bibinfo{title}{Raven's progressive matrices completion with latent
  gaussian process priors}, in: \bibinfo{booktitle}{Proceedings of the AAAI
  Conference on Artificial Intelligence}, pp. \bibinfo{pages}{9612--9620}.
\bibitem[{Sinha et~al.(2020)Sinha, Webb and Cohen}]{sinha2020memory}
\bibinfo{author}{Sinha, I.}, \bibinfo{author}{Webb, T.W.},
  \bibinfo{author}{Cohen, J.D.}, \bibinfo{year}{2020}.
\newblock \bibinfo{title}{A memory-augmented neural network model of abstract
  rule learning}.
\newblock \bibinfo{journal}{arXiv preprint arXiv:2012.07172} .
\bibitem[{Snow et~al.(1984)Snow, Kyllonen, Marshalek
  et~al.}]{snow1984topography}
\bibinfo{author}{Snow, R.E.}, \bibinfo{author}{Kyllonen, P.C.},
  \bibinfo{author}{Marshalek, B.}, et~al., \bibinfo{year}{1984}.
\newblock \bibinfo{title}{The topography of ability and learning correlations}.
\newblock \bibinfo{journal}{Advances in the psychology of human intelligence}
  \bibinfo{volume}{2}, \bibinfo{pages}{103}.
\bibitem[{Spratley et~al.(2020)Spratley, Ehinger and
  Miller}]{spratley2020closer}
\bibinfo{author}{Spratley, S.}, \bibinfo{author}{Ehinger, K.},
  \bibinfo{author}{Miller, T.}, \bibinfo{year}{2020}.
\newblock \bibinfo{title}{A closer look at generalisation in raven}, in:
  \bibinfo{booktitle}{Computer Vision--ECCV 2020: 16th European Conference,
  Glasgow, UK, August 23--28, 2020, Proceedings, Part XXVII 16},
  \bibinfo{organization}{Springer}. pp. \bibinfo{pages}{601--616}.
\bibitem[{Stabinger et~al.(2021)Stabinger, Peer, Piater and
  Rodr{\'\i}guez-S{\'a}nchez}]{stabinger2021evaluating}
\bibinfo{author}{Stabinger, S.}, \bibinfo{author}{Peer, D.},
  \bibinfo{author}{Piater, J.}, \bibinfo{author}{Rodr{\'\i}guez-S{\'a}nchez,
  A.}, \bibinfo{year}{2021}.
\newblock \bibinfo{title}{Evaluating the progress of deep learning for visual
  relational concepts}.
\newblock \bibinfo{journal}{Journal of Vision} \bibinfo{volume}{21},
  \bibinfo{pages}{8--8}.
\bibitem[{Steenbrugge et~al.(2018)Steenbrugge, Leroux, Verbelen and
  Dhoedt}]{steenbrugge2018improving}
\bibinfo{author}{Steenbrugge, X.}, \bibinfo{author}{Leroux, S.},
  \bibinfo{author}{Verbelen, T.}, \bibinfo{author}{Dhoedt, B.},
  \bibinfo{year}{2018}.
\newblock \bibinfo{title}{Improving generalization for abstract reasoning tasks
  using disentangled feature representations}.
\newblock \bibinfo{journal}{arXiv preprint arXiv:1811.04784} .
\bibitem[{van Steenkiste et~al.(2019)van Steenkiste, Locatello, Schmidhuber and
  Bachem}]{van2019disentangled}
\bibinfo{author}{van Steenkiste, S.}, \bibinfo{author}{Locatello, F.},
  \bibinfo{author}{Schmidhuber, J.}, \bibinfo{author}{Bachem, O.},
  \bibinfo{year}{2019}.
\newblock \bibinfo{title}{Are disentangled representations helpful for abstract
  visual reasoning?}
\newblock \bibinfo{journal}{arXiv preprint arXiv:1905.12506} .
\bibitem[{Stranneg{\aa}rd et~al.(2013)Stranneg{\aa}rd, Cirillo and
  Str{\"o}m}]{strannegaard2013anthropomorphic}
\bibinfo{author}{Stranneg{\aa}rd, C.}, \bibinfo{author}{Cirillo, S.},
  \bibinfo{author}{Str{\"o}m, V.}, \bibinfo{year}{2013}.
\newblock \bibinfo{title}{An anthropomorphic method for progressive matrix
  problems}.
\newblock \bibinfo{journal}{Cognitive Systems Research} \bibinfo{volume}{22},
  \bibinfo{pages}{35--46}.
\bibitem[{\textbf{Yang, Yuan} et~al.(2022)\textbf{Yang, Yuan}, Sanyal,
  Michelson, Ainooson and Kunda}]{yang2022end}
\bibinfo{author}{\textbf{Yang, Yuan}}, \bibinfo{author}{Sanyal, D.},
  \bibinfo{author}{Michelson, J.}, \bibinfo{author}{Ainooson, J.},
  \bibinfo{author}{Kunda, M.}, \bibinfo{year}{2022}.
\newblock \bibinfo{title}{An end-to-end imagery-based modeling of solving
  geometric analogy problems}, in: \bibinfo{booktitle}{Proceedings of the
  Annual Meeting of the Cognitive Science Society}.
\bibitem[{Tomai et~al.(2005)Tomai, Lovett, Forbus and
  Usher}]{tomai2005structure}
\bibinfo{author}{Tomai, E.}, \bibinfo{author}{Lovett, A.},
  \bibinfo{author}{Forbus, K.D.}, \bibinfo{author}{Usher, J.},
  \bibinfo{year}{2005}.
\newblock \bibinfo{title}{A structure mapping model for solving geometric
  analogy problems} .
\bibitem[{Wang et~al.(2020)Wang, Jamnik and Lio}]{wang2020abstract}
\bibinfo{author}{Wang, D.}, \bibinfo{author}{Jamnik, M.}, \bibinfo{author}{Lio,
  P.}, \bibinfo{year}{2020}.
\newblock \bibinfo{title}{Abstract diagrammatic reasoning with multiplex graph
  networks}.
\newblock \bibinfo{journal}{arXiv preprint arXiv:2006.11197} .
\bibitem[{Wang and Su(2015)}]{wang2015automatic}
\bibinfo{author}{Wang, K.}, \bibinfo{author}{Su, Z.}, \bibinfo{year}{2015}.
\newblock \bibinfo{title}{Automatic generation of raven’s progressive
  matrices}, in: \bibinfo{booktitle}{Twenty-Fourth International Joint
  Conference on Artificial Intelligence}.
\bibitem[{Webb et~al.(2020)Webb, Sinha and Cohen}]{webb2020emergent}
\bibinfo{author}{Webb, T.W.}, \bibinfo{author}{Sinha, I.},
  \bibinfo{author}{Cohen, J.}, \bibinfo{year}{2020}.
\newblock \bibinfo{title}{Emergent symbols through binding in external memory},
  in: \bibinfo{booktitle}{International Conference on Learning
  Representations}.
\bibitem[{Wechsler et~al.(2008)Wechsler, Coalson and
  Faiford}]{wechsler2008technical}
\bibinfo{author}{Wechsler, D.}, \bibinfo{author}{Coalson, D.L.},
  \bibinfo{author}{Faiford, S.E.}, \bibinfo{year}{2008}.
\newblock \bibinfo{title}{Wechsler Adult Intelligence Scale—Fourth Edition
  Technical and Interpretive Manual}.
\newblock \bibinfo{publisher}{Pearson}.
\bibitem[{Wu et~al.(2021)Wu, Dong, Grosse and Ba}]{wu2021scattering}
\bibinfo{author}{Wu, Y.}, \bibinfo{author}{Dong, H.}, \bibinfo{author}{Grosse,
  R.}, \bibinfo{author}{Ba, J.}, \bibinfo{year}{2021}.
\newblock \bibinfo{title}{The scattering compositional learner: Discovering
  objects, attributes, relationships in analogical reasoning}.
\newblock \bibinfo{journal}{arXiv preprint arXiv:2007.04212} .
\bibitem[{Yang et~al.(2020)Yang, McGreggor and Kunda}]{yang2020not}
\bibinfo{author}{Yang, Y.}, \bibinfo{author}{McGreggor, K.},
  \bibinfo{author}{Kunda, M.}, \bibinfo{year}{2020}.
\newblock \bibinfo{title}{Not quite any way you slice it: How different
  analogical constructions affect raven's matrices performance}, in:
  \bibinfo{booktitle}{Proceedings of the Eighth Annual Conference on Advances
  in Cognitive Systems (ACS)}.
\bibitem[{Yu et~al.(2021)Yu, Mo, Ahn and Shin}]{yu2021abstract}
\bibinfo{author}{Yu, S.}, \bibinfo{author}{Mo, S.}, \bibinfo{author}{Ahn, S.},
  \bibinfo{author}{Shin, J.}, \bibinfo{year}{2021}.
\newblock \bibinfo{title}{Abstract reasoning via logic-guided generation}.
\newblock \bibinfo{journal}{arXiv preprint arXiv:2107.10493} .
\bibitem[{Zhang et~al.(2019a)Zhang, Gao, Jia, Zhu and Zhu}]{zhang2019raven}
\bibinfo{author}{Zhang, C.}, \bibinfo{author}{Gao, F.}, \bibinfo{author}{Jia,
  B.}, \bibinfo{author}{Zhu, Y.}, \bibinfo{author}{Zhu, S.C.},
  \bibinfo{year}{2019}a.
\newblock \bibinfo{title}{Raven: A dataset for relational and analogical visual
  reasoning}, in: \bibinfo{booktitle}{Proceedings of the IEEE/CVF Conference on
  Computer Vision and Pattern Recognition}, pp. \bibinfo{pages}{5317--5327}.
\bibitem[{Zhang et~al.(2019b)Zhang, Jia, Gao, Zhu, Lu and
  Zhu}]{zhang2019learning}
\bibinfo{author}{Zhang, C.}, \bibinfo{author}{Jia, B.}, \bibinfo{author}{Gao,
  F.}, \bibinfo{author}{Zhu, Y.}, \bibinfo{author}{Lu, H.},
  \bibinfo{author}{Zhu, S.C.}, \bibinfo{year}{2019}b.
\newblock \bibinfo{title}{Learning perceptual inference by contrasting}.
\bibitem[{Zhang et~al.(2021)Zhang, Jia, Zhu and Zhu}]{zhang2021abstract}
\bibinfo{author}{Zhang, C.}, \bibinfo{author}{Jia, B.}, \bibinfo{author}{Zhu,
  S.C.}, \bibinfo{author}{Zhu, Y.}, \bibinfo{year}{2021}.
\newblock \bibinfo{title}{Abstract spatial-temporal reasoning via probabilistic
  abduction and execution}, in: \bibinfo{booktitle}{Proceedings of the IEEE/CVF
  Conference on Computer Vision and Pattern Recognition}, pp.
  \bibinfo{pages}{9736--9746}.
\bibitem[{Zhang et~al.(2020)Zhang, Xie, Jia, Zhu, Wu and
  Zhu}]{zhang2020learning}
\bibinfo{author}{Zhang, C.}, \bibinfo{author}{Xie, S.}, \bibinfo{author}{Jia,
  B.}, \bibinfo{author}{Zhu, Y.}, \bibinfo{author}{Wu, Y.N.},
  \bibinfo{author}{Zhu, S.C.}, \bibinfo{year}{2020}.
\newblock \bibinfo{title}{Learning algebraic representation for abstract
  spatial-temporal reasoning} .
\bibitem[{Zheng et~al.(2019)Zheng, Zha and Wei}]{zheng2019abstract}
\bibinfo{author}{Zheng, K.}, \bibinfo{author}{Zha, Z.J.}, \bibinfo{author}{Wei,
  W.}, \bibinfo{year}{2019}.
\newblock \bibinfo{title}{Abstract reasoning with distracting features}.
\newblock \bibinfo{journal}{Advances in Neural Information Processing Systems}
  \bibinfo{volume}{32}.
\bibitem[{Zhuo and Kankanhalli(2020)}]{zhuo2020solving}
\bibinfo{author}{Zhuo, T.}, \bibinfo{author}{Kankanhalli, M.},
  \bibinfo{year}{2020}.
\newblock \bibinfo{title}{Solving raven's progressive matrices with neural
  networks}.
\newblock \bibinfo{journal}{arXiv preprint arXiv:2002.01646} .
\bibitem[{Zhuo and Kankanhalli(2021)}]{zhuo2021effective}
\bibinfo{author}{Zhuo, T.}, \bibinfo{author}{Kankanhalli, M.},
  \bibinfo{year}{2021}.
\newblock \bibinfo{title}{Effective abstract reasoning with dual-contrast
  network}, in: \bibinfo{booktitle}{International Conference on Learning
  Representations}.

\end{thebibliography}

\end{document}